\documentclass{article}
\usepackage{times}
\usepackage{amsmath,amssymb}
\usepackage{mathrsfs}
\usepackage{mathabx}\changenotsign 
\usepackage{dsfont}
\usepackage{bbm}
\usepackage{scalerel}
\usepackage{graphicx}
\usepackage{algorithm}
\usepackage{adjustbox}
\usepackage{subfiles}
\usepackage{array}
\usepackage[noend]{algpseudocode}
\usepackage[T1]{fontenc}
\usepackage{nicefrac}
\usepackage{booktabs} 
\usepackage{amsfonts}
\usepackage{etex}
\usepackage{natbib}
\usepackage{prettyref}

\newcommand{\DERIV}{\mathrm{d}}

\newcommand{\LCHAR}{\L\text{ }}

\newcommand{\OPNORM}{\mathrm{op}{}}

\newcommand{\vecW}{\mathbf{w}}

\newcommand{\vecB}{\mathbf{B}}

\newcommand{\vecOrigin}{\vec{\mathbf{0}}}
\newcommand{\vecEps}{\pmb{\varepsilon}}

\newcommand{\CPI}{\normalfont \textsf{C}_{\textsc{pi}}}
\newcommand{\CWPI}{\normalfont \textsf{C}_{\textsc{wpi}}}
\newcommand{\CLSI}{\normalfont \textsf{C}_{\textsc{lsi}}}
\newcommand{\CPILOCAL}{\normalfont \textsf{C}_{\textsc{pi, local}}}

\newcommand{\deltalocal}{\normalfont \delta_{\textsc{local}}}
\newcommand{\CWPILOCAL}{\normalfont \textsf{C}_{\textsc{wpi, local}}}
\newcommand{\lambdareg}{\lambda_{\textsc{reg}}}
\newcommand{\cwgt}{c_{\textsc{wgt}}}

\newcommand{\tthresF}{t_{\textsc{thres}, F}}

\newcommand{\mprimenew}{m_{\textsc{new}}'}
\newcommand{\bprimenew}{b_{\textsc{new}}'}

\newcommand{\DIST}{\normalfont \textsf{d}}
\newcommand{\KL}{\normalfont \textsf{KL}}
\newcommand{\TV}{\normalfont \textsf{TV}}
\newcommand{\mubeta}{\mu_{\beta}}
\newcommand{\mubetalocal}{\mu_{\beta, \textsc{local}}}
\newcommand{\osc}{\text{osc}}

\newcommand{\MIN}{\mathrm{min}}
\newcommand{\tr}{\mathrm{tr}}

\newcommand{\matI}{\pmb{I}}

\newcommand{\matM}{\pmb{M}}
\newcommand{\poly}{\normalfont \text{poly}}
\newcommand{\matX}{\pmb{X}}
\newcommand{\vecy}{\pmb{y}}
\newcommand{\vecx}{\pmb{x}}
\newcommand{\vecTheta}{\pmb{\theta}}

\usepackage[letterpaper, left=1in, right=1in, top=1in, bottom=1in]{geometry}
\PassOptionsToPackage{hypertexnames=false}{hyperref}  % compatible with cref for multiple algo
\usepackage{parskip}
\usepackage[dvipsnames]{xcolor}
\usepackage[colorlinks=true, linkcolor=blue!70!black, citecolor=blue!70!black,urlcolor=black,breaklinks=true]{hyperref}
\hypersetup{linktoc=all}

\usepackage{microtype}
\usepackage{hhline}

\makeatletter
\newcommand{\neutralize}[1]{\expandafter\let\csname c@#1\endcsname\count@}
\makeatother

\usepackage{algorithm}
\usepackage{natbib}
\bibpunct{(}{)}{;}{a}{,}{,}
\usepackage{amsthm}
\usepackage{mathtools}
\usepackage{amsmath}
\usepackage{bbm}
\usepackage{amsfonts}
\usepackage{amssymb}

\usepackage{xpatch}

\usepackage{thmtools}
\usepackage{thm-restate}
\declaretheorem[name=Theorem,parent=section]{theorem}
\declaretheorem[name=Lemma,parent=section]{lemma}
\declaretheorem[name=Assumption, parent=section]{assumption}
\declaretheorem[name=Condition, parent=section]{condition}
\declaretheorem[name=Example,style=definition]{example}
\declaretheorem[name=Remark, style=definition]{remark}
\declaretheorem[name=Proposition, parent=section]{proposition}

\makeatletter
  \renewenvironment{proof}[1][Proof]%
  {%
   \par\noindent{\bfseries\upshape {#1.}\ }%
  }%
  {\qed \\ \newline}
  \makeatother

\newtheorem{definition}{Definition}[section]

\xpatchcmd{\proof}{\itshape}{\normalfont\proofnameformat}{}{}
\usepackage[nameinlink,capitalize]{cleveref}

\renewcommand{\eqref}[1]{\texorpdfstring{\hyperref[#1]{Eq.~(\ref*{#1})}}{Eq.~(\ref*{#1})}}

\crefformat{equation}{#2(#1)#3}
\Crefformat{equation}{#2(#1)#3}

\Crefformat{figure}{#2Figure #1#3}
\Crefname{assumption}{Assumption}{Assumptions}
\Crefformat{assumption}{#2Assumption #1#3}
\Crefname{subsubsection}{Section}{Sections}
\crefformat{subsubsection}{#2Section #1#3}
\Crefformat{subsubsection}{#2Section #1#3}

\usepackage{crossreftools}
\usepackage{xparse}

\ExplSyntaxOn
\DeclareDocumentCommand{\XDeclarePairedDelimiter}{mm}
 {
  \__egreg_delimiter_clear_keys: % reset to the default
  \keys_set:nn { egreg/delimiters } { #2 }
  \use:x % we want to expand the values of the token variables set with the keys
   {
    \exp_not:n {\NewDocumentCommand{#1}{sO{}m} }
     {
      \exp_not:n { \IfBooleanTF{##1} }
       {
        \exp_not:N \egreg_paired_delimiter_expand:nnnn
         { \exp_not:V \l_egreg_delimiter_left_tl }
         { \exp_not:V \l_egreg_delimiter_right_tl }
         { \exp_not:n { ##3 } }
         { \exp_not:V \l_egreg_delimiter_subscript_tl }
       }
       {
        \exp_not:N \egreg_paired_delimiter_fixed:nnnnn 
         { \exp_not:n { ##2 } }
         { \exp_not:V \l_egreg_delimiter_left_tl }
         { \exp_not:V \l_egreg_delimiter_right_tl }
         { \exp_not:n { ##3 } }
         { \exp_not:V \l_egreg_delimiter_subscript_tl }
       }
     }
   }
 }

\keys_define:nn { egreg/delimiters }
 {
  left      .tl_set:N = \l_egreg_delimiter_left_tl,
  right     .tl_set:N = \l_egreg_delimiter_right_tl,
  subscript .tl_set:N = \l_egreg_delimiter_subscript_tl,
 }

\cs_new_protected:Npn \__egreg_delimiter_clear_keys:
 {
  \keys_set:nn { egreg/delimiters } { left=.,right=.,subscript={} }
 }

\cs_new_protected:Npn \egreg_paired_delimiter_expand:nnnn #1 #2 #3 #4
 {% Fix the spacing issue with \left and \right (D. Arsenau, P. Stephani and H. Oberdiek)
  \mathopen{}
  \mathclose\c_group_begin_token
   \left#1
   #3
   \group_insert_after:N \c_group_end_token
   \right#2
   \tl_if_empty:nF {#4} { \c_math_subscript_token {#4} }
 }
\cs_new_protected:Npn \egreg_paired_delimiter_fixed:nnnnn #1 #2 #3 #4 #5
 {
  \mathopen{#1#2}#4\mathclose{#1#3}
  \tl_if_empty:nF {#5} { \c_math_subscript_token {#5} }
 }
\ExplSyntaxOff

\XDeclarePairedDelimiter{\supnorm}{
  left=\lVert,
  right=\rVert,
  subscript=\infty
  }
\usepackage{mathtools}
\usepackage{amsmath}
\usepackage{amsxtra}
\usepackage{enumitem}      
\usepackage{mathtools}
\usepackage{bbm}
\usepackage{amsfonts}
\usepackage{amssymb}
\let\vec\undefined
\usepackage{MnSymbol} %Actually conflicts with amssymb and others

\usepackage{xpatch}
\newcommand\blfootnote[1]{%
  \begingroup
  \renewcommand\thefootnote{}\footnote{#1}%
  \addtocounter{footnote}{-1}%
  \endgroup
}

{\newtheorem{conjecture}{Conjecture}} 
{\newtheorem{corollary}{Corollary}} 
\usepackage{prettyref}
\newcommand{\pref}[1]{\prettyref{#1}}

\newcommand{\savehyperref}[2]{\texorpdfstring{\hyperref[#1]{#2}}{#2}}
\newrefformat{eq}{\savehyperref{#1}{\textup{(\ref*{#1})}}}
\newrefformat{eqn}{\savehyperref{#1}{Equation~\ref*{#1}}}
\newrefformat{con}{\savehyperref{#1}{Conjecture~\ref*{#1}}}
\newrefformat{lem}{\savehyperref{#1}{Lemma~\ref*{#1}}}
\newrefformat{def}{\savehyperref{#1}{Definition~\ref*{#1}}}
\newrefformat{line}{\savehyperref{#1}{line~\ref*{#1}}}
\newrefformat{thm}{\savehyperref{#1}{Theorem~\ref*{#1}}}
\newrefformat{corr}{\savehyperref{#1}{Corollary~\ref*{#1}}}
\newrefformat{sec}{\savehyperref{#1}{Section~\ref*{#1}}}
\newrefformat{app}{\savehyperref{#1}{Appendix~\ref*{#1}}}
\newrefformat{ass}{\savehyperref{#1}{Assumption~\ref*{#1}}}
\newrefformat{ex}{\savehyperref{#1}{Example~\ref*{#1}}}
\newrefformat{fig}{\savehyperref{#1}{Figure~\ref*{#1}}}
\newrefformat{alg}{\savehyperref{#1}{Algorithm~\ref*{#1}}}
\newrefformat{rem}{\savehyperref{#1}{Remark~\ref*{#1}}}
\newrefformat{conj}{\savehyperref{#1}{Conjecture~\ref*{#1}}}
\newrefformat{prop}{\savehyperref{#1}{Proposition~\ref*{#1}}}
\newrefformat{proto}{\savehyperref{#1}{Protocol~\ref*{#1}}}
\newrefformat{prob}{\savehyperref{#1}{Problem~\ref*{#1}}}
\newrefformat{claim}{\savehyperref{#1}{Claim~\ref*{#1}}}
\newrefformat{subsec}{\savehyperref{#1}{Subsection~\ref*{#1}}}
\newrefformat{subsubsec}{\savehyperref{#1}{Subsection~\ref*{#1}}}
\newrefformat{sec}{\savehyperref{#1}{Section~\ref*{#1}}}
\newrefformat{ineq}{\savehyperref{#1}{Inequality~\ref*{#1}}}

\newcommand\numberthis{\addtocounter{equation}{1}\tag{\theequation}}
\allowdisplaybreaks

\DeclarePairedDelimiter{\abs}{\lvert}{\rvert} 
\DeclarePairedDelimiter{\brk}{[}{]}
\DeclarePairedDelimiter{\crl}{\{}{\}}
\DeclarePairedDelimiter{\prn}{(}{)}
\DeclarePairedDelimiter{\nrm}{\|}{\|}
\DeclarePairedDelimiter{\tri}{\langle}{\rangle}

\newcommand{\ldef}{\vcentcolon=}

\def\ddefloop#1{\ifx\ddefloop#1\else\ddef{#1}\expandafter\ddefloop\fi}
\def\ddef#1{\expandafter\def\csname bb#1\endcsname{\ensuremath{\mathbb{#1}}}}
\ddefloop ABCDEFGHIJKLMNOPQRSTUVWXYZ\ddefloop
\def\ddefloop#1{\ifx\ddefloop#1\else\ddef{#1}\expandafter\ddefloop\fi}
\def\ddef#1{\expandafter\def\csname b#1\endcsname{\ensuremath{\mathbf{#1}}}}
\ddefloop ABCDEFGHIJKLMNOPQRSTUVWXYZ\ddefloop
\def\ddef#1{\expandafter\def\csname c#1\endcsname{\ensuremath{\mathcal{#1}}}}
\ddefloop ABCDEFGHIJKLMNOPQRSTUVWXYZ\ddefloop
\def\ddef#1{\expandafter\def\csname h#1\endcsname{\ensuremath{\widehat{#1}}}}
\ddefloop ABCDEFGHIJKLMNOPQRSTUVWXYZabcdefghijklmnopqrsuvwxyz\ddefloop    % Not defined for t. 
\def\ddef#1{\expandafter\def\csname hc#1\endcsname{\ensuremath{\widehat{\mathcal{#1}}}}}
\ddefloop ABCDEFGHIJKLMNOPQRSTUVWXYZ\ddefloop
\def\ddef#1{\expandafter\def\csname t#1\endcsname{\ensuremath{\widetilde{#1}}}}
\ddefloop ABCDEFGHIJKLMNOPQRSTUVWXYZ\ddefloop
\def\ddef#1{\expandafter\def\csname tc#1\endcsname{\ensuremath{\widetilde{\mathcal{#1}}}}}
\ddefloop ABCDEFGHIJKLMNOPQRSTUVWXYZ\ddefloop

\newcommand{\ball}{\mathbb{B}}

\usepackage{tikz}

\newcommand{\grad}{\nabla}

\renewcommand{\epsilon}{\varepsilon}

\usepackage{times}
\usepackage{amsmath} 
\usepackage{adjustbox}
\usepackage{tikz}
\usepackage{pgfplots}
\pgfplotsset{compat=1.15}
\usepackage{mathrsfs}
\usetikzlibrary{arrows}
\usepackage{wrapfig}

\usetikzlibrary{shapes.geometric, arrows.meta, positioning, bending}

\begin{document}
\title{Optimization, Isoperimetric Inequalities, and Sampling via Lyapunov Potentials}
\author{%
August Y. Chen$^{\dag}$ \qquad Karthik Sridharan$^{\dag}$
\vspace{10pt} 
\\
\small{$^\dag$Cornell University} 
}

\vskip 0.3in

\maketitle
\begin{abstract}%
In this paper, we prove that optimizability of any function $F$ using Gradient Flow from all initializations implies a Poincar\'e Inequality for Gibbs measures $\mu_{\beta}\propto e^{-\beta F}$ at low temperature. 
In particular, under mild regularity assumptions on the convergence rate of Gradient Flow, we establish that $\mu_{\beta}$ satisfies a Poincar\'e Inequality with constant $O\prn*{\CPILOCAL+\frac1{\beta}}$ for $\beta \ge \Omega(d)$, where $\CPILOCAL$ is the Poincar\'e constant of $\mu_{\beta}$ restricted to a neighborhood of the global minimizers of $F$. 
Under an additional mild condition on $F$, we show that $\mu_{\beta}$ satisfies a Log-Sobolev Inequality with constant $O\prn*{\beta \max\crl*{S, 1} \max\crl*{\CPILOCAL, 1}}$ where $S$ denotes the second moment of $\mu_{\beta}$.
Here asymptotic notation hides $F$-dependent parameters.
At a high level, this establishes that optimizability via Gradient Flow from every initialization implies a Poincar\'e and Log-Sobolev Inequality for the low-temperature Gibbs measure, which in turn imply sampling from all initializations. 

Analogously, we establish that under the same assumptions, if $F$ can be initialized from everywhere except some set $\cS$, then $\mu_{\beta}$ satisfies a Weak Poincar\'e Inequality with parameters $\prn*{O\prn*{\CPILOCAL+\frac1{\beta}}, O\prn*{\mu_{\beta}(\cS)}}$ for $\beta \ge \Omega(d)$.
At a high level, this shows while optimizability from `most' initializations implies a Weak Poincar\'e Inequality, which in turn implies sampling from suitable warm starts.
Our regularity assumptions are mild and as a consequence, we show we can efficiently sample from several new natural and interesting classes of non-log-concave densities, an important setting with relatively few examples. 
As another corollary, we obtain efficient discrete-time sampling results for log-concave measures satisfying milder regularity conditions than smoothness, similar to \citet{lehec2023langevin}. 
\blfootnote{ $^\star$Alphabetical ordering. \null\;\; 
{\scriptsize~~\texttt{Emails:}~\{ayc74, ks999\}@cornell.edu}} 
\end{abstract}

\tableofcontents 

\section{Introduction}

Sampling from a high-dimensional distribution is a fundamental algorithmic problem in Machine Learning (ML) and statistics, with several applications such as Bayesian inference \citep{gilks1995markov, gamerman2006markov, stuart2010inverse, kroese2013handbook, chewi2024log}. 
Moreover, with the recent rise of generative AI methods such as diffusion models, this perspective on ML has become increasingly popular in practice; see e.g. \citet{song2019generative, ho2020denoising, song2020score, song2020denoising}. 
Recently, significant theoretical progress has been made in sampling from `nice enough' -- but still fairly general -- distributions in $\mathbb{R}^d$ via the gradient-based Langevin Monte Carlo (LMC) method, which can be viewed as a natural variant of Gradient Flow (GF) and Gradient Descent (GD). 
It has recently been shown LMC can sample from the Gibbs measure $\mu_{\beta} = e^{-\beta F}/Z$, where $Z$ denotes the partition function, $F$ denotes the log-density or the \textit{energy function}, and $\beta>0$ is the inverse temperature, given access to a gradient oracle $\grad F$\footnote{Similar but weaker guarantees hold given access to a stochastic gradient oracle, which is not the focus of our work.}, if $\mu_{\beta}$ satisfies certain nice properties.\footnote{As with the rest of the literature on this topic, for the rest of the paper we assume the existence of $\mu_{\beta}$ for all $\beta \ge \Omega(1)$. Moreover, for the rest of the paper, we work in $\mathbb{R}^d$.} 

In continuous time, LMC is the \textit{Langevin Diffusion}, the following Stochastic Differential Equation (SDE):
\[ \DERIV \vecW(t) = -\beta\grad F(\vecW(t))\DERIV t+\sqrt{2}\DERIV \vecB(t)\numberthis\label{eq:LangevinSDE}.\]
Here $\vecB(t)$ denotes a standard Brownian motion in $\mathbb{R}^d$. This is a natural method to sample from $\mu_{\beta}$: the continuous-time Langevin Diffusion with inverse temperature $\beta$, the SDE \pref{eq:LangevinSDE}, converges to $\mu_{\beta}$ \citep{chiang1987diffusion}. 
In discrete time, there are several discretizations of \pref{eq:LangevinSDE}. One natural discretization is \textit{Gradient Langevin Dynamics}, defined as follows:
\[ \vecW_{t+1} \leftarrow \vecW_t - \eta \beta \nabla F(\vecW_t)+\sqrt{2\eta}\vecEps_t\numberthis\label{eq:SGLDiterates}.\]
Here $\eta>0$ is the step size, $\vecEps_t \sim \mathcal{N}(0,\matI_d)$ is a $d$-dimensional standard Gaussian, and $\beta>0$ is the \textit{inverse temperature parameter} (when larger, noise is weighted less). Another is the \textit{Proximal Sampler} which we elaborate on in \pref{subsec:proximalsampler} \citep{lee2021structured, chen2022improved, liang2022proximal1, liang2022proximal2, fan2023improved, altschuler2024faster}. Yet another discrete-time sampler is the \textit{Weakly Dissipative Tamed Unadjusted Langevin Algorithm} and the \textit{Regularized Tamed Unadjusted Langevin Algorithm}, which we elaborate on in \pref{subsec:tamedsampler} \citep{lytras2024tamed}. Broadly, these algorithms are known as \textit{Langevin Dynamics} and aim to discrete \pref{eq:LangevinSDE}. Note as $\beta\rightarrow\infty$, reparametrizing \pref{eq:LangevinSDE} in terms of $t_{\text{new}}=\beta t$, \pref{eq:LangevinSDE} becomes GF with time $t_{\text{new}}$, and reparametrizing \pref{eq:SGLDiterates} in terms of $\eta_{\text{new}}=\eta \beta$, \pref{eq:SGLDiterates} becomes GD with step size $\eta_{\text{new}}$.

It is now established that continuous and discrete time LMC can sample from $\mu_{\beta}$ \textit{beyond log-concavity} (when $F$ is convex), to when $\mu_{\beta}$ satisfies an \textit{isoperimetric inequality}, which correspond to geometric properties of $F$ allowing the Markov process \pref{eq:LangevinSDE} to mix efficiently \citep{villani2009optimal, villani2021topics, bakry2014analysis}. 
\begin{itemize}
    \item The most general such inequality under which discrete-time LMC has been proved to be successful from \textit{arbitrary} initialization is when $\mu_{\beta}$ satisfies a \textit{Poincar\'e Inequality} (PI) \citep{chewi21analysis}. 
    \item A stronger, related inequality under which discrete-time LMC efficiently succeeds is when $\mu_{\beta}$ satisfies a \textit{Log-Sobolev Inequality} (LSI) \citep{vempala2019rapid}. This is referred to as the `sampling analogue of gradient domination', as it implies gradient domination in Wasserstein space \citep{jordan1998variational}. 
    \item Under a \textit{Weak Poincar\'e Inequality} (WPI), which generalizes a PI, continuous time LMC can efficiently sample from $\mu_{\beta}$ from a suitable \textit{warm start} \citep{rockner2001weak, wang2006functional, bakry2014analysis, mousavi2023towards, huang2024weak}. 
\end{itemize}
We defer more discussion on isoperimetric inequalities to \pref{subsec:isoperimetryineq}. Such sampling results have in turn been used to show appropriately-scaled LMC can optimize non-convex $F$ to tolerance $\tilde{O}\prn*{\frac{d}{\beta}}$ \citep{raginsky2017non, xu2018global, zou2021faster}.\footnote{In runtime worst-case exponential in $\beta$.}

However, it is not clear what this means more concretely. Classically, when $F$ is convex, $\mu_{\beta}$ satisfies a PI \citep{bobkov1999isoperimetric}; when $F$ is strongly convex, $\mu_{\beta}$ satisfies a LSI \citep{bakry2006diffusions}. But beyond convexity, do we have classes of energy functions/log-densities $F$ for which $\mu_{\beta}$ satisfies isoperimetry? For example, when $F$ satisfies gradient domination in the traditional sense of optimization -- which allows for GF and GD to optimize $F$ -- does $\mu_{\beta}$ satisfy a PI or LSI (and consequently we can sample from it)? 

Before highlighting our results, we mention that related works and a comparison to our results, including the concurrent works \citet{chewi2024ballistic, gong2024poincare}, can be found next in \pref{subsec:relatedworks}. 

\textbf{Convention.} For the rest of paper, by shifting we assume WLOG that $F$ attains a minimum value of $0$ on $\mathbb{R}^d$. We let $\vecW^{\star}$ denote any arbitrary global minimizer of $F$, thus $F(\vecW^{\star})=0$. 

\subsection{Overview of Results}\label{subsec:ourresultsoverview}
\paragraph{PI/LSI Results:} The similarity between Langevin Dynamics and GF/GD motivates the overarching:
\begin{conjecture}\label{conj:opttosamplingconj}
If $F$ is optimizable via Gradient Descent from arbitrary initialization, then $\mu_{\beta} := e^{-\beta F}/Z$ satisfies a PI for appropriate $\beta$. Thus we can efficiently sample from $\mu_{\beta}$ for such $\beta$ with oracle access to $\grad F$. 
\end{conjecture}
This is natural: if gradient-based methods succeed for optimization without getting stuck, LMC ought to not get stuck as well. Moreover, $\grad F$ is the exact same oracle as we have for GF/GD. 

We proceed to define optimizability of $F$ via GF following Definition 1 and Theorem 2 of \citet{priorpaper}. This following condition is shown in \citet{priorpaper} to be \textit{implied} by the existence of an appropriate rate function for the convergence of GF. The notion of appropriate rate function from \citet{priorpaper} is very generic -- for example, is is satisfied whenever GF enjoys an exponential rate -- and as such the following definition covers numerous examples in non-convex optimization. See \pref{sec:examplesapplications} for a subset of these examples.
\begin{definition}[Optimizability of $F$ via Gradient Flow]\label{def:optimizabilitydef}
For $F$ with minimum value 0, we say $F$ is optimizable by Gradient Flow if for all $\vecW\in\mathbb{R}^d$, there exists a Lyapunov Function $\Phi(\cdot)$ such that 
\[ \tri*{\grad \Phi(\vecW), \grad F(\vecW)} \ge g\prn*{F(\vecW)},\numberthis\label{eq:gfcondition}\]
where $g$ is monotonically non-decreasing with $g(0)=0$ and $g(x)>0$, $g(x) \ge m'x-b'$ for some $m',b'>0$ for all $x>0$.\footnote{We assume $g(x)$ has at least linear tail growth, as $g$ arises to handle when the rate function $R(\vecW, t)$ for GF is not integrable, e.g. for convex rate $t^{-1}$.} Moreover, we say $F$ is \textit{optimizable by GF from a set $\cQ\subset\mathbb{R}^d$} if \pref{eq:gfcondition} holds for all $\vecW\in\cQ$.
\end{definition}
\textbf{Convention.} We simply refer to $F$ as optimizable when $F$ is optimizable by GF in the sense of \pref{def:optimizabilitydef}. 

Moreover, to obtain a PI and therefore discrete-time sampling results, it is natural to assume discrete-time optimization via GD in addition to GF succeeds.
For GD to succeed in optimizing $F$ (i.e. for Taylor terms in GD to be controlled), we require that $\Phi$ and $F$ satisfy the following assumption:
\begin{assumption}[Self-Bounding Regularity]\label{ass:phiselfbounding}
For some monotonically non-decreasing $\rho_{\Phi}, \rho_F:\mathbb{R}_{\ge 0}\rightarrow\mathbb{R}_{\ge 0}$, we have $\nrm*{\grad \Phi(\vecW)}, \nrm*{\grad^2 \Phi(\vecW)}_{\OPNORM} \le \rho_{\Phi}\prn*{\Phi(\vecW)}$ and $\nrm*{\grad F(\vecW)}, \nrm*{\grad^2 F(\vecW)}_{\OPNORM} \le \rho_{F}\prn*{F(\vecW)}$.\footnote{In fact the bound on operator norm implies the bound on the gradient; see Lemma 11, \citet{priorpaper}.}
\end{assumption}
As shown in Theorem 3 of \cite{priorpaper}, assumptions analogous to \pref{ass:phiselfbounding} are actually \textit{necessary} for GD to succeed for discrete-time optimization, and hence come with little loss of generality. 
Note smoothness of $\Phi$ and $F$ (e.g. $\Phi=F$ for P\LCHAR functions) is a special case of \pref{ass:phiselfbounding}, but \pref{ass:phiselfbounding} is much more general.
Such a framework with dimension-independent $\rho_{\Phi}, \rho_F$ subsumes numerous examples in non-convex (and convex) optimization; see \pref{sec:examplesapplications} and \citet{priorpaper}.

We confirm \pref{conj:opttosamplingconj} in the following sense, stated formally in \pref{thm:generalgflinearizabletopi}. Under \pref{ass:phiselfbounding}, \pref{ass:Funimodal} (which subsumes the literature and is necessary, see \pref{rem:badlocalisoperimetry}), and \pref{ass:linearizableoutsideball}:
\[ \text{Optimizability of $F$ for all $\vecW$, i.e. \pref{eq:gfcondition}} \implies \text{ PI for $\mu_{\beta}$ for $\beta =\Omega(d)$ with PI constant $O\prn*{\textup{poly}(d,\beta)}$}.\numberthis\label{eq:informaltheorempi}\]
In \pref{thm:generalgflinearizabletopi}, we furthermore establish:
\[ \text{Above conditions + mild regularity on $F$ } \implies \text{ LSI for $\mu_{\beta}$ for $\beta =\Omega(d)$ with LSI constant $O\prn*{\textup{poly}(d,\beta)}$}.\]
In comparing optimization to sampling for $F$ optimizable by GF/GD, $\beta=\Omega(d)$ is the correct scaling (and has several applications); see \pref{subsec:roleoftemp}. When $\beta=\Omega(d)$ is written above, the asymptotic notation hides $F$-dependent constants; see e.g. \pref{rem:linearizableboostshort} and \pref{subsec:proofopttosampling} for full expressions. As a direct consequence of the literature, having established a PI and/or LSI, we obtain that discrete-time LMC can sample from $\mu_{\beta}$ for such $\beta$ in time polynomial in $d, \beta, \frac{1}{\epsilon}$ under very mild regularity assumptions; see \pref{corr:standardregularitysampling}, \pref{corr:newregularitysampling}. 

\textit{We view this as a core strength of our work: our result complements the literature and `plugs and plays' with sampling algorithms and their analysis that study sampling under isoperimetry.}
We further emphasize that \textit{the focus of our work is not to develop or analyze sampling algorithms, but rather to prove that geometric properties imply functional inequalities (PI/WPI)}, which are the \textit{crux} of LMC. 
To obtain \pref{corr:standardregularitysampling}, \pref{corr:newregularitysampling} we simply take results in the literature that, to the best of our knowledge, have the state-of-the-art results for LMC.

For these corollaries we make no warm start assumption, and instead explicitly describe the initialization, which does not depend on $\vecW^{\star}$. Our sampling algorithms succeed solely because $F$ is optimizable everywhere; intuitively, LMC `moves' us towards $\mu_{\beta}$ due to the optimizability condition $\tri*{\grad \Phi(\vecW), \grad F(\vecW)} \ge g\prn*{F(\vecW)}$. If optimizability only holds within $\ball(\vecW^{\star}, R)$ for some $R$, we show in \pref{rem:localconditionsampling} (with details in \pref{prop:obtainlocalcondition}) that by appropriately regularizing on $F$ outside $\ball(\vecW^{\star}, R)$ to yield $\hat{F}$, we can sample from $\hat{\mu}_{\beta} \propto \exp(-\beta \hat{F}) \approx \mu_{\beta}$ (the approximation holds for $R$ large). We view this as an interesting algorithmic implication of our work.

\paragraph{Weak Poincar\'e Inequalities:} In many non-convex landscapes, such as Phase Retrieval and Matrix Square Root, there is a set $\cS$ with small Lebesgue measure of bad initializations where GF/GD does not succeed, but everywhere else GF/GD works \citep{jain2017global, lee2019first, priorpaper}. It can be moreover verified that outside $\cS$, optimizability as per \pref{def:optimizabilitydef} holds \citep{priorpaper}. Little is known about sampling in such settings. As such a deeper understanding of these settings is very important and interesting.

A Weak Poincar\'e Inequality (WPI) captures this picture, corresponding to efficient sampling under a \textit{warm start} which has low density in $\cS$. It is crucial to note such a situation is \textit{not covered by a PI}, as a PI implies \textit{worst-case} mixing. Thus it is natural to expect:
\begin{conjecture}\label{conj:opttosamplingconjwpi}
If $F$ is optimizable via Gradient Descent from everywhere except a set $\cS$ with small Lebesgue measure, then $\mu_{\beta}$ satisfies a $(\CWPI, \delta)$-WPI with $\delta$ small for appropriate $\beta$. (See \pref{subsec:isoperimetryineq} for the formal definition of a WPI; here $\delta$ in the WPI controls the `error' we can sample to efficiently.) Thus we can efficiently sample from $\mu_{\beta}$ for such $\beta$ with oracle access to $\grad F$ with a \textit{warm start}. 
\end{conjecture}
For clarity on what we mean by $F$ being optimizable via Gradient Descent from everywhere except a set $\cS$ with small Lebesgue measure, we mean that for all $\vecW \in \mathbb{R}^d \setminus \cS$, \pref{eq:gfcondition} holds. That is, we have for some $\Phi$ and $g$ satisfying the conditions of \pref{def:optimizabilitydef},
\[ \tri*{\grad \Phi(\vecW), \grad F(\vecW)} \ge g(F(\vecW)) \text{ for all }\vecW \in \mathbb{R}^d \setminus \cS.\]
We denote this by `optimizability of $F$ from $\cS^c$'. As a concrete example, this holds if $F$ is P\L{ }outside of some $\cS \subset \mathbb{R}^d$.

Indeed, we confirm \pref{conj:opttosamplingconjwpi} in the following sense, stated formally in \pref{thm:generalgftowpi}. We show under \pref{ass:phiselfbounding}, \pref{ass:Funimodal}, \pref{ass:linearizableoutsideball} that
\[ \text{Optimizability of $F$ from $\cS^c$} \implies \text{ $(\CWPI, O(\mu_{\beta}(\cS)) )$-WPI for $\mu_{\beta}$, $\beta =\Omega(d)$, $\CWPI \approx \CPI$ from \pref{eq:informaltheorempi}}.\numberthis\label{eq:informaltheoremwpi}\]
Thus if $\mu_{\beta}(\cS)$ is small (e.g. if $\cS$ has small Lebesgue measure and $\inf_{\vecW \in \cS}F(\vecW)$ is not too small), the above shows we can sample to low error via LMC from a warm start. Again here, the $O(\cdot)$, $\Omega(\cdot)$ hide $F$-dependent parameters. With a WPI, sampling from a warm start follows via e.g. \citet{rockner2001weak, mousavi2023towards, huang2024weak}. Note $\cS$ is arbitrary; it can comprise of saddle points or even spurious local minima.

\paragraph{Applications and Significance:}
Our results \pref{thm:generalgflinearizabletopi}, \pref{thm:generalgftowpi} yield a natural, novel host of non-log-concave measures $\mu_{\beta}\propto \exp\{-\beta F(\vecW)\}$ where LMC samples in time $\textup{poly}(d,\beta, 1/\epsilon)$. The crux to establishing these polynomial guarantees is showing $\textup{poly}(d,\beta)$ bounds on isoperimetric constants. This is known for log-concave measures (convex $F$); beyond log-concavity, known methods (e.g. perturbation criteria) give $\exp(d)$ constants, even for e.g. P\LCHAR $F$. 
By contrast, \pref{def:optimizabilitydef}, and hence our results, subsumes the following general non-convex function classes for which GF/GD succeed for global optimization: Polyak-\L ojasiewicz (P\L) \citep{polyak1963gradient, lojasiewicz1963topological}, Kurdyka-\L ojasiewicz (K\L) \citep{kurdyka1998gradients}, and Linearizable \citep{sekhari2021sgd} functions (also known as Quasar-Convexity \citep{hinder2020near}). 
\begin{definition}[Polyak-\L ojasiewicz (P\L)]\label{def:PL}
A differentiable function $F$ is Polyak-\L ojasiewicz (P\L) with parameter $\lambda>0$ if $\nrm*{\grad F(\vecW)}^2 \ge \lambda F(\vecW)$ for all $\vecW\in\mathbb{R}^d$. (Take $\Phi=F$, $g(x)=\lambda x$ in \pref{def:optimizabilitydef}. Recall we shifted so $F$ has minimum value 0 before this section.)
\end{definition}
\begin{definition}[Kurdyka-\L ojasiewicz (K\L)]\label{def:KL}
A differentiable function $F$ is Kurdyka-\L ojasiewicz (K\L) with parameter $\lambda>0$, $\theta \in [0,1)$ if $\nrm*{\grad F(\vecW)}^2 \ge \lambda F(\vecW)^{1+\theta}$ for all $\vecW\in\mathbb{R}^d$. (Take $\Phi=F$, $g(x)=\lambda x^{1+\theta}$.)
\end{definition}
\begin{definition}[Linearizable]\label{def:Linearizable}
A differentiable function $F$ is $\lambda$-linearizable if for some global minimizer $\vecW^{\star}\in\mathbb{R}^d$ of $F$, $\tri*{\grad F(\vecW), \vecW-\vecW^{\star}} \ge \lambda F(\vecW)$ for all $\vecW\in\mathbb{R}^d$.\footnote{We make a change of variables compared to its definition in \citep{sekhari2021sgd}.} (Take $\Phi=\nrm*{\vecW-\vecW^{\star}}^2$, $g(x)=\lambda x$.)
\end{definition}
Consequently \pref{thm:generalgflinearizabletopi} yields a PI and thus $\textup{poly}(d,\beta,1/\epsilon)$ sampling guarantees for $\mu_{\beta} \propto \exp(-\beta F)$ when $F$ is in the above classes, under the conditions of \pref{thm:generalgflinearizabletopi}. Analogously, under the conditions of \pref{thm:generalgftowpi}, we obtain a WPI and sampling from a warm start for $\mu_{\beta} \propto \exp(-\beta F)$ when $F$ is in the above classes. Such a result has further applications and interpretation in Bayesian inference, as we detail in \pref{subsec:roleoftemp}. This alternate interpretation is: if Maximum a posteriori is optimizable by GF/GD, LMC efficiently samples from the posterior. Optimizability of ERM/regularized ERM by GF/GD has been studied extensively; our work lets us systematically use such results to fuel Bayesian inference.

For another application, note general convex functions are 1-Linearizable and automatically satisfy \pref{ass:Funimodal}. \pref{corr:generalexamplescorollary} thus gives \pref{corr:convexbeyondsmooth}, a $\textup{poly}\prn*{\beta,d,\frac1{\epsilon}}$ sampling guarantee for log-concave measures at low temperatures under relaxed regularity assumptions (beyond smoothness). Log-concave sampling beyond smoothness was studied in \citet{lehec2023langevin}; our regularity assumptions are in some sense more general.

\paragraph{Technical Approach:} We also highlight our technical approach. We utilize this exact Lyapunov function $\Phi$ from optimization (from \pref{def:optimizabilitydef}) to execute the Lyapunov potential technique from probability \citep{bakry2008simple} to prove a PI/LSI. Generally the technique of \citet{bakry2008simple} involves significantly different Lyapunov potentials than those from optimization, and often ad-hoc. Using the exact same potential from optimization gives crisp quantitative control over the isoperimetric constants of $\mu_{\beta}$. This crisp quantitative control stands in contrast to typical usages of this technique. We also further develop this technique to prove a WPI. To the best of our knowledge, our work is the first to develop the Lyapunov function technique to establish a WPI. Our means of using the Lyapunov function technique to establish a WPI is simple and user-friendly, and we expect that it will have further applications. \textit{As such, our work tightens the link between optimization, sampling, and probability in several ways.}

\paragraph{Connecting Optimization and Sampling:} Our results yield \textit{fundamental relationships at the algorithmic level, connecting optimizability via GF/GD to isoperimetry at low temperature (and hence the success of Langevin Dynamics).} There are several connections between sampling and optimization, from the Proximal Point Method of optimization inspiring the Proximal Sampler, to interior point methods for log-concave sampling \citep{kook2024gaussian}. Here, we address \pref{conj:opttosamplingconj}, \pref{conj:opttosamplingconjwpi} and deepen the connection between optimization, isoperimetry, and sampling from another angle.

\subsection{Related Works}\label{subsec:relatedworks}
Several other works have studied the connection between efficient optimization, isoperimetry, and sampling:
\begin{itemize}
\item \citet{ma2019sampling} studied this connection across \textit{different} temperature levels $\beta$, where the behavior of $\mu_{\beta}$ fundamentally changes (see e.g. Corollary 1 therein). In contrast, we study a given, fixed landscape for large $\beta$, and study the connection between optimization and sampling in this landscape.
\item Several recent works \citep{li2023riemannian, kinoshita2022improved, lytras2023taming, huang2023strong, sellke2024threshold} show that when the landscape of $-\log \mu_{\beta} = F$ is strict saddle in the sense of a constant order negative eigenvalue around spurious critical points, then combined with several other regularity assumptions, functional inequalities hold. Among these, \citet{kinoshita2022improved, lytras2023taming} studies the problem in Euclidean spaces. However, this does not capture our setting of general functions optimizable by GF/GD. Thus these settings are not comparable. Indeed, there are many functions where GF/GD succeed that are not strict-saddle, such as star-convex functions, smooth one-point-strongly convex functions, and even general convex functions. See \pref{ex:formalLinearizableexample} for further discussion of these examples. These results also contain many unnecessary regularity assumptions and/or suboptimal $F$-dependent parameters. We bypass these suboptimal dependencies via our novel use of the Lyapunov function method.

Moreover, the results of \citet{kinoshita2022improved, lytras2023taming, li2023riemannian} only hold for an unreasonably low temperature regime, $\beta \ge \Omega(d^6)$, where $\Omega(\cdot)$ again hides $F$-dependent parameters. This is often much larger than $\beta=\tilde{\Theta}\prn*{\frac{d}{\epsilon}}$ used for optimization via LMC to tolerance $\epsilon$. At such the algorithmic implications of their result simply is that optimization is possible in strict-saddle landscapes. By contrast, this is \textit{not} the case for $\beta \ge \Omega(d)$ as we consider, the regime where sampling is of interest in Bayesian inference; see \pref{subsec:roleoftemp}. 

\item The concurrent works \citet{chewi2024ballistic, gong2024poincare} study a special case of our problem, when $F$ is P\LCHAR and $\beta$ is large (a setting subsumed by our \pref{thm:generalgflinearizabletopi}), also proceeding through Lyapunov functions. 

\citet{gong2024poincare} studies this problem under a local P\LCHAR condition around local minima. However, they place several regularity assumptions on all of $\mathbb{R}^d$, which in they show in their Proposition 3.1 in fact imply unimodality analogous to our setting. Their Proposition 3.1 implies the existence of a connected set of local minima (see their note on page 3) and no saddle points. They further require a strictly negative lower bound on the Laplacian $\Delta F$ when the gradient is small, which factors into their quantitative dependencies; furthermore, such a situation can handled by our exact same proof, see \pref{sec:proofsketch}. Thus their work reduces to a setting analogous to ours. Their bound on the PI constant also implicitly incurs exponential $d$ dependence; it contains a term of the form $\exp(\overline{C})$ (their Theorem 2, Lemma 4), and $\overline{C} = \Omega(M_{\Delta}) = \Omega(d)$ for generic smooth $F$ (Lemmas 2, 3).\footnote{Note they adopt convention that smaller PI constant is worse.}

\citet{chewi2024ballistic} obtains a sharp characterization of the Poincar\'e and Log-Sobolev constants of $\mu_{\beta}$ when $F$ is P\LCHAR and has a unique minimizer $\vecW^{\star}$ in the \textit{asymptotic} limit $\beta\rightarrow\infty$. In this asymptotic limit, sampling degenerates into optimization and consequently the algorithmic implications of their result is relatively limited.\footnote{We point out their result will only hold for $\beta\ge\Omega(d)$ where asymptotic notation hides $F$-dependent parameters, since they require an upper bound on the Laplacian of $F$, which scales with $d$ even for e.g. quadratics.} We also remark that our \pref{thm:generalgflinearizabletopi} implies their upper bound on the Poincar\'e constant up to a universal constant factor of 2 (see e.g. \pref{rem:linearizableboostshort}), and that a Poincar\'e Inequality is sufficient to give an efficient sampling algorithm (see for instance \citet{chewi21analysis, lytras2024tamed}).

By contrast, our general optimizability condition is far more comprehensive and allows us to capture many examples under a single umbrella. It captures not only P\LCHAR but also K\L, Linearizable, Star-Convex, One-Point-Convex and general convex functions (see \pref{ex:formalKLexample}, \pref{ex:formalLinearizableexample}). As an extreme example, convex $F$ need not be P\L, but are readily subsumed by our setting (see \pref{ex:formalLinearizableexample}). 

Our method of using Lyapunov functions is also novel, in that we prove functional inequalities using the \textit{same} Lyapunov function arising from optimization, further highlighting the connection between optimization and sampling. Our techniques also yield improved quantitative dependencies on $F$-dependent parameters; see \pref{rem:ourimprovementonlyapunovfull}. As a consequence of our general optimizability condition, beyond a wide host of applications (\pref{ex:formalPLexample}, \pref{ex:formalKLexample}, \pref{ex:formalLinearizableexample}), \textit{we obtain fundamental relationships at the algorithmic level: that optimizability, at appropriate $\beta$, implies the success of Langevin Dynamics for sampling.}
\end{itemize}
Furthermore, none of these works connect optimizability outside of some unfavorable region $\cS$ (as is often the case in non-convex landscapes, e.g. Phase Retrieval) to a WPI, as we do in \pref{thm:generalgftowpi}. \citet{gong2024poincare} allows for local maxima outside a local region (which as remarked above can be readily handled by our proof), but do not permit saddle points or spurious local minima as we do in \pref{thm:generalgftowpi}. We also present algorithm implications of our result via regularization if we only have `local' optimizability in \pref{prop:obtainlocalcondition} but arbitrary stationary points/spurious local minima elsewhere, a perspective unexplored in these works.

\section{Preliminaries and Technical Background} 
\subsection{Isoperimetric Inequalities}\label{subsec:isoperimetryineq} 
\textit{Isoperimetric inequalities} define geometric properties of $F$ that enable LMC (or other Markov Chains, which we do not expand on here) to mix rapidly. These isoperimetric inequalities are governed by their \textit{isoperimetric constant}; in this work we adopt the notion that a smaller isoperimetric constant implies a stronger inequality.
From \textit{arbitrary} initializations, the most general condition under which LMC is successful is when $\mu_{\beta}$ satisfies a \textit{Poincar\'e Inequality} (PI) \citep{villani2021topics, bakry2014analysis}, defined as follows:
\begin{definition}[Poincar\'e  Inequality (PI)]
A probability measure $\mu$ on $\mathbb{R}^d$ satisfies a Poincar\'e Inequality (PI) with constant $\CPI(\mu)$ if for all infinitely differentiable functions $f:\mathbb{R}^d\rightarrow \mathbb{R}$, we have 
\[ \int_{\mathbb{R}^d} f^2 \DERIV\mu - \prn*{\int_{\mathbb{R}^d} f \DERIV\mu}^2 \le \CPI(\mu) \int_{\mathbb{R}^d} \nrm*{\grad f}^2 \DERIV\mu.\]
If the above is not satisfied, following the convention, we set $\CPI(\mu)=\infty$. 
\end{definition}
A PI corresponds to exponential contraction of variance for the Langevin Diffusion \pref{eq:LangevinSDE} (note the left hand side can be written as the variance $\mathbb{V}_{\mu}\prn*{f}$), and directly implies continuous-time sampling results in $\chi^2$-divergence via Langevin Dynamics \pref{eq:LangevinSDE}. In particular, letting $\pi_T$ denote the probability measure obtained after running the Langevin Diffusion \pref{eq:LangevinSDE} (with $-\log \mu$ in place of $\beta \grad F$) for time $T$ and $\pi_0$ denote the initialization, we have
\[ \chi^2\prn*{\pi_T||\mu} \le e^{-2T/\CPI(\mu)} \chi^2\prn*{\pi_0||\mu}.\]
For both of these results, see e.g. Chapter 4, \citet{bakry2014analysis}. By \citet{bobkov1999isoperimetric}, if $\mu$ is \textit{log-concave}, or equivalently $-\log \mu$ is a convex function of $\vecW$, then $\mu_{\beta}$ satisfies a PI.
We next define \textit{Log-Sobolev Inequality (LSI)}, which is stronger than PI.
\begin{definition}[Log-Sobolev Inequality (LSI)] 
A probability measure $\mu$ on $\mathbb{R}^d$ satisfies a Log-Sobolev Inequality (LSI) with Log-Sobolev constant $\CLSI(\mu)$ if for all infinitely differentiable functions $f:\mathbb{R}^d\rightarrow \mathbb{R}$, we have 
\[ \int_{\mathbb{R}^d} f \ln f \DERIV\mu - \int_{\mathbb{R}^d} f \ln \prn*{\int_{\mathbb{R}^d} f \DERIV \mu}\DERIV\mu \le 2\CLSI(\mu) \int_{\mathbb{R}^d} \nrm*{\grad f}^2 \DERIV\mu.\]
If the above is not satisfied, following the convention, we set $\CLSI(\mu)=\infty$.
\end{definition}
A LSI has been referred to as the `sampling analogue of the P\L\text{ }Inequality', since it implies gradient domination in Wasserstein space \citep{chewi2024log}. A LSI corresponds to exponential contraction of entropy $\text{ent}_{\mu}(f)$ for the Langevin Diffusion \pref{eq:LangevinSDE}, which again is the left hand side of the above, and directly implies exponential contraction for the $\KL$-divergence via the Langevin Diffusion \pref{eq:LangevinSDE} (run with $-\log \mu$ in place of $\beta \grad F$). Namely defining $\pi_T$, $\pi_0$ as earlier, a LSI implies
\[ \KL\prn*{\pi_T||\mu} \le e^{-2T/\CLSI(\mu)} \KL\prn*{\pi_0||\mu}.\]
See e.g. Chapter 5, \citet{bakry2014analysis}. A LSI is stronger than a PI with the same constant: a LSI with constant $\CLSI(\mu)$ implies that a PI with the same constant holds, thus $\CPI(\mu) \le \CLSI(\mu)$, but the reverse implication does not hold \citep{chewi2024log}. Obtaining a sampling result in $\KL$ (via LSI) is also stronger than in $\chi^2$ (via PI). Indeed, not all log-concave measures satisfy a LSI.

From a suitable \textit{warm-start}, the Langevin Diffusion can efficiently sample from $\mu_{\beta}$ under a \textit{Weak Poincar\'e Inequality} (WPI) \citep{rockner2001weak, wang2006functional, bakry2014analysis, mousavi2023towards, huang2024weak}, which captures \textit{beyond worst-case mixing}. Consider e.g. a mixture of two well-separated identity covariance Gaussians: mixing from arbitrary initialization is exponentially slow in $d$, but starting from a normal perfectly centered between the modes, we could conceivably obtain rapid mixing. Indeed, several works in probability have studied sampling from complicated distributions satisfying a WPI by `chaining together' warm starts \citep{alaoui2023fast, huang2024weak}. 
To define a WPI, we adopt convention from Definition 4.7, \citet{huang2024weak}.\footnote{The definition above in fact implies Definition 4.7 of \citet{huang2024weak}.} 
\begin{definition}[Weak Poincar\'e Inequality (WPI)]\label{def:wpidef}
A probability measure $\mu$ on $\mathbb{R}^d$ satisfies a $(\CWPI(\mu),\delta)$-Weak Poincar\'e Inequality (WPI) if for all infinitely differentiable functions $f:\mathbb{R}^d\rightarrow \mathbb{R}$, letting $\text{osc}(f)=\sup f - \inf f$, we have 
\[ \int_{\mathbb{R}^d} f^2 \DERIV\mu - \prn*{\int_{\mathbb{R}^d} f \DERIV\mu}^2 \le \CWPI(\mu) \int_{\mathbb{R}^d} \nrm*{\grad f}^2 \DERIV\mu + \delta \text{osc}(f)^2.\]
\end{definition}
Note $\text{osc}(f) \le 2 \sup(\abs*{f-\mathbb{E}\brk*{f}})$, so applying Theorem 2.1 of \citet{rockner2001weak} as in (2) of \citet{huang2024weak} and defining $\pi_T, \pi_0$ as earlier, we have the following mixing guarantee for the continuous-time Langevin Diffusion \pref{eq:LangevinSDE} (again, run with $-\log \mu$ in place of $\beta \grad F$)):
\[ \chi^2(\pi_T||\mu) \le e^{-T/\CWPI(\mu)} \chi^2(\pi_0||\mu) + 4\delta \nrm*{\tfrac{\DERIV \pi_0}{\DERIV \mu}-1}^2_{\infty}. \numberthis\label{eq:wpiguarantee}\]
Thus if $\pi_0$ is a suitable warm start in that $\nrm*{\tfrac{\DERIV \pi_0}{\DERIV \mu}-1}^2_{\infty}$ is small, then we obtain a mixing guarantee. Hence $\delta$ is the `error' or `slack' in the WPI, indicating how accurately we can sample efficiently with a warm start. Thus in \pref{thm:generalgftowpi}, if $\mu_{\beta}(\cS)$ is small, we can sample efficiently in continuous-time to small accuracy.
 
It is also worth discussing the tail growth of $F$ for which $\mu_{\beta} = e^{-\beta F}/Z$ satisfies an isoperimetric inequality \citep{chewi21analysis, mousavi2023towards}. A PI for $\mu_{\beta}$ goes hand-in-hand with $F$ having at least \textit{linear} tail growth (e.g. $F(\vecW)=\nrm*{\vecW}$). For example, we can prove $F$ has linear tail growth if $F$ is convex and $\mu_{\beta}$ exists; see Lemma 2.2, \citet{bakry2008simple}.
A LSI for $\mu_{\beta}$ goes hand-in-hand with $F$ having at least \textit{quadratic} tail growth (e.g. $F(\vecW)=\nrm*{\vecW}^2$). 
As such, it is natural to assume that $F$ has linear tail growth to prove a PI, and that $F$ has quadratic tail growth to prove a LSI.

\subsection{The Role of Temperature and Applications of Low-Temperature Sampling}\label{subsec:roleoftemp}
Notice in our earlier results that the inverse temperature $\beta = \Omega(d)$. Justification for this scaling to study the connection between optimization and sampling is severalfold:
\begin{itemize}
    \item Optimization is fundamentally performed at low temperature. Even at the initialization of optimization algorithms, the value of $F$ at initialization is often viewed as $O(1)$ in the literature \citep{priorpaper, bubeck2015convex, nesterov2018lectures}, which corresponds to the inverse temperature $\beta=\Omega(d)$; consider initializing at $\cN(\vecOrigin, \frac1{\beta}\matI_d)$. Furthermore the temperatures range we consider corresponds to \textit{initialization} of optimization where $\beta=\Omega(d)$, rather than \textit{output} of optimization to tolerance $\epsilon$ where $\beta=\Omega(d/\epsilon)$.
    \item We use $\beta=\Omega(d)$ simply to follow the above aforementioned scaling from optimization. It is possible to obtain an analogous result to ours in the $\beta=O(1)$ setting by changing \pref{ass:Funimodal} so that $\text{diam}(\cW^{\star}), r(l_b) = \Theta(\sqrt{d})$ rather than $\Theta(1)$ and $l_b = \Omega(\sqrt{d})$. Such a scaling is made for instance in \citet{huang2023strong}. Then one can simply follow the same proof as ours from \pref{sec:proofsketch}. 
\end{itemize}
Sampling at low temperature is also of independent theoretical interest and has been studied in several works, discussed in \pref{subsec:relatedworks}. Typically one expects that without convexity, as $\beta$ increases, the isoperimetric constants of $\mu_{\beta}$ become larger, or isoperimetric inequalities break altogether. This behavior has been rigorously confirmed in measures from statistical physics \citep{el2024bounds}. As we establish here, such behavior sharply contrasts to when $F$ is optimizable, despite the lack of global convexity. 

We remark our results or similar ones need not hold for small $\beta$. 
In \citet{bandeira2023free}, such a construction is provided, where $F$ is locally log-concave and `unimodal', but the corresponding $\mu_{\beta}\propto \exp(-\beta F)$ does not satisfy isoperimetry for particular $\beta$. 
In our result where $\beta$ is large enough and under our conditions on $F$ which hold for many examples of interest, the construction from \citet{bandeira2023free} is ruled out: the intermediate `bottleneck' region $\cW$ therein must contain comparable mass under $\mu_{\beta}(\cdot)$ as the initialization region $\cS$.
Note under the generality we work with here, finding the exact temperature threshold for $\beta$ where a PI/LSI holds is likely extremely difficult.\footnote{For example, such a threshold remains open for the Sherrington-Kirkpatrick model, for a PI with respect to the Glauber Dynamics.}

\paragraph{Applications of Low-Temperature Sampling:} Consider posterior sampling in high-dimensional regression. Following 
\citet{montanari2023posterior} (who use $p$ rather than $d$), we observe covariates $\matX \in \mathbb{R}^{n \times d}$ and response $\vecy_0\in\mathbb{R}^n$, under the linear model $\vecy_0 = \matX \vecTheta+\vecEps$, $\vecEps \sim \cN(\vecOrigin,\sigma^2 \matI_n)$. We consider the proportional asymptotics $n/d \rightarrow \delta \in(0,\infty)$, common in high-dimensional statistics (see e.g. \citet{barbier2018optimal, wainwright2019high}). Here, $\hat{F}_{\text{erm}}(\vecTheta)=\frac1n  \nrm*{\vecy_0-\matX \vecTheta}_2^2$ is empirical risk approximating population loss. Write the product measure $\pi_{\Theta}^{\otimes d}(\DERIV \vecTheta) \propto \exp(-R(\vecTheta))$. By eq. 12, p. 6 of \citet{montanari2023posterior}, the desired posterior to sample from is $\mu_{\matX, \vecy_0}(\DERIV \vecTheta) \propto \exp\prn*{-\frac1{2\sigma^2} \nrm*{\vecy_0-\matX \vecTheta}_2^2} \pi_{\Theta}^{\otimes d}(\DERIV \vecTheta)$. Thus as $n/d \rightarrow \delta$, 
\[ \mu_{\matX, \vecy_0}(\DERIV \vecTheta) \propto\exp\crl*{ -\frac{d \delta}{2\sigma^2} \prn*{\hat{F}_{\text{erm}}(\vecTheta)+\frac{2\sigma^2}{n} R(\vecTheta)}}.\]
This posterior is in the low temperature regime we consider, i.e. $\beta=d \cdot \frac{\delta}{2\sigma^2}=\Omega(d)$. The objective we consider is the regularized ERM objective.
Under the same asymptotics, the posterior is at low temperature $\beta=\Omega(d)$ more generally: for a Gibbs prior $\propto\exp(-R(\vecTheta))$ for $\vecTheta$ and the model $(\vecy_0)_i=f(\vecx_i;\vecTheta)+\vecEps_i$, $\vecEps_i \sim \cN(0, \sigma^2)$ for $1 \le i \le n$. For example, GLMs, where $f(\vecx_i;\vecTheta)=\psi(\vecx_i^{\top} \vecTheta)$, are non-convex but are optimizable by GF/GD under appropriate conditions on $\psi(\cdot)$ \citep{foster2018uniform, wang2023continuized}. Note our results apply when $\beta/d = \Theta(1)$ as $d\rightarrow \infty$. In the regime $\beta/d \rightarrow \infty$ as $d\rightarrow \infty$, as is the case for several related works mentioned in \pref{subsec:relatedworks}, posterior sampling is significantly easier \citep{montanari2023posterior, bontemps2011bernstein}.

One can interpret our result from a Bayesian inference perspective: if Maximum a posteriori is optimizable by GF/GD, LMC efficiently samples from the posterior. Optimizability of ERM and regularized ERM by GF/GD has been studied extensively; our work lets us systematically use such results to fuel Bayesian inference. 

We also note sampling at at low temperature is important in generative AI; the end of the reverse process in diffusion models is at low temperature. See e.g. \citet{song2019generative, song2020denoising, song2020score, yangsongblog}. Moreover, in the temperature restriction $\beta=\Omega(d)$, we can replace $d$ by the rank of $\grad^2 \Phi$ when $\Phi$ is smooth.

\section{Connecting Optimizability and Sampling}\label{sec:mainresults}
Before we state our results, we state the following unimodality assumption on $F$. Functional inequalities generally do not hold without exponential dimension-dependence when $F$ has well-separated modes \citep{bovier2004metastability, bovier2005metastability, menz2014poincare}. This is a probabilistic analogue to standard assumptions in non-convex optimization of good local behavior, such as $F$ being convex or P\L/K\L{ }near the global minima or near all saddle points, in e.g. \citet{damian2021label, ahnescape}.
\begin{assumption}\label{ass:Funimodal}
Let $\cW^{\star}$ denote the set of global minima. For all small enough $l>0$, there exists $r(l)>0$ such that $\{F \le l\} \subset \ball(\cW^{\star}, r(l))$ and $\mu_{\beta, \textsc{local}}(l)$, the restriction of $\mu_{\beta}$ on $\ball(\cW^{\star}, r(l))$, satisfies a Poincar\'e Inequality with constant $\CPILOCAL(l)$. Here $\ball(\cW^{\star}, r(l))=\{\vecW:d(\vecW,\cW^{\star}) \le r(l)\}$, where $d(\cdot,\cW^{\star})$ denotes the distance from $\vecW$ to the closest point in $\cW^{\star}$.
\end{assumption}
\begin{remark}
We believe the growth condition in the above is relatively unrestrictive. For example, if $F$ is P\LCHAR with parameter $\lambda$, by Theorem 2 of \cite{karimi2016linear}, $\{F \le l\} \subset \ball(\cW^{\star}, r(l))$ for $r(l) = 2\sqrt{l/\lambda}$.\footnote{See also \citet{chewi2024ballistic}, \citet{otto2000generalization}.} 
\end{remark}
Furthermore, there are several natural, general examples satisfying \pref{ass:Funimodal} subsuming standard settings of the literature with precise quantitative bounds on $\CPILOCAL(l)$. We explain fully in \pref{subsec:furtherexamples}:
\begin{example}[Local convexity]\label{ex:Fconvexnbhd}
Suppose $\cW^{\star}$ is convex and $F$ is convex on $\ball(\cW^{\star}, r(l))$ for some $l>0$. Then, we have that $\CPILOCAL(l) \le \frac{(\text{diam}(\cW^{\star})+2r(l_b))^2}{\pi^2} = O(1)$ if $\text{diam}(\cW)=O(1)$ (which is the case for $\beta=\Omega(d)$).
\end{example}
\begin{example}[Local strong convexity]\label{ex:justifyassumption}
Suppose also that $F$ is $\alpha$-strongly convex on $\ball(\cW^{\star}, r(l))$; then $\CPILOCAL(l)=O\prn*{\frac1{\beta}}$. As a special case, consider the following stronger assumption in \citet{lytras2023taming}, see also \citet{li2023riemannian}: $\cW^{\star}=\{\vecW^{\star}\}$, $F$ is $\alpha$-strongly convex at $\vecW^{\star}$, and $\grad^2 F$ is $L'$-Lipschitz in a $\Omega(1)$ neighborhood of $\vecW^{\star}$.\footnote{This applies for small enough $l$ such that $\ball(\cW^{\star}, r(l))$ is a subset of this $\Omega(1)$ neighborhood.}
\end{example}

We emphasize that \pref{ass:Funimodal} or analogous assumptions are in fact \textit{necessary}. For example, even $\cW^{\star}$ being connected is not enough for efficient sampling.
\begin{remark}\label{rem:badlocalisoperimetry}
Consider when $\cW^{\star}$ is dumbbell-shaped. 
Suppose $F(\vecW)=d(\vecW, \cW^{\star})^2$, where $d(\vecW,\vecW^{\star})$ denotes the distance from $\vecW$ to the closest point in $\cW^{\star}$. $F$ is optimizable -- its gradient is nonzero until reaching $\cW^{\star}$.\footnote{One can straightforwardly check this verifies optimizability in our sense.} However due to the poor isoperimetric constant of the dumbbell \citep{vempala2005geometric}, LMC will not mix rapidly upon reaching $\cW^{\star}$, and so the isoperimetric constants of $\mu_{\beta}$ behave poorly for $\beta$ large.
\end{remark}
To our knowledge, the only other related work handling multiple minimizers of $F$ is \citet{gong2024poincare}. Their result also deteriorates when $\cW^{\star}$ has poor isoperimetric constant. Moreover, \pref{ass:Funimodal} does not directly imply a PI; terrible isoperimetry elsewhere gives poor mixing times from arbitrarily initialization. It also does not imply a WPI in terms of $\cS$, the set where optimizability does not hold.

\textbf{Convention.} From here on out, asymptotic notation sometimes hides problem-dependent parameters; however we never suppress $\beta, d$-dependence. Explicit dependencies are written fully in the appendix.
\subsection{Main Results: Poincar\'e and Log-Sobolev Inequalities}\label{subsec:opttoisoperimetry}
Consider the following assumption on the tail growth of $F$, which corresponds to linear tail growth of $F$, which goes hand-in-hand with a PI. 
We note only the second part of this assumption is required for smooth $\Phi$. 
\begin{assumption}\label{ass:linearizableoutsideball}
Suppose that for some $r_1,r_2,R>0$, for all $\vecW \in \ball(\vecW^{\star}, R)^c$, we have $\tri*{\grad F(\vecW),\vecW-\vecW^{\star}} \ge r_1 F(\vecW)\text{ and }F(\vecW) \ge r_2\nrm*{\vecW-\vecW^{\star}}$ for some $\vecW^{\star} \in \cW^{\star}$. 
\end{assumption}
This assumption is very general in the context of optimization, and can be enforced via suitable regularization outside $\ball(\vecW^{\star}, R)$ \citep{raginsky2017non}. The standard dissipativity assumption made in many prior works on non-convex optimization \citep{raginsky2017non, xu2018global, zou2021faster, mou2022improved} are a special case of \pref{ass:linearizableoutsideball}; consequently we present the assumption in the above form. Note as per \citet{raginsky2017non}, for $\beta=\Omega(d)$, the dissipativity assumption (even dissipativity with $b=0$) implies $\mu_{\beta}$ satisfies a PI, but with constant worst-case exponential in dimension.
\begin{theorem}[Establishing PI and LSI under optimizability from all initializations]\label{thm:generalgflinearizabletopi}
Suppose $F$ is optimizable in the sense of \pref{def:optimizabilitydef} for all $\vecW$ and satisfies \pref{ass:linearizableoutsideball}, the corresponding $\Phi$ satisfies \pref{ass:phiselfbounding} ($F$ satisfying \pref{ass:phiselfbounding} is unnecessary here; see \pref{rem:noFregularity}), and \pref{ass:Funimodal} is satisfied for some $l_b>0$. Then for $\beta \ge \Omega(d)$: 
\begin{enumerate}
    \item $\mu_{\beta}$ satisfies a PI with $\CPI =O\prn*{\CPILOCAL+\frac1{\beta}}$, where $\CPILOCAL$ is the Poincar\'e constant of $\mu_{\beta}$ restricted to $\ball(\cW^{\star}, r(l_b))$.
    \item Suppose $F$ is $L$-weakly-convex, that is $\grad^2 F(\vecW) \succeq -L \matI_d$ for some $L>0$, and $F$ has quadratic tail growth, that is, $F(\vecW) \ge m\nrm*{\vecW}^2-b$ for some $m,b>0$.\footnote{Recall quadratic tail growth goes hand-in-hand with a LSI.} Let $S<\infty$ be the second moment of $\mu_{\beta}$. Then $\mu_{\beta}$ satisfies a LSI with constant $\CLSI = O\prn*{\prn*{S+\frac{d}{\beta}+1}\prn*{\beta \CPILOCAL+1}}$.
\end{enumerate}
\end{theorem}
From \pref{thm:generalgflinearizabletopi}, we have established that optimizability of $F$ via GF/GD (under the conditions from above, among which \pref{ass:Funimodal} and \pref{ass:phiselfbounding} are needed) implies PI/LSI at low temperature. These inequalities are the \textit{crux} of non-log-concave sampling via LMC. Central to this proof is the optimizability condition $\tri*{\grad \Phi(\vecW), \grad F(\vecW)} \ge g(F(\vecW))$ from \pref{def:optimizabilitydef}; see \pref{sec:proofsketch}. As such, \pref{thm:generalgflinearizabletopi} confirms our initial \pref{conj:opttosamplingconj}. Later in \pref{subsec:isoperimetrytosampling}, we present corollaries of \pref{thm:generalgflinearizabletopi} for sampling. 

Explicit constants are in the proof in \pref{subsec:proofopttosampling}; they are not included for simplicity. To demonstrate one such example, consider when $\Phi$ is $L$-smooth, which as explained in \pref{sec:examplesapplications} subsumes many cases of interest. Then we have the following, which we expand further on in \pref{rem:linearizableboost}. 
\begin{remark}\label{rem:linearizableboostshort}
Suppose $\Phi$ is $L$-smooth, $g(x)=\lambda x$ for $\lambda \le 1$, and WLOG that $r_1 \le 1/2$. Then $\mu_{\beta}$ satisfies a PI with
\[ \CPI = 2\CPILOCAL + \frac{2}{\beta}\prn*{1+\frac{L}{\lambda l_b}} \text{ for }\beta \ge 2\prn*{1+\frac{L}{\lambda l_b}} \prn*{d+\frac{8R^2}{r_1 L} \lor \frac{2L}{r_1 r_2^2 \lambda^2}}. \numberthis\label{eq:PhismoothPIrequire}\]
\end{remark}
\begin{remark}
We also consider the sharpness of our guarantees. For $F(\vecW)=\nrm*{\vecW}^2$, and more generally strongly convex $F$, the PI constant from \pref{thm:generalgflinearizabletopi} is tight up to $O(\cdot)$. The LSI constant is lossy by around a $\beta$ factor.
\end{remark}
The proof of \pref{thm:generalgflinearizabletopi} uses the Lyapunov function technique in a fairly novel way. Typically one uses a particular ad-hoc Lyapunov function such as $e^{\beta F}$, $F$, or similar, as in \citet{chewi2024ballistic, gong2024poincare, lytras2023taming, li2023riemannian}. Rather, we use $\Phi$ from \pref{def:optimizabilitydef} -- the \textit{exact same} Lyapunov function arising from optimization (recall \pref{def:optimizabilitydef}, from \citet{priorpaper}). We present the main ideas for the proof in \pref{sec:proofsketch} and the full proof in \pref{sec:omittedproofs}.

\subsection{Main Results: Weak Poincar\'e Inequalities}\label{subsec:opttowpi}
We now discuss how to extend our work to when optimizability in the form of \pref{def:optimizabilitydef} holds in some region $\cS$, where we prove a WPI. 
We establish the following; the proof is in \pref{subsec:proofopttosamplingwpi}:
\begin{theorem}[Establishing WPI under optimizability from most initializations]\label{thm:generalgftowpi}
Suppose $F$ is optimizable in the sense of \pref{def:optimizabilitydef} for all $\vecW$ not in some $\cS \subseteq \mathbb{R}^d$, $F$ satisfies \pref{ass:linearizableoutsideball}, $F$ and the corresponding $\Phi$ satisfy \pref{ass:phiselfbounding}, and \pref{ass:Funimodal} is satisfied for some $l_b>0$. 
Then for all $\beta \ge \Omega(d)$, $\mu_{\beta}$ satisfies a $(\CWPI, \delta)$-WPI with $\CWPI=O\prn*{\CPILOCAL+\frac1{\beta}}$, $\delta=O(\mu_{\beta}(\cS))$.
\end{theorem}
$\cS$ typically has small Lebesgue measure $\nu$, for example in the landscape of Phase Retrieval, Matrix Square Root, or the set of `bad initializations' around a saddle point where Gradient Descent does not escape \citep{jain2017global, jin2017escape, lee2019first, priorpaper}. For $\beta \ge \Omega(d)$, $\mu_{\beta}(\cS) \le \frac1Z \exp(-\beta\inf_{\vecW\in\cS} F(\vecW)) \nu(\cS)$, where $Z=\int e^{-\beta F} \DERIV \vecW$. Unless $\cS$ already comprises of favorable near-global-optima or $\nu(\cS)$ is large, this term is small. A crude upper bound follows from Markov's Inequality. Moreover if $F$ is $L$-smooth, for $\beta=\tilde{\Omega}(d)$, we can lower bound $Z \ge e^{-d\ln(\beta L/2\pi)}$; see 3.21, \citet{raginsky2017non}.\footnote{This requires an additional polylog factors.} Thus in this case, the term $\frac1Z \exp(-\beta\inf_{\vecW\in\cS} F(\vecW)) = e^{-\Omega(\beta)}$ is exponentially small. 

Thus by \pref{eq:wpiguarantee}, LMC can sample to accuracy $4\mu_{\beta}(\cS) \nrm*{\tfrac{\DERIV \pi_0}{\DERIV \mu_{\beta}}-1}^2_{\infty} \le \frac1Z \exp(-\beta\inf_{\vecW\in\cS} F(\vecW)) \nu(\cS) \nrm*{\tfrac{\DERIV \pi_0}{\DERIV \mu_{\beta}}-1}^2_{\infty}$. Thus if $\nu(\cS)$ is small and we have a warm start in that $\nrm*{\tfrac{\DERIV \pi_0}{\DERIV \mu_{\beta}}-1}^2_{\infty}$ is controlled, LMC can sample to high accuracy. This confirms the intuition in \pref{conj:opttosamplingconjwpi}.

\begin{remark}\label{rem:linearizableboostshortwpi}
Suppose $\Phi, F$ are $L$-smooth, $g(x)=\lambda x$ for $\lambda \le 1$, and WLOG that $r_1 \le 1/2$. Then $\mu_{\beta}$ satisfies a 
\[\prn*{ 2\CPILOCAL + \frac{2}{\beta}\prn*{1+\frac{B}{\lambda l_b}}, 6\prn*{1+\frac{B}{ \lambda l_b}}\mu_{\beta}(\cS) }-\text{WPI for }\beta \ge 2\prn*{1+\frac{B}{ \lambda l_b}} (d+C''),\]
where $B=L \lor G_F G_{\Phi} \lor 1$, $G_F = \sup_{\vecW\in\cS \cap \ball(\vecW^{\star}, R+1)} \nrm*{\grad F(\vecW)}$, $G_{\Phi} = \sup_{\vecW\in\cS \cap \ball(\vecW^{\star}, R+1)}\nrm*{\grad \Phi(\vecW)}$, $C'' = (\lambda+1)\prn*{\frac{8R^2}{r_1 L} \lor \frac{2L}{r_1 r_2^2 \lambda^2}} + \lambda G_F^2 $. Notice in $\cS$, the region where GF/GD do not succeed, we except $G_F$ to be very small; if $\Phi=F$ (e.g. for P\L, K\L{ }functions), we also obtain that $G_{\Phi}$ is small.
\end{remark}
\begin{corollary}[Of the proof; relaxing \pref{ass:Funimodal}]\label{corr:relaxingFunimodal}
Suppose $\mubetalocal$ satisfies a $(\CWPILOCAL, \deltalocal)$-WPI rather than \pref{ass:Funimodal}. Then in the setting of \pref{thm:generalgflinearizabletopi}, $\mu_{\beta}$ satisfies a $\prn*{O\prn*{\CWPILOCAL+\frac1{\beta}}, 2\deltalocal}$-WPI. Analogously in the setting of \pref{thm:generalgftowpi}, $\mu_{\beta}$ satisfies a $\prn*{O\prn*{\CWPILOCAL+\frac1{\beta}}, O\prn*{\mu_{\beta}(\cS)+2\deltalocal}}$-WPI.
\end{corollary}
\begin{remark}[Sampling with only Local Optimizability]\label{rem:localconditionsampling}
We further note that upon examining the proofs of \pref{thm:generalgflinearizabletopi}, \pref{thm:generalgftowpi}, we only need \pref{def:optimizabilitydef} within $\ball(\vecW^{\star}, R+1)$ for some $\vecW^{\star}\in\cW^{\star}$. This suggests the following interesting algorithmic implication: if \pref{def:optimizabilitydef} only holds locally in $\ball(\vecW^{\star}, R)$, with advance knowledge of $\vecW^{\star}$ and $R$, one can still approximately sample from $\mu_{\beta}$ by regularizing $F$ so \pref{ass:linearizableoutsideball} holds. We elaborate further in \pref{subsec:localoptimizability}; in particular see \pref{prop:obtainlocalcondition}, \pref{corr:localconditionsampling}.
\end{remark}

\subsection{Algorithmic Implications for Sampling}\label{subsec:isoperimetrytosampling}
We now state direct algorithmic implications of \pref{thm:generalgflinearizabletopi}, \pref{thm:generalgftowpi}. We remark \pref{thm:generalgftowpi} yields sampling results for the Langevin Diffusion \pref{eq:LangevinSDE} under a suitable warm start, via \pref{eq:wpiguarantee} (from Theorem 2.1, \citet{rockner2001weak}). Now we will focus on the implications of \pref{thm:generalgflinearizabletopi}. Note establishing improved sampling algorithms under isoperimetry is \textit{not} the main focus of our work; the following results are rather \textit{corollaries} of \pref{thm:generalgflinearizabletopi} via the literature. Again, we believe this is a core \textit{strength} of our work; our results \textit{complement} the literature. Note several recent works have shown the success of discrete-time LMC under solely a PI and smoothness in $\TV$ and $\KL$ divergences, e.g. \citet{chewi21analysis, chen2022improved, altschuler2024faster}. We now are in position to state these implications.
\begin{assumption}[$L$-H\"{o}lder-smoothness]\label{ass:holdercontinuous}
For any $\vecW_1, \vecW_2 \in \mathbb{R}^d$, $\nrm*{\grad F(\vecW_1)-\grad F(\vecW_2)} \le L\nrm*{\vecW_1-\vecW_2}^s$.
\end{assumption}
\begin{corollary}\label{corr:standardregularitysampling}
Suppose $F$ is optimizable by GF in the sense of \pref{def:optimizabilitydef}, the other conditions of \pref{thm:generalgflinearizabletopi} hold, and $F$ satisfies \pref{ass:holdercontinuous}. Then for all $\beta \ge \Omega(d)$, where the $\Omega(\cdot)$ hides $F$-dependent parameters, discrete-time LMC initialized at a distribution $\pi_0 \sim \cN\prn*{\vecOrigin, \frac{1}{2\beta L+\gamma} \matI_d}$ with appropriate step size has the following guarantees, where $\gamma \le 1$ is defined in our proof in \pref{subsec:proofofcorollaries}.
\begin{enumerate}
    \item Suppose $F$ satisfies \pref{ass:holdercontinuous}, that is, $F$ is $L$-H\"{o}lder-continuous with parameter $s$ in $(0,1]$. Then with access to a gradient oracle $\grad F$, the recursion \pref{eq:SGLDiterates} yields a distribution $\pi_T$ with $\TV\prn*{\pi_T||\mu_{\beta}} \le \epsilon \text{ after }T=\tilde{O}\prn*{d \prn*{\CPILOCAL+\frac1{\beta}}^{1+\frac1{s}} \beta^{1+\frac3{s}} \max\crl*{1,\frac{\beta^{s/2}}{d}} \epsilon^{-\frac2{s}}}\text{ iterations}$.
    \item Suppose that $F$ is $L$-smooth. Given additional access to a Proximal Oracle, the Proximal Sampler yields $\mu_T$ with $\DIST\prn*{\pi_T||\mu_{\beta}} \le \epsilon\text{ after } T = \tilde{O}\prn*{\prn*{\CPILOCAL+\frac1{\beta}} \beta d^{1/2} \crl*{\beta+d+ \log\prn*{\frac1{\epsilon}}}} \text{ iterations,}$
    in the metrics $\DIST \in \{\TV, \sqrt{\KL}, \sqrt{\chi^2}\}$. See \pref{subsec:proximalsampler} for more details on the Proximal Sampler.
\end{enumerate}
\end{corollary}
We discuss further details on how the above follows from the literature in \pref{subsec:proofofcorollaries}. 
Furthermore, note \pref{ass:holdercontinuous} does not capture many (optimizable) $F$ of interest, for example simply $F(x)=x^{2p}$ for any $p\ge 1$ in one dimension. 
In \pref{subsec:furthersamplingimplications} we discuss how we can adapt the recent work \citet{lytras2024tamed} to such situations; see \pref{corr:newregularitysampling}. 
Note in both of \pref{corr:standardregularitysampling}, \pref{corr:newregularitysampling}, we do not use information about $\cW^{\star}$ in the initialization, and do not make a warm start hypothesis. The initialization $\frac1{\beta} \matI_d$ is for similar scaling as $\mu_{\beta}$, needed to control initialization, and $\vecOrigin$ is arbitrary.\footnote{The initial divergence can be controlled in \pref{lem:initdivergencecontrolled}, \pref{lem:controlregularizedklinit}, and these divergences already factor into our runtime bounds.}
Our sampling algorithms do not use knowledge of $F$ in initialization; they succeed because the success of GF/GD imply isoperimetry, as per \pref{thm:generalgflinearizabletopi}. Intuitively, the optimizability condition $\tri*{\grad \Phi(\vecW), \grad F(\vecW)} \ge g(F(\vecW))$ allows gradient-based LMC to `find' $\cW^{\star}$ without a warm start.

\section{Examples and Applications}\label{sec:examplesapplications}
The framework of `optimizability' from \pref{def:optimizabilitydef} and \pref{ass:phiselfbounding} subsumes many interesting examples in non-convex (and convex) optimization, from smooth P\L{ }and K\L{ }functions to Phase Retrieval and Matrix Square Root to \textit{all} Linearizable functions; see \citet{priorpaper}. In all these examples \pref{eq:gfcondition} holds, and \pref{ass:phiselfbounding} is satisfied with \textit{dimension-independent} $\rho_{\Phi}$. 
Combining with the conditions of \pref{thm:generalgflinearizabletopi}, \pref{corr:standardregularitysampling}, \pref{corr:newregularitysampling}, we obtain results on isoperimetry and sampling via LMC for many examples.
\begin{example}[P\LCHAR functions]\label{ex:formalPLexample}
Consider smooth P\LCHAR functions $F$, that is with $\nrm*{\grad F(\vecW)}^2 \ge \lambda F(\vecW)$. Then \pref{def:optimizabilitydef} holds with $\Phi(\vecW)=F(\vecW)$ and $g(x)=\lambda x$. Note \pref{ass:phiselfbounding} holds as $F$ is smooth. Note also that $F$ need not be smooth; we only need \pref{ass:phiselfbounding} to hold with $F$ in place of $\Phi$. For example, for $\rho_{\Phi}(x)=A' x + B'$, \pref{ass:phiselfbounding} allows for arbitrary polynomial tail growth of $F$ in $\nrm*{\vecW}$. 
\end{example}
\begin{example}[K\LCHAR functions]\label{ex:formalKLexample}
Now consider K\LCHAR functions $F$, that is with $\nrm*{\grad F(\vecW)}^2 \ge \lambda F(\vecW)^{1+\theta}$ for $\theta \ge 0$. The main difference between the P\LCHAR and K\LCHAR conditions is that the K\LCHAR condition is weaker near the global minima. For K\LCHAR functions $F$, we can take $\Phi(\vecW)=\frac{F(\vecW)}{\lambda}$ in the above, and \pref{def:optimizabilitydef} holds with $g(x)=x^{1+\theta}$, if $F$ satisfies \pref{ass:phiselfbounding} with $\Phi$ in place of $F$. Again, note \pref{ass:phiselfbounding} holds if $F$ is smooth by the definition of smoothness and \pref{lem:smoothnessgradbound}, but that $F$ satisfying \pref{ass:phiselfbounding} is much more general than $F$ being smooth, and in particular allows for any polynomial tail growth of $F$ in $\nrm*{\vecW}$.
\end{example}
\begin{example}[Linearizable/Quasar-Convex Functions]\label{ex:formalLinearizableexample}
Consider $\lambda$-Linearizable functions $F$ \citep{sekhari2021sgd}, that is s.t. $\tri*{\grad F(\vecW), \vecW-\vecW^{\star}} \ge \lambda F(\vecW)$ (also known as Quasar-Convexity, see Definition 3 of \citet{hinder2020near}, or Weak Quasi-Convexity, see \citet{hardt2018gradient}). Here $\Phi(\vecW)=\nrm*{\vecW-\vecW^{\star}}^2$ and $g(x)=\lambda x$, and \pref{def:optimizabilitydef} holds. Note $\Phi$, being 2-smooth, vacuously satisfies \pref{ass:phiselfbounding} by \pref{lem:smoothnessgradbound}. For a PI (\pref{thm:generalgflinearizabletopi}), \pref{ass:phiselfbounding} is not needed on $F$, and thus we obtain a PI with \textit{no} regularity assumptions on $F$. One can obtain the $\beta$-range for which one obtains a PI from our results by taking $L=2$ in \pref{eq:PhismoothPIrequire}.
This setting generalizes numerous other non-convex function classes from optimization, such as star-convex functions \citep{lee2016optimizing} and smooth one-point-strongly convex functions \citep{kleinberg2018alternative}. See \citet{hinder2020near} for further discussion. 
\end{example}
Applying our main results \pref{thm:generalgflinearizabletopi}, \pref{thm:generalgftowpi}, we obtain isoperimetry for all these examples (under the conditions of those Theorems). Noting \pref{ass:Funimodal} is satisfied automatically for all convex $F$, combining \pref{thm:generalgflinearizabletopi} with \pref{corr:newregularitysampling} gives sampling results for log-concave measures beyond smoothness. Formal statements of these corollaries are in \pref{corr:generalexamplescorollary}, \pref{corr:convexbeyondsmooth}.

\section{High-Level Proof Ideas}\label{sec:highlevelpfsketch}
Here, we give high-level proof ideas. A more detailed proof sketch and full proofs are in \pref{sec:proofsketch} and \pref{sec:omittedproofs} respectively.
The central idea is to prove a PI via the Lyapunov potential $\Phi$ from optimization using the Lyapunov function technique from \citet{bakry2008simple}. We develop the technique to fully exploit the property \pref{eq:gfcondition} implied by success of GF/GD, which gives us sharp quantitative control of the isoperimetric constant. For simplicity, we suppose here that $\Phi$ is $L$-smooth and that $g(x)=x$. 

\paragraph{Proving a PI:}
Let $\cU = \ball(\cW^{\star},r(l_b))$, where $l_b$ is any $l$ satisfying \pref{ass:Funimodal}. Consider an arbitrary test function $\psi$. Let $f = \psi - \alpha$, where $\alpha=\int_{\cU} \psi \DERIV \mubetalocal$. For $B>0$ to be chosen later, note as $\frac{t}{t+B}$ is increasing in $t \ge 0$, we have
\begin{align*}
\frac{l_b}{l_b+B} \mathbb{V}_{\mubeta}\brk*{\psi} \le \frac{l_b}{l_b+B}  \int f^2 \DERIV \mubeta &\le \frac{l_b}{l_b+B}  \int_{\cU} f^2 \DERIV \mubeta + \frac{l_b}{l_b+B} \int_{\cU^c} f^2 \DERIV \mubeta\\
&\le \frac{l_b}{l_b+B}  \int_{\cU} f^2 \DERIV \mubeta + \int f^2 \frac{F(\vecW)}{F(\vecW)+B} \DERIV \mubeta.
\end{align*}
We upper bound the first integral $\int_{\cU} f^2 \DERIV \mubeta$ by \pref{ass:Funimodal} and the choice of $\alpha$. For the second integral, note by the condition \pref{eq:gfcondition} and letting $\cL$ denote the so-called \textit{infinitesimal generator} of \pref{eq:LangevinSDE}, we have
\[ F(\vecW)+B \le \tri*{\grad \Phi(\vecW), F(\vecW)}+B = -\frac1{\beta}\cL\Phi(\vecW)+\frac1{\beta}\Delta\Phi(\vecW)+B \le -\frac1{\beta}\cL\Phi(\vecW)+\frac1{\beta}\abs*{\Delta\Phi(\vecW)}+B.\]
We divide by $F(\vecW)+B>0$, multiply both sides by $f^2 \ge 0$, and integrate with respect to $\mu_{\beta}$ to obtain 
\[ \int f^2 \frac{F(\vecW)}{F(\vecW)+B} \DERIV \mubeta \le \frac1{\beta}\int f(\vecW)^2 \frac{-\cL\Phi(\vecW)}{F(\vecW)+B} \DERIV\mu_{\beta} + \frac1{\beta}\int f(\vecW)^2 \frac{\abs*{\Delta\Phi(\vecW)}}{F(\vecW)+B} \DERIV\mu_{\beta}.\]
We upper bound the first integral above using properties of the infinitesimal generator, \pref{lem:intbyparts}. We upper bound the second integral above using smoothness of $\Phi$ and that $\beta=\Omega(d)$. Rearranging and converting back to $\psi$ gives the desired PI. To generalize the proof to non-smooth $\Phi$, we `interpolate' $\Phi$ with the smooth function $\nrm*{\vecW-\vecW^{\star}}^2$ and use \pref{ass:linearizableoutsideball}. 

\paragraph{Proving a WPI:} We follow the same steps as above, except we apply the above inequality \textit{pointwise}, for $\vecW\in\cS^c$ where it holds. We use this to upper bound $\int f^2 \DERIV \mu_{\beta}$ in a similar fashion as above, which in turn lets us upper bound $\int f^2 \frac{F(\vecW)}{F(\vecW)+B} \DERIV \mubeta$. The difference is that we pick up an `error term' $\int_{\cS} f^2 \DERIV \mu_{\beta}$. However by definition of $f$, we have $f^2 \le \text{osc}(\psi)^2$, and so the error term is at most $\text{osc}(\psi)^2 \mu_{\beta}(\cS)$. 

\section{Acknowledgements}
We would like to thank Ayush Sekhari for collaboration on initial stages of the project and generously reading an earlier manuscript, Ahmed El Alaoui for a related collaboration which inspired part of our work, and Brice Huang and Robert Kleinberg for discussions.

\bibliography{sources.bib}

\begin{thebibliography}{88}
\providecommand{\natexlab}[1]{#1}
\providecommand{\url}[1]{\texttt{#1}}
\expandafter\ifx\csname urlstyle\endcsname\relax
  \providecommand{\doi}[1]{doi: #1}\else
  \providecommand{\doi}{doi: \begingroup \urlstyle{rm}\Url}\fi

\bibitem[Ahn et~al.(2024)Ahn, Jadbabaie, and Sra]{ahnescape}
Kwangjun Ahn, Ali Jadbabaie, and Suvrit Sra.
\newblock How to {E}scape {S}harp {M}inima with {R}andom {P}erturbations.
\newblock \emph{Forty-first {I}nternational {C}onference on {M}achine {L}earning}, 2024.

\bibitem[Alaoui et~al.(2025+)Alaoui, Eldan, Gheissari, and Piana]{alaoui2023fast}
Ahmed~El Alaoui, Ronen Eldan, Reza Gheissari, and Arianna Piana.
\newblock Fast relaxation of the random field {I}sing dynamics.
\newblock \emph{Annals of {P}robability (to appear)}, 2025+.

\bibitem[Altschuler and Chewi(2024)]{altschuler2024faster}
Jason~M Altschuler and Sinho Chewi.
\newblock Faster high-accuracy log-concave sampling via algorithmic warm starts.
\newblock \emph{Journal of the ACM}, 71\penalty0 (3):\penalty0 1--55, 2024.

\bibitem[Bakry and {\'E}mery(2006)]{bakry2006diffusions}
Dominique Bakry and Michel {\'E}mery.
\newblock Diffusions hypercontractives.
\newblock In \emph{S{\'e}minaire de Probabilit{\'e}s XIX 1983/84: Proceedings}, pages 177--206. Springer, 2006.

\bibitem[Bakry et~al.(2008)Bakry, Barthe, Cattiaux, and Guillin]{bakry2008simple}
Dominique Bakry, Franck Barthe, Patrick Cattiaux, and Arnaud Guillin.
\newblock A simple proof of the {P}oincar{\'e} inequality for a large class of probability measures.
\newblock \emph{Electronic Communications in Probability}, 13:\penalty0 60--66, 2008.

\bibitem[Bakry et~al.(2014)Bakry, Gentil, and Ledoux]{bakry2014analysis}
Dominique Bakry, Ivan Gentil, and Michel Ledoux.
\newblock \emph{Analysis and geometry of Markov diffusion operators}, volume 103.
\newblock Springer, 2014.

\bibitem[Bandeira et~al.(2023)Bandeira, Maillard, Nickl, and Wang]{bandeira2023free}
Afonso~S Bandeira, Antoine Maillard, Richard Nickl, and Sven Wang.
\newblock On free energy barriers in {G}aussian priors and failure of cold start {MCMC} for high-dimensional unimodal distributions.
\newblock \emph{Philosophical Transactions of the Royal Society A}, 381\penalty0 (2247):\penalty0 20220150, 2023.

\bibitem[Barbier et~al.(2018)Barbier, Krzakala, Macris, Miolane, and Zdeborov{\'a}]{barbier2018optimal}
Jean Barbier, Florent Krzakala, Nicolas Macris, L{\'e}o Miolane, and Lenka Zdeborov{\'a}.
\newblock Optimal errors and phase transitions in high-dimensional generalized linear models.
\newblock In \emph{Conference on {L}earning {T}heory}, 2018.

\bibitem[Bobkov(1999)]{bobkov1999isoperimetric}
Sergey~G Bobkov.
\newblock Isoperimetric and analytic inequalities for log-concave probability measures.
\newblock \emph{The Annals of Probability}, 27\penalty0 (4):\penalty0 1903--1921, 1999.

\bibitem[Bonnefont(2022)]{funcineqgeometrysurvey}
Michel Bonnefont.
\newblock Poincar\'e inequality {W}ith {E}xplicit {C}onstant {I}n {D}imension $d\ge1$.
\newblock \emph{https://www.math.u-bordeaux.fr/~mibonnef/Poincar\'e\_\_Toulouse.pdf}, 2022.

\bibitem[Bontemps(2011)]{bontemps2011bernstein}
Dominique Bontemps.
\newblock Bernstein--von {M}ises theorems for {G}aussian regression with increasing number of regressors.
\newblock \emph{Annals of Statistics}, 39, 2011.

\bibitem[Bovier et~al.(2004)Bovier, Eckhoff, Gayrard, and Klein]{bovier2004metastability}
Anton Bovier, Michael Eckhoff, V{\'e}ronique Gayrard, and Markus Klein.
\newblock Metastability in reversible diffusion processes. {I}. {S}harp asymptotics for capacities and exit times.
\newblock \emph{Journal of the European Mathematical Society}, 6\penalty0 (4):\penalty0 399--424, 2004.

\bibitem[Bovier et~al.(2005)Bovier, Gayrard, and Klein]{bovier2005metastability}
Anton Bovier, V{\'e}ronique Gayrard, and Markus Klein.
\newblock Metastability in reversible diffusion processes {I}{I}: {P}recise asymptotics for small eigenvalues.
\newblock \emph{Journal of the European Mathematical Society}, 7\penalty0 (1):\penalty0 69--99, 2005.

\bibitem[Boyd and Vandenberghe(2004)]{boyd2004convex}
Stephen Boyd and Lieven Vandenberghe.
\newblock \emph{Convex optimization}.
\newblock Cambridge {U}niversity press, 2004.

\bibitem[Brascamp and Lieb(1976)]{brascamp1976extensions}
Herm~Jan Brascamp and Elliott~H Lieb.
\newblock On extensions of the {B}runn-{M}inkowski and {P}r{\'e}kopa-leindler theorems, including inequalities for log concave functions, and with an application to the diffusion equation.
\newblock \emph{Journal of Functional Analysis}, 22\penalty0 (4):\penalty0 366--389, 1976.

\bibitem[Bubeck et~al.(2015)]{bubeck2015convex}
S{\'e}bastien Bubeck et~al.
\newblock Convex optimization: Algorithms and complexity.
\newblock \emph{Foundations and Trends in Machine Learning}, 8\penalty0 (3-4):\penalty0 231--357, 2015.

\bibitem[Cattiaux et~al.(2010)Cattiaux, Guillin, and Wu]{cattiaux2010note}
Patrick Cattiaux, Arnaud Guillin, and Li-Ming Wu.
\newblock A note on {T}alagrand’s transportation inequality and logarithmic {S}obolev inequality.
\newblock \emph{Probability Theory and Related Fields}, 148:\penalty0 285--304, 2010.

\bibitem[Chen et~al.(2024)Chen, Sekhari, and Sridharan]{chen2024langevin}
August~Y Chen, Ayush Sekhari, and Karthik Sridharan.
\newblock Langevin {D}ynamics: {A} {U}nified {P}erspective on {O}ptimization via {L}yapunov {P}otentials.
\newblock \emph{arXiv preprint arXiv:2407.04264}, 2024.

\bibitem[Chen et~al.(2022)Chen, Chewi, Salim, and Wibisono]{chen2022improved}
Yongxin Chen, Sinho Chewi, Adil Salim, and Andre Wibisono.
\newblock Improved analysis for a proximal algorithm for sampling.
\newblock In \emph{Conference on Learning Theory}, pages 2984--3014. PMLR, 2022.

\bibitem[Chewi(2024)]{chewi2024log}
Sinho Chewi.
\newblock Log-concave sampling.
\newblock \emph{Book draft available at https://chewisinho. github.io}, 2024.

\bibitem[Chewi and Stromme(2024)]{chewi2024ballistic}
Sinho Chewi and Austin~J Stromme.
\newblock The ballistic limit of the log-{S}obolev constant equals the {P}olyak-{L}ojasiewicz constant.
\newblock \emph{arXiv preprint arXiv:2411.11415}, 2024.

\bibitem[Chewi et~al.(2024)Chewi, Erdogdu, Li, Shen, and Zhang]{chewi21analysis}
Sinho Chewi, Murat~A Erdogdu, Mufan Li, Ruoqi Shen, and Matthew~S Zhang.
\newblock Analysis of {L}angevin {M}onte {C}arlo from {P}oincar\'e to {L}og-{S}obolev.
\newblock \emph{Foundations of Computational Mathematics}, pages 1--51, 2024.

\bibitem[Chiang et~al.(1987)Chiang, Hwang, and Sheu]{chiang1987diffusion}
Tzuu-Shuh Chiang, Chii-Ruey Hwang, and Shuenn~Jyi Sheu.
\newblock Diffusion for global optimization in r\^{}n.
\newblock \emph{SIAM Journal on Control and Optimization}, 25\penalty0 (3):\penalty0 737--753, 1987.

\bibitem[Damian et~al.(2021)Damian, Ma, and Lee]{damian2021label}
Alex Damian, Tengyu Ma, and Jason~D Lee.
\newblock Label {N}oise {S}{G}{D} {P}rovably {P}refers {F}lat {G}lobal {M}inimizers.
\newblock \emph{Advances in Neural Information Processing Systems}, 34:\penalty0 27449--27461, 2021.

\bibitem[Das et~al.(2023)Das, Nagaraj, and Raj]{das2023utilising}
Aniket Das, Dheeraj~M Nagaraj, and Anant Raj.
\newblock Utilising the {C}{L}{T} {S}tructure in {S}tochastic {G}radient based {S}ampling: {I}mproved {A}nalysis and {F}aster {A}lgorithms.
\newblock In \emph{The Thirty Sixth Annual Conference on Learning Theory}, pages 4072--4129. PMLR, 2023.

\bibitem[De~Sa et~al.(2022)De~Sa, Kale, Lee, Sekhari, and Sridharan]{priorpaper}
Christopher~M De~Sa, Satyen Kale, Jason~D Lee, Ayush Sekhari, and Karthik Sridharan.
\newblock From {G}radient {F}low on {P}opulation {L}oss to {L}earning with {S}tochastic {G}radient {D}escent.
\newblock \emph{Advances in Neural Information Processing Systems}, 35:\penalty0 30963--30976, 2022.

\bibitem[El~Alaoui and Gaitonde(2024)]{el2024bounds}
Ahmed El~Alaoui and Jason Gaitonde.
\newblock Bounds on the covariance matrix of the {S}herrington--{K}irkpatrick model.
\newblock \emph{Electronic Communications in Probability}, 29:\penalty0 1--13, 2024.

\bibitem[Fan et~al.(2023)Fan, Yuan, and Chen]{fan2023improved}
Jiaojiao Fan, Bo~Yuan, and Yongxin Chen.
\newblock Improved dimension dependence of a proximal algorithm for sampling.
\newblock In \emph{The Thirty Sixth Annual Conference on Learning Theory}, pages 1473--1521. PMLR, 2023.

\bibitem[Foster et~al.(2018)Foster, Sekhari, and Sridharan]{foster2018uniform}
Dylan~J Foster, Ayush Sekhari, and Karthik Sridharan.
\newblock Uniform convergence of gradients for non-convex learning and optimization.
\newblock \emph{Advances in Neural Information Processing Systems}, 2018.

\bibitem[Gamerman and Lopes(2006)]{gamerman2006markov}
Dani Gamerman and Hedibert~F Lopes.
\newblock \emph{Markov {C}hain Monte Carlo: stochastic simulation for {B}ayesian inference}.
\newblock Chapman and Hall/CRC, 2006.

\bibitem[Gilks et~al.(1995)Gilks, Richardson, and Spiegelhalter]{gilks1995markov}
Walter~R Gilks, Sylvia Richardson, and David Spiegelhalter.
\newblock \emph{Markov {C}hain Monte Carlo in practice}.
\newblock CRC press, 1995.

\bibitem[Gong et~al.(2024)Gong, He, and Shen]{gong2024poincare}
Yun Gong, Niao He, and Zebang Shen.
\newblock Poincar\'e {I}nequality for {L}ocal {L}og-{P}olyak-{L}ojasiewicz {M}easures: {N}on-asymptotic {A}nalysis in {L}ow-temperature {R}egime.
\newblock \emph{arXiv preprint arXiv:2501.00429}, 2024.

\bibitem[Hardt et~al.(2018)Hardt, Ma, and Recht]{hardt2018gradient}
Moritz Hardt, Tengyu Ma, and Benjamin Recht.
\newblock Gradient descent learns linear dynamical systems.
\newblock \emph{Journal of Machine Learning Research}, 19\penalty0 (29):\penalty0 1--44, 2018.

\bibitem[Hinder et~al.(2020)Hinder, Sidford, and Sohoni]{hinder2020near}
Oliver Hinder, Aaron Sidford, and Nimit Sohoni.
\newblock Near-optimal methods for minimizing star-convex functions and beyond.
\newblock In \emph{Conference on Learning Theory}, pages 1894--1938. PMLR, 2020.

\bibitem[Ho et~al.(2020)Ho, Jain, and Abbeel]{ho2020denoising}
Jonathan Ho, Ajay Jain, and Pieter Abbeel.
\newblock Denoising {D}iffusion {P}robabilistic {M}odels.
\newblock \emph{Advances in Neural Information Processing Systems}, 33:\penalty0 6840--6851, 2020.

\bibitem[Huang and Sellke(2025+)]{huang2023strong}
Brice Huang and Mark Sellke.
\newblock Strong topological trivialization of multi-species spherical spin glasses.
\newblock \emph{Annals of Probability (to appear)}, 2025+.

\bibitem[Huang et~al.(2025)Huang, Mohanty, Rajaraman, and Wu]{huang2024weak}
Brice Huang, Sidhanth Mohanty, Amit Rajaraman, and David~X Wu.
\newblock Weak {P}oincar\'e inequalities, {S}imulated {A}nnealing, and {S}ampling from {S}pherical {S}pin {G}lasses.
\newblock In \emph{Symposium on Theory of Computation}, 2025.

\bibitem[Huang et~al.(2024)Huang, Zou, Dong, Ma, and Zhang]{huangfaster}
Xunpeng Huang, Difan Zou, Hanze Dong, Yian Ma, and Tong Zhang.
\newblock Faster {S}ampling via {S}tochastic {G}radient {P}roximal {S}ampler.
\newblock \emph{Forty-first International Conference on Machine Learning}, 2024.

\bibitem[Jain et~al.(2017)Jain, Jin, Kakade, and Netrapalli]{jain2017global}
Prateek Jain, Chi Jin, Sham Kakade, and Praneeth Netrapalli.
\newblock Global convergence of non-convex gradient descent for computing matrix squareroot.
\newblock In \emph{Artificial Intelligence and Statistics}, pages 479--488. PMLR, 2017.

\bibitem[Jin et~al.(2017)Jin, Ge, Netrapalli, Kakade, and Jordan]{jin2017escape}
Chi Jin, Rong Ge, Praneeth Netrapalli, Sham~M Kakade, and Michael~I Jordan.
\newblock How to {E}scape {S}addle {P}oints {E}fficiently.
\newblock In \emph{International Conference on Machine Learning}, pages 1724--1732. PMLR, 2017.

\bibitem[Jordan et~al.(1998)Jordan, Kinderlehrer, and Otto]{jordan1998variational}
Richard Jordan, David Kinderlehrer, and Felix Otto.
\newblock The variational formulation of the {F}okker--{P}lanck equation.
\newblock \emph{SIAM Journal on Mathematical Analysis}, 29\penalty0 (1):\penalty0 1--17, 1998.

\bibitem[Kale et~al.(2021)Kale, Sekhari, and Sridharan]{sekhari2021sgd}
Satyen Kale, Ayush Sekhari, and Karthik Sridharan.
\newblock S{G}{D}: {T}he {R}ole of {I}mplicit {R}egularization, {B}atch-size and {M}ultiple-epochs.
\newblock \emph{Advances In Neural Information Processing Systems}, 34:\penalty0 27422--27433, 2021.

\bibitem[Karimi et~al.(2016)Karimi, Nutini, and Schmidt]{karimi2016linear}
Hamed Karimi, Julie Nutini, and Mark Schmidt.
\newblock Linear convergence of gradient and proximal-gradient methods under the {P}olyak-{\l}ojasiewicz condition.
\newblock In \emph{Machine Learning and Knowledge Discovery in Databases: European Conference, ECML PKDD 2016, Riva del Garda, Italy, September 19-23, 2016, Proceedings, Part I 16}, pages 795--811. Springer, 2016.

\bibitem[Kinoshita and Suzuki(2022)]{kinoshita2022improved}
Yuri Kinoshita and Taiji Suzuki.
\newblock Improved convergence rate of stochastic gradient {L}angevin dynamics with variance reduction and its application to optimization.
\newblock \emph{Advances in Neural Information Processing Systems}, 35:\penalty0 19022--19034, 2022.

\bibitem[Kleinberg et~al.(2018)Kleinberg, Li, and Yuan]{kleinberg2018alternative}
Bobby Kleinberg, Yuanzhi Li, and Yang Yuan.
\newblock An alternative view: {W}hen does {S}{G}{D} escape local minima?
\newblock In \emph{International Conference on Machine Learning}, pages 2698--2707. PMLR, 2018.

\bibitem[Kook and Vempala(2024)]{kook2024gaussian}
Yunbum Kook and Santosh~S Vempala.
\newblock Gaussian {C}ooling and {D}ikin {W}alks: {T}he {I}nterior-{P}oint {M}ethod for {L}ogconcave {S}ampling.
\newblock In \emph{The Thirty Seventh Annual Conference on Learning Theory}, pages 3137--3240. PMLR, 2024.

\bibitem[Kroese et~al.(2013)Kroese, Taimre, and Botev]{kroese2013handbook}
Dirk~P Kroese, Thomas Taimre, and Zdravko~I Botev.
\newblock \emph{Handbook of monte carlo methods}.
\newblock John Wiley \& Sons, 2013.

\bibitem[Kurdyka(1998)]{kurdyka1998gradients}
Krzysztof Kurdyka.
\newblock On gradients of functions definable in o-minimal structures.
\newblock \emph{Annales de l'institut Fourier}, 48\penalty0 (3):\penalty0 769--783, 1998.

\bibitem[Lee et~al.(2019)Lee, Panageas, Piliouras, Simchowitz, Jordan, and Recht]{lee2019first}
Jason~D Lee, Ioannis Panageas, Georgios Piliouras, Max Simchowitz, Michael~I Jordan, and Benjamin Recht.
\newblock First-order methods almost always avoid strict saddle points.
\newblock \emph{Mathematical Programming}, 176:\penalty0 311--337, 2019.

\bibitem[Lee and Valiant(2016)]{lee2016optimizing}
Jasper~CH Lee and Paul Valiant.
\newblock Optimizing star-convex functions.
\newblock In \emph{2016 IEEE 57th Annual Symposium on Foundations of Computer Science}, pages 603--614. IEEE, 2016.

\bibitem[Lee et~al.(2021)Lee, Shen, and Tian]{lee2021structured}
Yin~Tat Lee, Ruoqi Shen, and Kevin Tian.
\newblock Structured logconcave sampling with a restricted {G}aussian oracle.
\newblock In \emph{Conference on Learning Theory}, pages 2993--3050. PMLR, 2021.

\bibitem[Lehec(2023)]{lehec2023langevin}
Joseph Lehec.
\newblock The {L}angevin {M}onte {C}arlo algorithm in the non-smooth log-concave case.
\newblock \emph{The Annals of Applied Probability}, 33\penalty0 (6A):\penalty0 4858--4874, 2023.

\bibitem[Li and Erdogdu(2023)]{li2023riemannian}
Mufan Li and Murat~A Erdogdu.
\newblock Riemannian langevin algorithm for solving semidefinite programs.
\newblock \emph{Bernoulli}, 29\penalty0 (4):\penalty0 3093--3113, 2023.

\bibitem[Liang and Chen(2022{\natexlab{a}})]{liang2022proximal1}
Jiaming Liang and Yongxin Chen.
\newblock A proximal algorithm for sampling.
\newblock \emph{arXiv preprint arXiv:2202.13975}, 2022{\natexlab{a}}.

\bibitem[Liang and Chen(2022{\natexlab{b}})]{liang2022proximal2}
Jiaming Liang and Yongxin Chen.
\newblock A proximal algorithm for sampling from non-smooth potentials.
\newblock In \emph{2022 Winter Simulation Conference}, pages 3229--3240. IEEE, 2022{\natexlab{b}}.

\bibitem[Liu et~al.(2022)Liu, Zhu, and Belkin]{liu2022loss}
Chaoyue Liu, Libin Zhu, and Mikhail Belkin.
\newblock Loss landscapes and optimization in over-parameterized non-linear systems and neural networks.
\newblock \emph{Applied and Computational Harmonic Analysis}, 59:\penalty0 85--116, 2022.

\bibitem[Lojasiewicz(1963)]{lojasiewicz1963topological}
Stanislaw Lojasiewicz.
\newblock A topological property of real analytic subsets.
\newblock \emph{Coll. du CNRS, Les {\'e}quations aux d{\'e}riv{\'e}es partielles}, 117\penalty0 (87-89):\penalty0 2, 1963.

\bibitem[Lytras and Mertikopoulos(2024)]{lytras2024tamed}
Iosif Lytras and Panayotis Mertikopoulos.
\newblock Tamed {L}angevin sampling under weaker conditions.
\newblock \emph{arXiv preprint arXiv:2405.17693}, 2024.

\bibitem[Lytras and Sabanis(2023)]{lytras2023taming}
Iosif Lytras and Sotirios Sabanis.
\newblock Taming under isoperimetry.
\newblock \emph{arXiv preprint arXiv:2311.09003}, 2023.

\bibitem[Ma et~al.(2019)Ma, Chen, Jin, Flammarion, and Jordan]{ma2019sampling}
Yi-An Ma, Yuansi Chen, Chi Jin, Nicolas Flammarion, and Michael~I Jordan.
\newblock Sampling can be faster than optimization.
\newblock \emph{Proceedings of the National Academy of Sciences}, 116\penalty0 (42):\penalty0 20881--20885, 2019.

\bibitem[Menz and Schlichting(2014)]{menz2014poincare}
Georg Menz and Andr{\'e} Schlichting.
\newblock Poincar{\'e} and logarithmic {S}obolev inequalities by decomposition of the energy landscape.
\newblock \emph{Annals of Probability}, 42\penalty0 (5):\penalty0 1809--1884, 2014.

\bibitem[Montanari and Wu(2023)]{montanari2023posterior}
Andrea Montanari and Yuchen Wu.
\newblock Posterior {S}ampling in {H}igh {D}imension via {D}iffusion {P}rocesses.
\newblock \emph{arXiv:2304.11449}, 2023.

\bibitem[Mou et~al.(2022)Mou, Flammarion, Wainwright, and Bartlett]{mou2022improved}
Wenlong Mou, Nicolas Flammarion, Martin~J Wainwright, and Peter~L Bartlett.
\newblock Improved bounds for discretization of {L}angevin diffusions: {N}ear-optimal rates without convexity.
\newblock \emph{Bernoulli}, 28\penalty0 (3):\penalty0 1577--1601, 2022.

\bibitem[Mousavi-Hosseini et~al.(2023)Mousavi-Hosseini, Farghly, He, Balasubramanian, and Erdogdu]{mousavi2023towards}
Alireza Mousavi-Hosseini, Tyler~K Farghly, Ye~He, Krishna Balasubramanian, and Murat~A Erdogdu.
\newblock Towards a complete analysis of {L}angevin {M}onte {C}arlo: {B}eyond {P}oincar{\'e} {I}nequality.
\newblock In \emph{The Thirty Sixth Annual Conference on Learning Theory}, pages 1--35. PMLR, 2023.

\bibitem[Nesterov et~al.(2018)]{nesterov2018lectures}
Yurii Nesterov et~al.
\newblock \emph{Lectures on convex optimization}, volume 137.
\newblock Springer, 2018.

\bibitem[Otto and Villani(2000)]{otto2000generalization}
Felix Otto and C{\'e}dric Villani.
\newblock Generalization of an inequality by {T}alagrand and links with the logarithmic {S}obolev inequality.
\newblock \emph{Journal of Functional Analysis}, 173\penalty0 (2):\penalty0 361--400, 2000.

\bibitem[Payne and Weinberger(1960)]{payne1960optimal}
Lawrence~E Payne and Hans~F Weinberger.
\newblock An optimal {P}oincar{\'e} inequality for convex domains.
\newblock \emph{Archive for Rational Mechanics and Analysis}, 5\penalty0 (1):\penalty0 286--292, 1960.

\bibitem[Polyak(1963)]{polyak1963gradient}
Boris~T Polyak.
\newblock Gradient methods for the minimisation of functionals.
\newblock \emph{USSR Computational Mathematics and Mathematical Physics}, 3\penalty0 (4):\penalty0 864--878, 1963.

\bibitem[Raginsky et~al.(2017)Raginsky, Rakhlin, and Telgarsky]{raginsky2017non}
Maxim Raginsky, Alexander Rakhlin, and Matus Telgarsky.
\newblock Non-convex learning via {S}tochastic {G}radient {L}angevin {D}ynamics: a nonasymptotic analysis.
\newblock In \emph{Conference on Learning Theory}, pages 1674--1703. PMLR, 2017.

\bibitem[R{\"o}ckner and Wang(2001)]{rockner2001weak}
Michael R{\"o}ckner and Feng-Yu Wang.
\newblock Weak {P}oincar{\'e} inequalities and {L}2-convergence rates of {M}arkov semigroups.
\newblock \emph{Journal of Functional Analysis}, 185\penalty0 (2):\penalty0 564--603, 2001.

\bibitem[Sellke(2024)]{sellke2024threshold}
Mark Sellke.
\newblock The threshold energy of low temperature {L}angevin dynamics for pure spherical spin glasses.
\newblock \emph{Communications on Pure and Applied Mathematics}, 77\penalty0 (11):\penalty0 4065--4099, 2024.

\bibitem[Song et~al.(2021{\natexlab{a}})Song, Meng, and Ermon]{song2020denoising}
Jiaming Song, Chenlin Meng, and Stefano Ermon.
\newblock Denoising {D}iffusion {I}mplicit {M}odels.
\newblock \emph{International Conference on Learning Representations}, 2021{\natexlab{a}}.

\bibitem[Song(2021)]{yangsongblog}
Yang Song.
\newblock Generative {M}odeling by {E}stimating {G}radients of the {D}ata {D}istribution.
\newblock \emph{https://yang-song.net/blog/2021/score/}, 2021.

\bibitem[Song and Ermon(2019)]{song2019generative}
Yang Song and Stefano Ermon.
\newblock Generative {M}odeling by {E}stimating {G}radients of the {D}ata {D}istribution.
\newblock \emph{Advances in Neural Information Processing Systems}, 32, 2019.

\bibitem[Song et~al.(2021{\natexlab{b}})Song, Sohl-Dickstein, Kingma, Kumar, Ermon, and Poole]{song2020score}
Yang Song, Jascha Sohl-Dickstein, Diederik~P Kingma, Abhishek Kumar, Stefano Ermon, and Ben Poole.
\newblock Score-based generative modeling through stochastic differential equations.
\newblock \emph{International Conference on Learning Representations}, 2021{\natexlab{b}}.

\bibitem[Srebro et~al.(2010)Srebro, Sridharan, and Tewari]{srebro2010smoothness}
Nathan Srebro, Karthik Sridharan, and Ambuj Tewari.
\newblock Smoothness, low noise and fast rates.
\newblock \emph{Advances in Neural Information Processing Systems}, 23, 2010.

\bibitem[Stuart(2010)]{stuart2010inverse}
Andrew~M Stuart.
\newblock Inverse problems: a {B}ayesian perspective.
\newblock \emph{Acta Numerica}, 19:\penalty0 451--559, 2010.

\bibitem[Titsias and Papaspiliopoulos(2018)]{titsias2018auxiliary}
Michalis~K Titsias and Omiros Papaspiliopoulos.
\newblock Auxiliary gradient-based sampling algorithms.
\newblock \emph{Journal of the Royal Statistical Society Series B: Statistical Methodology}, 80\penalty0 (4):\penalty0 749--767, 2018.

\bibitem[Vempala(2005)]{vempala2005geometric}
Santosh Vempala.
\newblock Geometric random walks: a survey.
\newblock \emph{Combinatorial and {C}omputational {G}eometry}, 52\penalty0 (573-612):\penalty0 2, 2005.

\bibitem[Vempala and Wibisono(2019)]{vempala2019rapid}
Santosh Vempala and Andre Wibisono.
\newblock Rapid {C}onvergence of the {U}nadjusted {L}angevin {A}lgorithm: {I}soperimetry {S}uffices.
\newblock \emph{Advances in Neural Information Processing Systems}, 32, 2019.

\bibitem[Villani(2009)]{villani2009optimal}
C{\'e}dric Villani.
\newblock \emph{Optimal transport: old and new}, volume 338.
\newblock Springer, 2009.

\bibitem[Villani(2021)]{villani2021topics}
C{\'e}dric Villani.
\newblock \emph{Topics in optimal transportation}, volume~58.
\newblock American Mathematical Society, 2021.

\bibitem[Wainwright(2019)]{wainwright2019high}
Martin~J Wainwright.
\newblock \emph{High-{D}imensional {S}tatistics: {A} {N}on-{A}symptotic {V}iewpoint}, volume~48.
\newblock Cambridge University Press, 2019.

\bibitem[Wang(2006)]{wang2006functional}
Fengyu Wang.
\newblock \emph{Functional {I}nequalities, {M}arkov {S}emigroups and {S}pectral {T}heory}.
\newblock Elsevier, 2006.

\bibitem[Wang and Wibisono(2023)]{wang2023continuized}
Jun-Kun Wang and Andre Wibisono.
\newblock Continuized {A}cceleration for {Q}uasar-{C}onvex {F}unctions in {N}on-{C}onvex {O}ptimization.
\newblock \emph{International Conference on Learning Representations}, 2023.

\bibitem[Xu et~al.(2018)Xu, Chen, Zou, and Gu]{xu2018global}
Pan Xu, Jinghui Chen, Difan Zou, and Quanquan Gu.
\newblock Global {C}onvergence of {L}angevin {D}ynamics {B}ased {A}lgorithms for {N}onconvex {O}ptimization.
\newblock \emph{Advances in Neural Information Processing Systems}, 31, 2018.

\bibitem[Zhang et~al.(2020)Zhang, He, Sra, and Jadbabaie]{zhang2019gradient}
Jingzhao Zhang, Tianxing He, Suvrit Sra, and Ali Jadbabaie.
\newblock Why gradient clipping accelerates training: {A} theoretical justification for adaptivity.
\newblock \emph{International Conference on Learning Representations}, 2020.

\bibitem[Zou et~al.(2021)Zou, Xu, and Gu]{zou2021faster}
Difan Zou, Pan Xu, and Quanquan Gu.
\newblock Faster {C}onvergence of {S}tochastic {G}radient {L}angevin {D}ynamics for {N}on-{L}og-{C}oncave {S}ampling.
\newblock In \emph{Uncertainty in Artificial Intelligence}, pages 1152--1162. PMLR, 2021.

\end{thebibliography}

\newpage
\appendix

\paragraph{Notation.} The domain is $\mathbb{R}^d$, with origin $\vecOrigin$. Let $\nu$ denote Lebesgue measure on $\mathbb{R}^d$. When we write $\nrm*{\cdot}$ without explicitly specifying, we mean the $l_2$ Euclidean norm of a vector. For vectors $\mathbf{a}, \mathbf{b}$, let $\theta\tri*{\vec{a}, \vec{b}}$ denote the directed angle they make in $[0, \pi]$. We denote the Laplacian (sum of second derivatives) of a twice-differentiable function $f$ by $\Delta f$. We denote the Euclidean $l_2$ ball centered at $p\in\mathbb{R}^d$ with radius $R \ge 0$ by $\ball(p,R)$. When $\mathcal{P}$ is a set, $\ball(\mathcal{P}, R)=\{\vecW:\inf_{\vecW' \in \mathcal{P}}\nrm*{\vecW-\vecW'} \le R\}$. We denote the surface of the $d$-dimensional sphere with radius $r$ by $\mathcal{S}^{d-1}(r)$. 
For some $f$ differentiable to $k$ orders, we will let $\nabla^{k} f$ denote the tensor of all the $k$-th order derivatives of $f$, and $\nrm*{\cdot}_{\OPNORM}$ denotes the corresponding tensor's operator norm. 
For a matrix $\matM$, let $\lambda_{\min}(\matM)$ denote its minimum eigenvalue, and $\tr(\matM)$ denote its trace. For matrices $\matM_1, \matM_2$, we let $\succeq$ denote the PSD order, that is $\matM_1 \succeq \matM_2$ if and only if $\matM_1 - \matM_2$ is positive semi-definite. We denote Total Variation distance, Kullback–Leibler divergence, and Chi-squared divergence by $\TV$, $\KL$, $\chi^2$ respectively.

For an arbitrary function $f$, let $\text{osc}(f)=\sup f - \inf f$. Here, $\widetilde{\Omega}$, $\widetilde{\Theta}$, $\widetilde{O}$ hide universal constants and $\log$ factors in $\beta, d, \epsilon$. 
We denote the set of all global minimizers $\vecW^{\star}$ of $F$ by $\cW^{\star}$. We say $F$ is \textit{smooth} ($L$-smooth) if the magnitude of the eigenvalues of its Hessian are universally bounded by a constant (when this constant is at most $L$). We let $Z$ denote the partition function of the corresponding measure, which may change line-by-line (e.g. for different $\beta$). 

\section{Additional Results and Discussion}\label{sec:moreresults}
\subsection{Further Algorithmic Implications of Main Results}\label{subsec:furthersamplingimplications}
The assumption of smoothness or H\"{o}lder continuity does not capture many optimizable $F$ of interest, for example simply $F(x)=x^{2p}$ for any $p>1$ in one dimension. See e.g. \citet{zhang2019gradient} and follow-ups for a study of optimizable $F$ which are not smooth. We thus consider a more general assumption from \citet{lytras2024tamed} (their Assumption 1, slightly simplified) which goes far beyond, allowing for tail growth of $F$ that is any arbitrary polynomial in $\nrm*{\vecW}$. In particular, this assumption can be verified for $F(x)=x^{2p}$, which is not true for \pref{ass:holdercontinuous}. Under this assumption, we obtain less sharp, but still polynomial, convergence rates:
\begin{assumption}[Almost Assumption 1, \citet{lytras2024tamed}]\label{ass:tamedassumptionsimpler}
$F$ satisfies the following:
\begin{itemize}
\item Weak Dissipativity: for some $s_2 \ge 1$, $A_2, b_2>0$, we have for all $\vecW\in\mathbb{R}^d$, $\tri*{\grad F(\vecW), \vecW} \ge A_2 \nrm*{\vecW}^{s_2}-b_2$.
\item Polynomial Jacobian Growth: for some $L_3, s_3>0$ and all $k \ge 2$ for which the following is well-defined, we have for all $\vecW\in\mathbb{R}^d$, $\max\prn*{\nrm*{\grad F(\vecW)}, \nrm*{\grad^k F(\vecW)}_{\OPNORM}} \le L_3(1+\nrm*{\vecW})^{2s_3}$.
\end{itemize}
\end{assumption}
We emphasize we do \textit{not} use these assumptions to obtain isoperimetry in \pref{thm:generalgflinearizabletopi}. Rather, they are just different regularity assumptions under which we obtain different rates for discrete-time LMC. Under these assumptions, and recalling all dependence on $d,\beta$ is polynomial in \pref{thm:generalgflinearizabletopi}, we obtain from \pref{thm:generalgflinearizabletopi} that:
\begin{corollary}\label{corr:newregularitysampling}
Suppose the conditions of \pref{thm:generalgflinearizabletopi} hold and $F$ satisfies \pref{ass:tamedassumptionsimpler}. Moreover, suppose we initialize at a distribution $\pi_0 \propto \exp\prn*{-2\nrm*{\vecW}^{2s_3'}}$ where $s_3' = \max(s_3+\frac12, r+1)$, $r \ge \max(2s_3+1, s_3+2)$. Then assuming knowledge of $A_2, s_1, s_2, s_3$ from \pref{ass:tamedassumptionsimpler} and with this initialization $\pi_0$, for $\beta=\Omega(d)$, discrete-time LMC enjoys the following guarantees:
\begin{enumerate}
    \item Via the discrete-time algorithm Regularized Tamed Unadjusted Langevin (reg-TULA) of \citet{lytras2024tamed}, we have $\TV\prn*{\pi_T || \mu_{\beta}} \le \epsilon\text{ after }T = \tilde{O}\prn*{\textup{poly}\prn*{d, \beta, \CPILOCAL, \frac1{\epsilon}}\log\prn*{\frac{1}{\epsilon}}}$ iterations.
    \item Suppose the assumptions in point 2 of \pref{thm:generalgflinearizabletopi} also hold, which implies $\mu_{\beta}$ satisfies a Log-Sobolev Inequality with constant $\CLSI = O(\beta \max\crl*{S, 1} \max\crl*{\CPILOCAL,1})$. Then via the discrete-time algorithm Weakly Dissipative Tamed Unadjusted Langevin Algorithm (wd-TULA) of \citet{lytras2024tamed}, we have $\TV\prn*{\pi_T || \mu_{\beta}} \le \epsilon$ after $T = \tilde{O}\prn*{\frac{\textup{poly}(d, \beta) \max\crl*{S, 1} \max\crl*{\CPILOCAL,1}}{\epsilon^2} \log\prn*{\frac{1}{\epsilon}}}$ iterations.
\end{enumerate}
\end{corollary}
Both of these sampling algorithms from \citet{lytras2024tamed} are fully detailed in \pref{subsec:tamedsampler}.
Explicit polynomial dependencies can be found in the proof of Theorems 2, 3 from \citet{lytras2024tamed}; the degrees of these polynomials depend (polynomially) on $s_2, s_3$. 

\subsection{Sampling Under Local Optimizability}\label{subsec:localoptimizability}
Suppose rather than global optimizability, $F$ is optimizable by GF only in a large region around $\vecW^{\star}$. Such a situation has been often observed in non-convex landscapes, for example in neural networks \citep{kleinberg2018alternative, liu2022loss}. 
Rather than a WPI, we aim to prove a PI/LSI here for a regularized version of $\mu_{\beta}$, and discuss its algorithmic implications.
We impose the following regularity assumption on $F$:
\begin{assumption}\label{ass:localconditionassumption}
$F$ is $L$-smooth for all $\vecW$, and for some $R>0$:
\begin{itemize}
    \item $F$ is optimizable in $\ball(\vecW^{\star}, R)$ where $g$ in \pref{eq:gfcondition} is of the form $g(x)=\lambda x$ for $\lambda \le 1$.
    \item $\tri*{\grad F(\vecW), \vecW-\vecW^{\star}} \ge 0$ for all $\vecW$ with $R-1 \le \nrm*{\vecW-\vecW^{\star}} \le R$.
    \item $F(\vecW) \ge r_2 \nrm*{\vecW-\vecW^{\star}}$ for some $r_2>0$.
\end{itemize}
\end{assumption}
We can replace the smoothness assumption with \pref{ass:tamedassumptionsimpler} by changing the regularization added to $F$ appropriately, and can also replace 1 in the second condition $R-1 \le \nrm*{\vecW-\vecW^{\star}} \le R$ above by an arbitrary universal constant; see the proof in \pref{subsec:localconditionproofs}. The condition on $g(\cdot)$ above is made for simplicity, and already captures several examples, e.g. P\L{ }and Linearizable functions; again, by suitably modifying the proof one can extend this to general $g(\cdot)$ satisfying the conditions of \pref{def:optimizabilitydef}. We stick with the above and discuss in \pref{rem:localconditionsamplingextend} how to generalize the proof. 

Note here that outside $\ball(\vecW^{\star}, R)$, besides smoothness and a lower bound on growth, $F$ could have arbitrarily many points with vanishing gradient, saddle points and local minima.\footnote{Smoothness and the lower bound on growth do not `sandwich' $F$ in a way that implies a lack of critical points.} This contrasts to the main result of \citet{gong2024poincare}, where despite its supposed `local' nature, their Assumption 5 lower bounds on $\nrm*{\grad F}$ and the lack of saddle points are assumed outside a compact set.

By regularizing $F$ appropriately, we establish:
\begin{proposition}\label{prop:obtainlocalcondition}
Suppose \pref{ass:localconditionassumption} holds, the corresponding $\Phi$ satisfies \pref{ass:phiselfbounding}, and \pref{ass:Funimodal} is satisfied for some $l_b>0$ with $\ball(\cW^{\star}, r(l_b)) \subseteq \ball(\vecW^{\star}, R-1)$ for some $\vecW^{\star} \in \cW^{\star}$. 
Let $\hat{F}(\vecW) = F(\vecW)+\chi_F(\vecW) \cdot L (\nrm*{\vecW-\vecW^{\star}}^2+1)$ where $\chi_F \in [0,1]$ is a suitable interpolant which depends on problem parameters, defined in our proof in \pref{eq:interpolateFlocalconditiondef}. 

Then for $\beta \ge \Omega(d)$, $\hat{\mu}_{\beta} \propto e^{-\hat{F}}/Z$ satisfies a PI with constant $O\prn*{\CPILOCAL+\frac1{\beta}}$. Furthermore, $\hat{F}$ is smooth with $O(1)$ smoothness constant.
\end{proposition}
Explicit constants are in the proof in \pref{subsec:localconditionproofs}. We note that under the conditions of point 2 of \pref{thm:generalgflinearizabletopi} and via the same proof, one can also extend this to an LSI. 
As an algorithmic implication, \pref{prop:obtainlocalcondition} directly shows the following.
\begin{corollary}\label{corr:localconditionsampling}
Let $\delta = \mu_{\beta}\prn*{\ball(\vecW^{\star}, R-1)^c}$. Given oracle access to $\grad F$, $F$ and knowledge of $\vecW^{\star}\in\cW^{\star}$ satisfying the conditions of \pref{prop:obtainlocalcondition}, $R$, and $g(\cdot)$, then running LMC in both continuous and discrete time with $\grad \hat{F}$ in place of $\grad F$ yields a distribution $\pi$ such that $\TV(\pi, \mu_{\beta}) \le \epsilon+3\delta$, in time $O\prn*{\poly(d,\beta,\frac1{\epsilon})}$.
\end{corollary}
\proof[Proof of \pref{corr:localconditionsampling}]{
By \pref{prop:obtainlocalcondition} and \pref{corr:standardregularitysampling}, \pref{corr:newregularitysampling}, in continuous and discrete time, LMC yields a distribution $\pi$ such that $\TV(\pi, \hat{\mu}_{\beta}) \le \epsilon$ in time $O\prn*{\poly (d,\beta,\frac1{\epsilon})}$. 
Note LMC is implementable because we can construct $\grad \hat{F}$ using knowledge of $\grad F$, $\vecW^{\star}\in\cW^{\star}$ satisfying the conditions of \pref{prop:obtainlocalcondition}, $R$, and problem-dependent parameters.
The problem-dependent parameters are defined in the proof of \pref{subsec:localconditionproofs}, and can be computed with oracle access to $F, \grad F$, knowledge of $\vecW^{\star}, R, g(\cdot)$, and appropriate cross validation; we expand on this in \pref{rem:computeprobdependentparam} in \pref{subsec:localconditionproofs}.
Hence we can implement LMC and produce a hypothesis $\pi$ which approximately samples from $\hat{\mu}_{\beta}$ as per the above.
Thus, we have 
\[ \TV(\pi, \mu_{\beta}) \le \TV(\pi, \hat{\mu}_{\beta}) + \TV(\hat{\mu}_{\beta}, \mu_{\beta}) \le \epsilon+3\delta,\]
where the last step is verified in \pref{lem:closetvlocalcondition}. $\hfill\blacksquare$
}

We conclude from \pref{corr:localconditionsampling} that optimizability from appropriate neighborhoods of the global minima yields sampling guarantees, via running LMC on a regularized version of $F$. Running LMC on a regularized version of $F$ has seen recent interest, as a way to sample from $\mu_{\beta}$ under relaxed regularity assumptions \citep{lytras2023taming, lytras2024tamed}. Here we offer a novel perspective justifying the benefit of regularization for LMC as a way we can sample from a regularized Gibbs measure if we only have `local optimizability', and possibly adversarial behavior outside of this neighborhood.

\subsection{Further Discussion of Examples and Implications}\label{subsec:furtherexamples}
We first expand on why the natural settings \pref{ex:Fconvexnbhd}, \pref{ex:justifyassumption} are subsumed by \pref{ass:Funimodal}:
\begin{itemize}
    \item \pref{ex:Fconvexnbhd} (local convexity): Suppose $\cW^{\star}$ is convex and $F$ is convex on $\ball(\cW^{\star}, r(l))$ for some $l>0$. Note convexity of $\cW^{\star}$ implies convexity of $\ball(\cW^{\star}, r(l))$ (Exercise 2.14, \citet{boyd2004convex}). By the Payne-Weinberger Theorem \citep{payne1960optimal}, in the form of Theorem 6.2 of \citet{funcineqgeometrysurvey}, we see $\CPILOCAL(l) \le \frac{(\text{diam}(\cW^{\star})+2r(l_b))^2}{\pi^2} = O(1)$ if $\text{diam}(\cW)=O(1)$. Note $\text{diam}(\cW)=O(1)$ is the case for $\beta=\Omega(d)$.
    \item \pref{ex:justifyassumption} (local strong convexity): As a special case of the above, suppose additionally that $F$ is $\alpha$-strongly convex on $\ball(\cW^{\star}, r(l))$. Then $\CPILOCAL(l)=O\prn*{\frac1{\beta}}$ by Brascamp-Lieb \citep{brascamp1976extensions} in the form of Theorem 5.1, \citet{funcineqgeometrysurvey}.\footnote{Which applies to a domain of $\mathbb{R}^d$ with convex boundary, see page 20, \citet{funcineqgeometrysurvey}.} A special case of this is the following stronger assumption in \citet{lytras2023taming}, also considered in \citet{li2023riemannian}: $\cW^{\star}=\{\vecW^{\star}\}$, $F$ is $\alpha$-strongly convex at $\vecW^{\star}$, and the Hessian of $F$ is $L'$-Lipschitz in a $\Omega(1)$ neighborhood of $\vecW^{\star}$. 
To see why, consider $l_b>0$ small enough so that in $\ball(\cW^{\star}, r(l_b))$, the Hessian of $F$ is $L'$-Lipschitz, and $r(l_b) \le \frac{\alpha}{2L'}$. This is possible by taking $l_b$ small enough. Using that eigenvalues are 1-Lipschitz in the Hessian, we see for any $\vecW$ and arbitrary $\vecW^{\star} \in \cW^{\star}$ that
\begin{align*}
\abs*{\lambda_{\MIN}(\grad^2 F(\vecW))}=\abs*{\lambda_{\MIN}(\grad^2 F(\vecW))-\lambda_{\MIN}(\grad^2 F(\vecW^{\star}))} &\le \nrm*{\grad^2 F(\vecW)-\grad^2 F(\vecW^{\star})}_{\OPNORM} \le L'\nrm*{\vecW-\vecW^{\star}}.
\end{align*}
It follows for all $\vecW$ with $\nrm*{\vecW-\vecW^{\star}} \le \frac{\alpha}{2L'}$, $F$ is $\alpha/2$-strongly convex. 
\end{itemize}
We next formally instantiate the corollaries of \pref{thm:generalgflinearizabletopi}, \pref{thm:generalgftowpi} for the examples from \pref{sec:examplesapplications}. Directly applying \pref{thm:generalgflinearizabletopi}, \pref{thm:generalgftowpi} for $F$ satisfying the conditions of \pref{ex:formalPLexample}, \pref{ex:formalKLexample}, \pref{ex:formalLinearizableexample} implies the following.
\begin{corollary}[Implications for Isoperimetry and Sampling]\label{corr:generalexamplescorollary}
Suppose $F$ satisfies the conditions of \pref{ex:formalPLexample} (P\L), \pref{ex:formalKLexample} (K\L), or \pref{ex:formalLinearizableexample} (Linearizable/Quasar-Convex), and $F$ also satisfies \pref{ass:Funimodal}, \pref{ass:linearizableoutsideball}.\footnote{Recall \pref{ass:linearizableoutsideball} is unnecessary for $F$ smooth.} Then for $\beta=\Omega(d)$, we have the following:
\begin{itemize}
    \item $\mu_{\beta}$ satisfies a PI with $\CPI =O\prn*{\CPILOCAL+\frac1{\beta}}$. 
    \item Under the conditions of point 2 of \pref{thm:generalgflinearizabletopi}, $\mu_{\beta}$ satisfies a LSI with $\CLSI = O\prn*{\prn*{S+\frac{d}{\beta}+1}\prn*{\beta \CPILOCAL+1}}$, where $S$ is the second moment of $\mu_{\beta}$.
    \item Suppose that $F$ satisfies the conditions of \pref{ex:formalPLexample}, \pref{ex:formalKLexample}, or \pref{ex:formalLinearizableexample} outside some set $\cS$. In this case, $\mu_{\beta}$ satisfies a $O\prn*{\prn*{\CPILOCAL+\frac1{\beta}}, O(\mu_{\beta}(\cS))}$-WPI.
    \item As per \pref{corr:relaxingFunimodal}, we can obtain a WPI for all these examples if $\mubetalocal$ does not satisfy \pref{ass:Funimodal} but instead satisfies a $(\CWPILOCAL, \deltalocal)$-WPI. 
    \item Via \pref{corr:standardregularitysampling}, \pref{corr:newregularitysampling}, we obtain $\textup{poly}\prn*{d,\beta,\frac1{\epsilon}}$ sampling guarantees for discrete-time LMC under \pref{ass:holdercontinuous}, \pref{ass:tamedassumption}.
\end{itemize}
\end{corollary}
Our sampling results hold when $F$ satisfies \pref{ass:Funimodal}. While this or analogous conditions are necessary as per \pref{rem:badlocalisoperimetry}, note convex $F$ automatically satisfy \pref{ass:Funimodal}. Thus taking $l_b=1$ in \pref{thm:generalgflinearizabletopi}, using the result of Payne and Weinberger \citep{payne1960optimal} which states $\CPILOCAL = O\prn*{\text{diam}(\cW^{\star})^2+r(l_b)^2}$, and combining with \pref{ex:Fconvexnbhd}, we directly obtain the following.
\begin{corollary}[Sampling from non-smooth convex functions via LMC]\label{corr:convexbeyondsmooth}
Suppose $F$ is convex. Then
\[ \CPI(\mu_{\beta})=  O\prn*{\text{diam}(\cW^{\star})^2+r(l_b)^2+ \frac{1}{\beta}} \text{ for }\beta \ge \Omega\prn*{d+4R^2 \lor \frac{2}{r_2^2}}.\]
Furthermore, as a direct corollary of \pref{corr:newregularitysampling}, we obtain results on sampling from particular log-concave measures (with the temperature restriction) where the potential is not smooth, similar to \citet{lehec2023langevin}. In fact, in some senses our results are stronger; those of \citet{lehec2023langevin} (see Theorem 5) do not permit tail growth of $F$ that is an arbitrary polynomial in $\nrm*{\vecW}$. 
\end{corollary}

\subsection{Sampling Under a Stochastic Gradient Oracle}\label{subsec:stochasticgradsampling}
We can also use our results on a Log-Sobolev Inequality, in particular part 2 of \pref{thm:generalgflinearizabletopi} for $F$ optimizable from all initializations, to show we can sample from $\mu_{\beta}$ when we only have a stochastic gradient oracle $\grad f \approx \grad F$. To the best of knowledge, the most recent guarantees in this setting are \citet{das2023utilising, huangfaster}, where a variety of discretizations of \pref{eq:LangevinSDE} are considered. For the algorithms themselves, we refer the reader to these papers. 

Under standard assumptions on bounded variance of a stochastic gradient oracle, to the best of our knowledge, the state-of-the-art guarantees for LMC in this setting are Theorems 4.1 and 4.2 of \citet{huangfaster}. The results of \citet{huangfaster} state the following. Suppose $F$ satisfies $L$-smoothness and $\mu_{\beta}$ satisfies a Log-Sobolev Inequality with constant $\CLSI$, and that $f$ is written as a finite sum log-density. Then letting $\sigma$ be an upper bound on the variance of the stochastic gradients $\grad f$, we can sample in $\TV$-error $\epsilon$ from $\mu_{\beta}$ using $\tilde{O}\prn*{\frac{\beta^3 \CLSI^3 d^{1/2} \min\crl*{d+\beta^2 \sigma^2, d^{1/2} \beta^2 \sigma^2}}{\epsilon^2}}$ expected queries to the stochastic gradient oracle.

Combine this with the second part of our \pref{thm:generalgflinearizabletopi} for optimizable $F$. Under the assumptions of the second part of \pref{thm:generalgflinearizabletopi}, and that $F$ is finite-sum and $L$-smooth, we obtain the following from \pref{thm:generalgflinearizabletopi}:
\begin{itemize}
    \item In the setting of \pref{ex:Fconvexnbhd} (local convexity): Here $\CPILOCAL=O(1)$ and so $\CLSI(\mu_{\beta})=O\prn*{\prn*{S+\frac{d}{\beta}+1} \beta}$. As a direct coollary of the above, we obtain a sampling guarantee in $\TV$ of $\tilde{O}\prn*{\frac{\beta^8 d \prn*{S+\frac{d}{\beta}+1}^3 \sigma^2}{\epsilon^2}}$ for the algorithm given in Theorem 4.1 of \citet{huangfaster}.
    \item In the setting of \pref{ex:justifyassumption} (local strong convexity): Here $\CPILOCAL=O\prn*{1/\beta}$ and so $\CLSI(\mu_{\beta})=O\prn*{S+\frac{d}{\beta}+1}$. As a direct coollary of the above, we obtain a sampling guarantees in $\TV$ of $\tilde{O}\prn*{\frac{\beta^5 d  \prn*{S+\frac{d}{\beta}+1}^3 \sigma^2}{\epsilon^2}}$ for the same algorithm from Theorem 4.1 of \citet{huangfaster}.
\end{itemize}
Note if we also assume the standard dissipativity condition in \citet{raginsky2017non, xu2018global, zou2021faster}, by Lemma 1 of \citet{raginsky2017non}, we can take $S=O\prn*{d/\beta}$ in the above.

\section{Additional Background}\label{sec:additionalbackground}
\subsection{Markov Semigroup Theory}\label{subsec:backgroundmarkovsemigroup}
We introduce the concept of the infinitesimal generator of a Markov process, which will make this exposition and our proofs much more natural. We give only what is needed for our work and refer the reader to \citet{chewi2024log, bakry2014analysis} for more details. 
\begin{definition}
The infinitesimal generator of a Markov process $\vecW(t)$ is the operator $\mathcal{L}$ defined on all (sufficiently differentiable) functions $\phi$ by 
\[ \mathcal{L}\phi(\vecW) = \lim_{t\rightarrow0}\frac{\mathbb{E}\brk*{\phi(\vecW(t))}-\phi(\vecW)}{t}.\]
\end{definition}
This can be thought of as the instantaneous derivative of the Markov process in expectation. It is well-known that for the Langevin Diffusion \pref{eq:LangevinSDE}, the generator takes the following form:
\[ \mathcal{L}\phi(\vecW) = -\tri*{\beta \grad F(\vecW), \grad \phi(\vecW)} + \Delta \phi(\vecW).\numberthis\label{eq:generatorlangevin}\]
For example, this calculation can be found in Example 1.2.4 of \citet{chewi2024log}.

We also need to introduce the idea of symmetry of the measure $\mu$ with respect to the stochastic process. In particular, we say $\mu$ is \textit{symmetric} with respect to the Langevin Diffusion \pref{eq:LangevinSDE} if for all infinitely differentiable $f, g$, 
\[ \int f \cL g \DERIV\mu = \int \cL f g \DERIV\mu.\]
It is well-known that $\mu_{\beta}$ is symmetric. See e.g. Example 1.2.18 of \citet{chewi2024log}. This is used in \pref{lem:intbyparts}.

\subsection{The Proximal Sampler}\label{subsec:proximalsampler}
Earlier we only discussed the discretization \pref{eq:SGLDiterates} of the Langevin Diffusion \pref{eq:LangevinSDE}, which as shown in \citet{chewi21analysis, vempala2019rapid}, succeeds in sampling from $\mu_{\beta}$ if $\mu_{\beta}$ satisfies an isoperimetric inequality. Another discretization of \pref{eq:LangevinSDE} that can sample from $\mu_{\beta}$ if $\mu_{\beta}$ satisfies an isoperimetric inequality is the \textit{Proximal Sampler}, first introduced in \citet{titsias2018auxiliary, lee2021structured}. See \citet{lee2021structured, chen2022improved, liang2022proximal1, liang2022proximal2, fan2023improved, altschuler2024faster} for a variety of important developments on the proximal sampler. To the best of our knowledge, the state-of-the-art guarantees for the Proximal Sampler with exact gradients are in \citet{altschuler2024faster}, \citet{fan2023improved}. For state-of-the-art guarantees for the Proximal Sampler with stochastic gradients, see \citet{huangfaster}. The Proximal Sampler is motivated by the Proximal Point Method in optimization, and works as follows: fix $h>0$ and consider the joint distribution $\pi$ on $\mathbb{R}^d \times \mathbb{R}^d$ defined as follows:
\[ \pi(\vecW, \vecW') := \frac1{Z} \exp\prn*{-\beta F(\vecW)-\frac1{2h}\nrm*{\vecW-\vecW'}^2}.\]
Initialize $\vecW_0 \sim \pi_0$ and perform the following recursion between two sequences $\vecW_k$ (the samples of interest) and $\vecW'_k$ (an auxiliary sequence) for $k \ge 0$:
\begin{enumerate}
\item Sample $\vecW'_k \sim \pi^{\vecW'|\vecW}\prn*{\cdot|\vecW_k}=\cN\prn*{\vecW_k, h I_d}$.
\item Sample the next iterate $\vecW_{k+1} \sim \pi^{\vecW|\vecW'}\prn*{\cdot|\vecW'_k}$.
\end{enumerate}
Notice the second step is implementable if $F$ is $L$-smooth for small enough $h\le \frac1{2 \beta L}$, as for such $h$, $\pi^{\vecW|\vecW'}\prn*{\cdot|\vecW}$ is log-concave. In fact in \citet{altschuler2024faster} and many other works on the proximal sampler, it is shown the Proximal Sampler is implementable with a \textit{Proximal Oracle}, which given $\vecW' \in \mathbb{R}^d$, returns
\[ \text{argmin}_{\vecW \in \mathbb{R}^d}\prn*{F(\vecW)+\frac1{2h}\nrm*{\vecW-\vecW'}^2}.\]
A Proximal Oracle is implementable if $F$ is smooth, as for small enough $h$, the above optimization problem is smooth and strongly convex. When we cite Theorems 5.3, 5.4 from \citet{altschuler2024faster}, we assume $F$ is smooth.

\subsection{The Tamed Unadjusted Langevin Algorithm}\label{subsec:tamedsampler}
Here, we describe in detail the Weakly-Dissipative/Regularized Tamed Unadjusted Langevin Algorithm from \citet{lytras2024tamed}. In recent years, works such as \citet{lehec2023langevin, lytras2023taming, lytras2024tamed} have aimed to develop sampling algorithms that succeed beyond the relatively restrictive smoothness or H\"{o}lder continuity conditions in a variety of settings. As shown in 2.3 of \citet{lytras2024tamed}, one needs to modify the sampling algorithm beyond \pref{eq:SGLDiterates} to sample from the Gibbs measure when $F$ grows faster than a quadratic in $\nrm*{\vecW}$. To our knowledge, the most general guarantees are in \citet{lytras2024tamed}, and so we go with the results from there. The idea of these tamed sampling schemes is to split the gradient into two parts: one that grows at most linearly, and another part which we `tame'. This allows for convergence results under far milder regularity conditions, Assumption 1 of \citet{lytras2024tamed}, which we fully present in \pref{ass:tamedassumption}, though we note \pref{ass:tamedassumption} is implied by \pref{ass:tamedassumptionsimpler}.

The Weakly-Dissipative Tamed Unadjusted Langevin Algorithm (wd-TULA) from their work gives an algorithm with more efficient guarantees under weak convexity of $F$ or a LSI, and is defined as follows. Letting $\eta$ denote the step size, we first let 
\[ f(\vecW) := \beta \grad F(\vecW) - \beta A_2 \vecW (1+\nrm*{\vecW}^2)^{\frac{s_2}2-1}, f_{\eta}(\vecW) = \frac{f(\vecW)}{1+\sqrt{\eta}\nrm*{\vecW}^{2s_3}},\]
where $A_2, s_2, s_3$ are defined in \pref{ass:tamedassumptionsimpler}.
We then let
\[ h_{\eta}(\vecW) := \beta A_2 \vecW (1+\nrm*{\vecW}^2)^{\frac{s_2}2-1} + f_{\eta}(\vecW),\]
and use $h_{\eta}(\vecW)$ in place of $\beta \grad F(\vecW)$ in \pref{eq:SGLDiterates}. That is, for standard $d$-dimensional normals $\vecEps_t$,
\[ \vecW_{t+1}=\vecW_t-\eta h_{\eta}(\vecW_t)+\sqrt{2\eta}\vecEps_t.\numberthis\label{eq:tamediterates}\]
We use Theorem 2 of \citet{lytras2024tamed}, which obtains a nonasymptotic polynomial-time guarantee for \pref{eq:tamediterates} under \pref{ass:tamedassumption} and a LSI for $\mu_{\beta}$. The guarantee depends on the initialization $\KL(\pi_0||\mu_{\beta})$, but we argue in \pref{lem:controlregularizedklinit} that this can be controlled for appropriate $\pi_0$.

However, the Weakly-Dissipative Tamed Unadjusted Langevin Algorithm (wd-TULA) does not succeed when $\mu_{\beta}$ satisfies a PI. To this end, for large enough $r$ (for example, $r=4s_3+4$ is enough), we instead define \pref{eq:tamediterates} the same way as above, except $F$ is replaced by a regularized version, $F(\vecW)+\frac{\lambda}{\beta}\nrm*{\vecW}^{2r+2}$. That is, in defining $f(\vecW)$ above, we take $\grad \prn*{F(\vecW)+\frac{\lambda}{\beta}\nrm*{\vecW}^{2r+2}}$ rather than $\grad F(\vecW)$. This yields the Regularized Tamed Unadjusted Langevin Algorithm (reg-TULA). In Theorem 3 of \citet{lytras2024tamed}, reg-TULA was shown to succeed in sampling from $\mu_{\beta}$ under \pref{ass:tamedassumption} and a PI for $\mu_{\beta}$. Again, we argue in \pref{lem:controlregularizedklinit} that the initialization error can be controlled for appropriate $\pi_0$.

\section{Proof Ideas}\label{sec:proofsketch}
Here, we sketch our proof; our full proofs are in \pref{sec:omittedproofs}. We invite the reader interested in learning our proofs to first read this subsection, as we will build off the work here in \pref{sec:omittedproofs}.

\subsection{Proving a PI}
The central idea is to prove a PI via the Lyapunov potential arising from optimization, a similar idea to \cite{bakry2008simple}. However, we modify their technique in a novel way to fully exploit local geometric properties implied by success of Gradient Descent, which gives us sharper quantitative control of the isoperimetric constant. Rather than building an ad-hoc Lyapunov potential from $F$, we instead utilize $\Phi$ as our potential in proving the functional inequality.

In our setting, recall we have a twice-differentiable and non-negative \textit{Lyapunov function} $\Phi(\vecW)$ such that 
\[ \tri*{ \nabla\Phi(\vecW),\nabla F(\vecW)} \ge g\prn*{F(\vecW)} \]
for a non-negative, monotonically increasing $g$ with $g(x) \ge m'x-b'$, $g(0)=0$, and $g(x)>0$ for $x>0$. 

Recall the definition of the infinitesimal generator $\cL$ of \pref{eq:LangevinSDE}, defined as the following operator on any sufficiently differentiable test function $\phi$:
\[ \cL\phi(\vecW) = \Delta \phi(w) - \tri*{ \beta \nabla F(\vecW), \nabla \phi(\vecW)}.\]
Crucial to our analysis is the following Integration by Parts identity, which holds by reversibility of the Langevin Diffusion \pref{eq:LangevinSDE}:
\begin{lemma}[Theorem 1.2.14, \citet{chewi2024log}]\label{lem:intbyparts}
For all functions $f, g$ for which $\cL f$, $\cL g$ are defined,
\[ \int (-\cL) fg \DERIV \mu_{\beta} = \int f(-\cL) g \DERIV \mu_{\beta} = \int \tri*{\grad f, \grad g}\DERIV \mu_{\beta}.\]
\end{lemma}
For more background on the infinitesimal generator and the above identity, see \pref{subsec:backgroundmarkovsemigroup}.

Now, we outline our argument. Following the discussion from \pref{sec:highlevelpfsketch}, it remains to upper bound a term of the form $\int f(\vecW)^2 \frac{g\prn*{F(\vecW)}}{g\prn*{F(\vecW)}+B} \DERIV\mu_{\beta}$. We do so via \pref{lem:lyapunovmethodsetup}, which crucially uses \pref{lem:intbyparts}. Consider any $B>0$, and let $h(\vecW) = g(F(\vecW))+B$. \pref{lem:lyapunovmethodsetup} gives
\begin{align*}
\int f(\vecW)^2 \frac{g\prn*{F(\vecW)}}{g\prn*{F(\vecW)}+B} \DERIV\mu_{\beta} &\le \frac1{\beta} \int \prn*{ \nrm*{\nabla f(\vecW)}^2+\frac{f(\vecW)^2}{h(\vecW)^2}\nrm*{\nabla \Phi(\vecW)}^2 - \frac{f(\vecW)^2}{h(\vecW)^2}\tri*{ \nabla h(\vecW), \nabla \Phi(\vecW)}} \DERIV\mu_{\beta} \\
&\hspace{1in}+ \frac1{\beta}\int f(\vecW)^2 \frac{\abs*{\Delta\Phi(\vecW)}}{h(\vecW)} \DERIV\mu_{\beta}.
\end{align*}
With \pref{lem:lyapunovmethodsetup} in hand and following the ideas from \pref{sec:highlevelpfsketch}, we prove \pref{thm:generalgflinearizabletopi} by using \pref{ass:linearizableoutsideball} to upper bound
\[ \frac{\nrm*{\nabla \Phi(\vecW)}^2}{h(\vecW)^2} - \frac{\tri*{ \nabla h(\vecW), \nabla \Phi(\vecW)}}{h(\vecW)^2} \le C',\]
for some $C'>0$. Plugging this upper bound into the implication of \pref{lem:lyapunovmethodsetup} above and combining with the discussion from \pref{sec:highlevelpfsketch}, rearranging and converting from $f$ back to $\psi$ eventually gives the desired PI. Using the `tightening' technique of \citet{cattiaux2010note}, we can strengthen the PI into a LSI for $\mu_{\beta}$ under the assumption of quadratic tail growth for $F$ -- which goes hand-in-hand with a LSI -- and weak-convexity. Finally, once we have proved a PI or LSI, sampling from $\mu_{\beta}$ via LMC is known from the literature. This is fully detailed in \pref{sec:omittedproofs}. 

\subsection{Proving a WPI}
We also extend the Lyapunov technique to prove an WPI, which may be of independent interest. The idea is as follows: if \pref{eq:gfcondition} does not hold in $\cS$ but otherwise holds in $\cS^c$, instead consider arbitrary test function $\psi$ and let $f=\psi-\alpha$ be defined exactly as in \pref{sec:highlevelpfsketch}. 

Considering any $B>0$, for all $\vecW\in \cS^c$, we have:
\[ 0<g\prn*{F(\vecW)}+B \le \tri*{\grad \Phi(\vecW), F(\vecW)}+B = -\frac1{\beta}\cL\Phi(\vecW)+\frac1{\beta}\Delta\Phi(\vecW)+B \le -\frac1{\beta}\cL\Phi(\vecW)+\frac1{\beta}\abs*{\Delta\Phi(\vecW)}+B.\]
Defining again $h(\vecW)=g(F(\vecW))+B>0$, we obtain that
\[ 1 \le \frac1{\beta} \cdot \frac{-\cL \Phi}{h} + \frac1{\beta} \cdot \frac{\abs*{\Delta \Phi}}{h} + \frac{B}{h}. \]
Rather than integrating this inequality everywhere, we integrate it only where this holds, in $\cS^c$. Multiplying the above by $f^2$ and integrating w.r.t. $\mu_{\beta}$ over $\cS^c$ gives
\begin{align*}
\int f^2 \DERIV \mubeta &= \int_{\cS} f^2 \DERIV \mubeta + \int_{\cS^c} f^2 \DERIV \mubeta  \\
&\le \int_{\cS} f^2 \DERIV \mubeta +\frac1{\beta}\int_{\cS^c} f^2 \frac{-\cL\Phi}{h} \DERIV\mu_{\beta} + \frac1{\beta}\int_{\cS^c} f^2 \frac{\abs*{\Delta\Phi}}{h} \DERIV\mu_{\beta} + \int_{\cS^c} f^2 \frac{B}{h} \DERIV\mu_{\beta} \\
&\le \frac1{\beta}\int f^2 \frac{-\cL\Phi}{h} \DERIV\mu_{\beta} + \frac1{\beta}\int f^2 \frac{\abs*{\Delta\Phi}}{h} \DERIV\mu_{\beta} + \int f^2 \frac{B}{h} \DERIV\mu_{\beta}  + \prn*{\int_{\cS} f^2 \DERIV \mubeta -\frac1{\beta}\int_{\cS} f^2 \frac{-\cL\Phi}{h} \DERIV\mu_{\beta}}\\
&\le \frac1{\beta}\int f^2 \frac{-\cL\Phi}{h} \DERIV\mu_{\beta} + \frac1{\beta}\int f^2 \frac{\abs*{\Delta\Phi}}{h} \DERIV\mu_{\beta} + \int f^2 \frac{B}{h} \DERIV\mu_{\beta}  + \prn*{\int_{\cS} f^2 \DERIV \mubeta + \frac1{\beta}\abs*{\int_{\cS} f^2 \frac{-\cL\Phi}{h} \DERIV\mu_{\beta}}}.
\end{align*}
The key difference is that the condition above not holding everywhere implies we picked up the `error term'
\[\int_{\cS} f^2 \DERIV \mubeta + \frac1{\beta}\abs*{\int_{\cS} f^2 \frac{-\cL\Phi}{h} \DERIV\mu_{\beta}},\]
which we wish to relate to $\text{osc}(\psi)$ to establish a WPI. 

Notice for $\beta=\Omega(d)$, $\frac1{\beta} \abs*{\frac{-\cL \Phi}{h}}$ can be controlled by a constant depending on problem-dependent parameters involving supremums over $\cS$ (which is typically thought of as small). 

Now we aim to see why $f^2$ can be related to $\text{osc}(\psi)^2$. Indeed, since $f=\psi-\alpha$ where $\psi$ is an expectation of $\psi$ w.r.t a probability measure, namely $\mubetalocal$, we obtain that $\abs*{f} \le \text{osc}(\psi)$ pointwise. Consequently we can upper bound the error term by 
\[ \int_{\cS} f^2 \DERIV \mubeta + \frac1{\beta}\abs*{\int_{\cS} f^2 \frac{-\cL\Phi}{h} \DERIV\mu_{\beta}} \le \text{problem dependent parameters} \cdot \text{osc}(\psi)^2 \cdot \mu_{\beta}(\cS).\]
Thus, rearranging the above, we obtain
\begin{align*}
&\int f^2 \cdot \frac{g(F(\vecW))}{h(\vecW)} \DERIV \mu_{\beta}\\ &\le \frac1{\beta}\int f^2 \frac{-\cL\Phi}{h} \DERIV\mu_{\beta} + \frac1{\beta}\int f^2 \frac{\abs*{\Delta\Phi}}{h} \DERIV\mu_{\beta} + \prn*{\int_{\cS} f^2 \DERIV \mubeta + \frac1{\beta}\abs*{\int_{\cS} f^2 \frac{-\cL\Phi}{h} \DERIV\mu_{\beta}}} \\
&\le \frac1{\beta}\int f^2 \frac{-\cL\Phi}{h} \DERIV\mu_{\beta} + \frac1{\beta}\int f^2 \frac{\abs*{\Delta\Phi}}{h} \DERIV\mu_{\beta} +\text{problem dependent parameters} \cdot \text{osc}(\psi)^2 \cdot \mu_{\beta}(\cS).
\end{align*}
From here, we proceed similarly to our proof of the PI from earlier. 
Finally, to prove \pref{corr:relaxingFunimodal}, rather than applying a PI for $\mubetalocal$ to upper bound $\int_{\cU} f^2 \DERIV \mu_{\beta}$ from \pref{ass:Funimodal}, apply the hypothesis that $\mubetalocal$ satisfies a WPI and use the same steps as above.

\section{Proofs}\label{sec:omittedproofs}
In all of these proofs, we define $\cU = \ball(\cW^{\star},r(l_b))$ for $l_b$ satisfying \pref{ass:Funimodal}, as done in \pref{sec:proofsketch}.
\subsection{Proof of \pref{thm:generalgflinearizabletopi}}\label{subsec:proofopttosampling}
We first introduce the following Lemma.
\begin{lemma}\label{lem:lyapunovmethodsetup}
Consider any Lyapunov function $\Phi(\cdot)$ and $g(\cdot)$ satisfying \pref{eq:gfcondition} for all $\vecW\in\mathbb{R}^d$. Then for any $B>0$ and any test function $f$, we have  
\begin{align*}
\int f(\vecW)^2 \frac{g\prn*{F(\vecW)}}{g\prn*{F(\vecW)}+B} \DERIV\mu_{\beta} &\le \frac1{\beta} \int \prn*{ \nrm*{\nabla f(\vecW)}^2+\frac{f(\vecW)^2}{h(\vecW)^2}\nrm*{\nabla \Phi(\vecW)}^2 - \frac{f(\vecW)^2}{h(\vecW)^2}\tri*{ \nabla h(\vecW), \nabla \Phi(\vecW)}} \DERIV\mu_{\beta} \\
&\hspace{1in}+ \frac1{\beta}\int f(\vecW)^2 \frac{\abs*{\Delta\Phi(\vecW)}}{h(\vecW)} \DERIV\mu_{\beta}.\numberthis\label{eq:provesamplingsetup}
\end{align*}
\end{lemma}
Note when we apply \pref{lem:lyapunovmethodsetup}, we will apply it with $\tilde{\Phi}$ in place of $\Phi$ and $\tilde{g}$ in place of $g$, where $\tilde{\Phi}, \tilde{g}$ are such that \pref{eq:gfcondition} holds with $\tilde{\Phi}$ in place of $\Phi$ and $\tilde{g}$ in place of $g$. 
\proof[Proof of \pref{lem:lyapunovmethodsetup}]{
By the condition \pref{eq:gfcondition}, we obtain
\[ g\prn*{F(\vecW)}+B \le \tri*{\grad \Phi(\vecW), F(\vecW)}+B = -\frac1{\beta}\cL\Phi(\vecW)+\frac1{\beta}\Delta\Phi(\vecW)+B\le -\frac1{\beta}\cL\Phi(\vecW)+\frac1{\beta}\abs*{\Delta\Phi(\vecW)}+B.\numberthis\label{eq:generatorconditionusage}\]
Denote $h(\vecW) := g\prn*{F(\vecW)}+B >0$. 
Therefore for any $f$, as $f^2 \ge 0$, we obtain
\begin{align*}
\int f(\vecW)^2 \DERIV\mu_{\beta} &\le \int f(\vecW)^2 \frac{-\frac1{\beta}\cL\Phi(\vecW)+\frac1{\beta}\abs*{\Delta\Phi(\vecW)}+B}{h(\vecW)} \DERIV\mu_{\beta} \\
&\le \frac1{\beta}\int f(\vecW)^2 \frac{-\cL\Phi(\vecW)}{h(\vecW)} \DERIV\mu_{\beta} + \frac1{\beta}\int f(\vecW)^2 \frac{\abs*{\Delta\Phi(\vecW)}}{h(\vecW)} \DERIV\mu_{\beta} + \int f(\vecW)^2 \frac{B}{h(\vecW)} \DERIV\mu_{\beta}.
\end{align*}
For the first term, we use \pref{lem:intbyparts} in the second equality below to obtain
\begin{align*}
\int f(\vecW)^2 \frac{-\cL\Phi(\vecW)}{h(\vecW)} \DERIV\mu_{\beta} &= \int \frac{f(\vecW)^2}{h(\vecW)} \cdot -\cL\Phi(\vecW) \DERIV\mu_{\beta}\\
&= \int \tri*{\nabla \prn*{\frac{f(\vecW)^2}{h(\vecW)}}, \nabla \Phi(\vecW)}\DERIV\mu_{\beta} \\
&= \int \prn*{\frac{2f(\vecW)}{h(\vecW)} \tri*{ \nabla f(\vecW), \nabla \Phi(\vecW)} - \frac{f(\vecW)^2}{h(\vecW)^2} \tri*{\nabla  h(\vecW), \nabla \Phi(\vecW)}} \DERIV\mu_{\beta}\\
&\le \int \prn*{2 \abs*{\frac{f(\vecW)}{h(\vecW)}} \nrm*{\grad f(\vecW)} \nrm*{\grad \Phi(\vecW)} - \frac{f(\vecW)^2}{h(\vecW)^2} \tri*{\nabla  h(\vecW), \nabla \Phi(\vecW)}} \DERIV\mu_{\beta}\\
&\le \int  \nrm*{\nabla f(\vecW)}^2+\frac{f(\vecW)^2}{h(\vecW)^2}\nrm*{\nabla \Phi(\vecW)}^2 - \frac{f(\vecW)^2}{h(\vecW)^2}\tri*{ \nabla h(\vecW), \nabla \Phi(\vecW)} \DERIV\mu_{\beta}.%\numberthis\label{eq:intbypartsineq}.
\end{align*}
Combining the above two inequalities and rearranging gives
\begin{align*}
\int f(\vecW)^2 \frac{g\prn*{F(\vecW)}}{g\prn*{F(\vecW)}+B} \DERIV\mu_{\beta} &\le \frac1{\beta} \int \prn*{ \nrm*{\nabla f(\vecW)}^2+\frac{f(\vecW)^2}{h(\vecW)^2}\nrm*{\nabla \Phi(\vecW)}^2 - \frac{f(\vecW)^2}{h(\vecW)^2}\tri*{ \nabla h(\vecW), \nabla \Phi(\vecW)}} \DERIV\mu_{\beta} \\
&\hspace{1in}+ \frac1{\beta}\int f(\vecW)^2 \frac{\abs*{\Delta\Phi(\vecW)}}{h(\vecW)} \DERIV\mu_{\beta},
\end{align*}
and this proves \pref{lem:lyapunovmethodsetup}. $\hfill\blacksquare$
}

Now, we prove \pref{thm:generalgflinearizabletopi}.
\proof[Proof of \pref{thm:generalgflinearizabletopi}]{
Our proof proceeds in three parts: 
\begin{itemize}
    \item Appropriately modifying $\Phi$ to make it more regular (which does \textit{not} require additional regularity assumptions beyond those stated in \pref{thm:generalgflinearizabletopi}).
    \item Using the Lyapunov function technique in a novel manner as sketched in \pref{sec:proofsketch} to prove a PI.
    \item Turning a PI into an LSI using established methods.
\end{itemize}
\paragraph{Part 1: Modifying $\Phi$ to introduce additional regularity.} 

The first part of our proof is to show we can create a smooth (bounded Hessian eigenvalues) Lyapunov function $\tilde{\Phi}$ with that satisfies \pref{eq:gfcondition}. The dependence on the allowed $\beta$ and the resulting isoperimetric constants will in turn depend on $\tilde{\Phi}$. We emphasize this step is \textit{only necessary when $\Phi$ is not smooth}.

First note without loss of generality we can take $m' \leftarrow \min\prn*{m',\frac12}$. Also note we can without loss of generality replace $g$ with a lower bound $\tilde{g}$ such that $\tilde{g}(0)=0$, $\tilde{g}(x)>0$ for $x>0$, is increasing, and has exactly linear tail growth. We do so by constructing $\tilde{g}$ as follows. First define
\[ x'=\frac1{m'}\prn*{g\prn*{r_2 R}+b'},\numberthis\label{eq:xprimedeftail}\]
and notice that $m' x' - b' =g(r_2 R)> 0$.
We now construct $\tilde{g}(\cdot)$ as follows:
\begin{itemize}
    \item If $r_2 R \ge x'$, define:
\[ \tilde{g}(x)=\begin{cases} \frac12 g(x)&\text{for } x \le r_2 R  \\\text{smoothed version}&\text{for }x \in [r_2 R, r_2 R+\delta] \\ m'x - b' &\text{for }x \ge r_2 R + \delta\end{cases}\]
for a small enough universal constant $\delta>0$. By `smoothed version' we just mean interpolating between the relevant two functions to preserve that $\tilde{g}(x)$ is differentiable and increasing while staying under the line $m'x-b'$, which we can easily see is possible because $m'x'-b' > \frac12 g(r_2 R) = \tilde{g}(r_2 R)$.
\item Otherwise if $r_2 R < x'$, define:
\[ \tilde{g}(x) =\begin{cases}\frac12 g(x)&\text{for }x \le r_2 R \\ \text{smoothed version }1&\text{for }x \in [r_2 R, r_2 R+\delta]  \\  \frac{\frac{9}{10}\prn*{m'x'-b'}-\frac34 g\prn*{r_2 R}}{x'-r_2 R}(x-r_2 R)+\frac34 g\prn*{r_2 R}&\text{for }x\in [r_2 R+\delta, x' - \delta]\\ \text{smoothed version }2&\text{for }x \in [x'-\delta,x']\\m'x-b'&\text{for }x \ge x'\end{cases}\]
for a small enough universal constant $\delta>0$. Similarly as before, by `smoothed version 1' we just mean interpolating between the relevant two functions to preserve that $\tilde{g}(x)$ is differentiable and increasing while staying under the line $\frac{\frac{9}{10}\prn*{m'x'-b'}-\frac34 g\prn*{r_2 R}}{x'-r_2 R}(x-r_2 R)+\frac34 g\prn*{r_2 R}$, and likewise by `smoothed version 2' we just mean interpolating between the relevant two functions to preserve that $\tilde{g}(x)$ is differentiable and increasing while staying under the line $m' x - b'$. This is possible because 1) $\frac12 g(r_2 R) < \frac34 g(r_2 R) < \frac{9}{10} g(r_2 R) < g(r_2 R)=m' x' - b'$, 2) $\frac{\frac{9}{10}\prn*{m'x'-b'}-\frac34 g\prn*{r_2 R}}{x'-r_2 R}(x'-r_2 R)+\frac34 g\prn*{r_2 R} = \frac{9}{10}(m'x'-b') = \frac{9}{10} g(r_2 R) < g(r_2 R)$, and 3) $\frac{9}{10}\prn*{m'x'-b'}-\frac34 g\prn*{r_2 R} = \frac3{20} g(r_2 R)>0$. In particular, 1), 2) and 3) ensure we can always interpolate so that $\tilde{g}$ is increasing, and 2) also ensures that $\tilde{g}(x) \le g(x)$.
\end{itemize}
Finally, take $\tilde{g}(x)\leftarrow r\tilde{g}(x)$ where 
\[ r = \min\prn*{1,\inf_{x\in [r_2 R, x']}\frac{x}{\tilde{g}(x)}}.\numberthis\label{eq:rdeftail}\]
Note $r>0$ since $g(r_2 R)>0$ and as $[r_2 R, x']$ is compact. These parameters also all behave in a dimension free way if $m',b', r_2, R$ do. 

In either case, the constructed $\tilde{g}(x)$ is increasing, differentiable, and has linear tail growth. In particular note $\tilde{g}(x) \ge r\prn*{m'(x-x')-b'}=m' rx-r(m'x'+b')$. Moreover, by this construction, we can check that for $x \ge r_2 R$ we have $\tilde{g}(x) \le x$, and for all $x \ge 0$ we have $g(x) \ge \tilde{g}(x)$. By \pref{ass:linearizableoutsideball}, for all $\vecW \in \ball(\vecW^{\star}, R)^c$ we have $F(\vecW) \ge r_2 R$, therefore 
\[ \tri*{\frac1{r_1}\prn*{\vecW-\vecW^{\star}}, \grad F(\vecW)} \ge F(\vecW) \ge \tilde{g}\prn*{F(\vecW)}\]
outside $\ball(\vecW^{\star}, R)$. Also, since for all $x$ we have $g(x) \ge \tilde{g}(x)$, this implies for all $\vecW$ that
\[ \tri*{\grad \Phi(\vecW), \grad F(\vecW)} \ge g\prn*{F(\vecW)} \ge \tilde{g}\prn*{F(\vecW)}.\]
Let $\Phi_2(\vecW) = \frac{1}{2r_1} \nrm*{\vecW-\vecW^{\star}}^2+M'$ where 
\[ M' :=\sup_{\vecW \in \ball(\vecW^{\star}, R+1)} \Phi(\vecW).\numberthis\label{eq:Mprimedef}\]
Therefore, we have $\tri*{\grad\Phi_2(\vecW), \grad F(\vecW)} \ge \tilde{g}\prn*{F(\vecW)}$ outside $\ball(\vecW^{\star}, R)$, and also that $\Phi_2(\vecW) \ge \Phi(\vecW)$ on $\ball(\vecW^{\star}, R+1)$. Furthermore, note the above construction of $\tilde{g}(x)$ is unnecessary if $g(x)=\lambda x$, by taking $\lambda = \min(\lambda, 1)$, which is the case in many of our examples e.g. \pref{ex:formalPLexample}, \pref{ex:formalLinearizableexample}.

From here on out, if $g(x)=\lambda x$ for $\lambda \le 1$ we define
\[ \mprimenew=m', \bprimenew=b'.\numberthis\label{eq:tailprimesdef1}\]
Otherwise if the above construction of $\tilde{g}$ was needed we define
\[ \mprimenew = m' r, \bprimenew = r(m'x'+b'),\numberthis\label{eq:tailprimesdef2}\]
where $r, x'$ are defined as per \pref{eq:rdeftail}, \pref{eq:xprimedeftail}. Consequently we always have 
\[ \tilde{g}(x) \ge \mprimenew x - \bprimenew.\numberthis\label{eq:tildegtailgrowth}\]
Now, we let $\chi(\vecW) \in [0,1]$ be a bump function interpolating between $\ball(\vecW^{\star}, R)$ and $\ball(\vecW^{\star}, R+1)$ in the natural way, such that $\chi \equiv 0$ on $\ball(\vecW^{\star}, R)$ and $\chi \equiv 1$ on $\ball(\vecW^{\star}, R+1)^c$. In \pref{lem:constructsuchachi}, we explicitly construct a $\chi(\vecW)$ such that:
\begin{itemize}
    \item $\chi(\vecW)$ is differentiable to all orders.
    \item $\nrm*{\grad \chi(\vecW)}, \nrm*{\grad^2 \chi(\vecW)}_{\OPNORM} \le B$ where $B>0$ is a universal constant.
    %for some function $\psi$ with no explicit $d$ dependence for $\vecW \in \ball(\vecW^{\star}, R)^c \cap \ball(\vecW^{\star}, R+1)$.
    \item $\tri*{\grad \chi(\vecW), \grad F(\vecW)} \ge 0$ for $\vecW \in \ball(\vecW^{\star}, R)^c \cap \ball(\vecW^{\star}, R+1)$.
\end{itemize}
Now, define 
\[ \tilde{\Phi}(\vecW) := \chi(\vecW) \Phi_2(\vecW) +(1-\chi(\vecW))\Phi(\vecW).\]
We break into cases and show that $\tilde{\Phi}$ is still a valid Lyapunov function.
\begin{itemize}
    \item For $\vecW \in \ball(\vecW^{\star}, R)$, as $\chi \equiv 0$ holds identically in this set, we have 
\[ \tri*{\grad \tilde{\Phi}(\vecW), \grad F(\vecW)} \equiv \tri*{\grad \Phi(\vecW), \grad F(\vecW)} \ge \tilde{g}\prn*{F(\vecW)}.\]
\item For $\vecW \in \ball(\vecW^{\star}, R+1)^c$, as $\chi \equiv 1$ identically in this set, we have 
\[ \tri*{\grad \tilde{\Phi}(\vecW), \grad F(\vecW)} = \tri*{\grad \Phi_2(\vecW), \grad F(\vecW)} \ge \tilde{g}\prn*{F(\vecW)}.\]
\item For $\vecW \in \ball(\vecW^{\star}, R)^c \cap \ball(\vecW^{\star}, R+1)$, we have 
\[ \grad \tilde{\Phi}(\vecW)=\chi(\vecW) \grad \Phi_2(\vecW)+(1-\chi(\vecW))\grad\Phi(\vecW)+\grad \chi(\vecW) \Phi_2(\vecW) - \grad \chi(\vecW) \Phi(\vecW).\]
This means
\begin{align*}
\tri*{\grad \tilde{\Phi}(\vecW), \grad F(\vecW)} &= \chi(\vecW)\tri*{\grad \Phi_2(\vecW), \grad F(\vecW)}+(1-\chi(\vecW))\tri*{\grad \Phi(\vecW), \grad F(\vecW)} \\
&\hspace{1in}+(\Phi_2(\vecW)-\Phi(\vecW)) \tri*{\grad \chi(\vecW), \grad F(\vecW)} \\
&\ge (\chi(\vecW)+1-\chi(\vecW))\tilde{g}\prn*{F(\vecW)}=\tilde{g}\prn*{F(\vecW)}.
\end{align*}
The above uses that $\Phi_2(\vecW) \ge \Phi(\vecW)$ for $\vecW\in \ball(\vecW^{\star}, R+1)$, and the property of $\chi$ that $\tri*{\grad \chi(\vecW), \grad F(\vecW)} \ge 0$ for $\vecW \in \ball(\vecW^{\star}, R)^c \cap \ball(\vecW^{\star}, R+1)$.
\end{itemize}
Therefore, for all $\vecW\in\mathbb{R}^d$ we have
\[ \tri*{\grad \tilde{\Phi}(\vecW), \grad F(\vecW)} \ge \tilde{g}\prn*{F(\vecW)}.\]%\numberthis\label{eq:interpolatedphigfcondition}\]
That is, $\tilde{\Phi}(\cdot)$, together with $\tilde{g}(\cdot)$, satisfies \pref{eq:gfcondition}. 

Moreover, we claim $\tilde{\Phi}$ is smooth. Note $\nrm*{\grad^2 \Phi_2(\vecW)}_{\OPNORM}=\frac{1}{r_1}$ where $r_1$ was defined above. Let 
\[ L' = \sup_{\vecW\in \ball(\vecW^{\star}, R+1)} \rho_{\Phi}\prn*{\Phi(\vecW)} \le \rho_{\Phi}(M'),\numberthis\label{eq:Lprimedef}\]
where $M'$ is as in \pref{eq:Mprimedef}.
%Let $M' = \sup_{\vecW \in \ball(\vecW^{\star}, R+1)} \Phi\prn*{\vecW}$. 
\begin{itemize}
    \item In $\ball(\vecW^{\star}, R) \cup \ball(\vecW^{\star}, R+1)^c$ we have $\nrm*{\grad^2 \tilde{\Phi}(\vecW)}_{\OPNORM} \le \max\prn*{L', \frac{1}{r_1}}$. 
    \item In $\ball(\vecW^{\star}, R)^c \cap \ball(\vecW^{\star}, R+1)$, we can compute
\begin{align*}
\grad^2 \tilde{\Phi}(\vecW) &= \grad^2 \Phi(\vecW) + \prn*{\Phi_2(\vecW)-\Phi(\vecW)} \grad^2 \chi(\vecW) + \grad^2 \prn*{\Phi_2(\vecW)-\Phi(\vecW)} \chi(\vecW) \\
&\hspace{1in}+ 2\grad \chi(\vecW) \grad \prn*{\Phi_2(\vecW)-\Phi(\vecW)}^T.
\end{align*}
By Triangle Inequality for operator norm and the inequality $\nrm*{\mathbf{a}\mathbf{b}^T}_{\OPNORM} \le \nrm*{\mathbf{a}}\nrm*{\mathbf{b}}$, it follows that
\begin{align*}
&\nrm*{\grad^2 \tilde{\Phi}(\vecW)}_{\OPNORM} \\
&\le \nrm*{\grad^2 \Phi(\vecW)}_{\OPNORM} + \prn*{\abs*{\Phi_2(\vecW)}+\abs*{\Phi(\vecW)}} \nrm*{\grad^2 \chi(\vecW)}_{\OPNORM} + \prn*{\nrm*{\grad^2 \Phi_2(\vecW)}_{\OPNORM}+\nrm*{\grad^2\Phi(\vecW)}_{\OPNORM}} \chi(\vecW) \\
&\hspace{1in}+ 2\nrm*{\grad\chi(\vecW)} \nrm*{\grad\prn*{\Phi_2(\vecW)-\Phi(\vecW)}} \\
&\le L' + B\prn*{\frac{(R+1)^2}{2r_1} + 2M'}+\prn*{\frac1{r_1}+L'}\cdot 1+2B\prn*{L'+\frac{R+1}{r_1}}.
\end{align*}
\end{itemize}
Recalling $L'$ from \pref{eq:Lprimedef}, define 
\[ \tilde{L} := \crl*{L' + B\prn*{\frac{(R+1)^2}{2r_1} + 2M'}+\prn*{\frac1{r_1}+L'}+2B\prn*{L'+\frac{R+1}{r_1}}} \lor 2\bprimenew \lor 1, \numberthis\label{eq:tildeLdefpi}\] 
where $\bprimenew$ defines the linear univariate tail growth of $\tilde{g}$. Here, we recall the definitions of $L'$ in \pref{eq:Lprimedef}, $M'$ in \pref{eq:Mprimedef}, $\bprimenew$ from \pref{eq:tailprimesdef1} or \pref{eq:tailprimesdef2} (whichever applies here), and $B$ which is a universal constant coming from the construction of $\chi$. Thus, $\tilde{\Phi}$ is $\tilde{L}$-smooth. Clearly $\tilde{\Phi}$ is non-negative as well.\\

\paragraph{Part 2: Proving a PI with the new Lyapunov function.} 
Now we go back to our setup to prove a Poincar\'e Inequality. 
Consider any test function $\psi$. Let 
\[ f=\psi-\alpha\text{ where }\alpha=\frac1{\mu_{\beta}(\cU)}\int_{\cU} \psi \DERIV\mu_{\beta}.\numberthis\label{eq:fdeflyapunov}\]
Recall that $\tilde{\Phi}(\cdot)$ together with $\tilde{g}(\cdot)$ satisfies \pref{eq:gfcondition}. Thus, applying \pref{lem:lyapunovmethodsetup} with $\tilde{\Phi}(\cdot)$, $\tilde{g}(\cdot)$, $B = \tilde{L}>0$ and $h(\vecW)=\tilde{g}\prn*{F(\vecW)}+\tilde{L}$ gives
\begin{align*}
\int f(\vecW)^2 \frac{\tilde{g}\prn*{F(\vecW)}}{\tilde{g}\prn*{F(\vecW)}+\tilde{L}} \DERIV\mu_{\beta} &\le \frac1{\beta} \int \prn*{ \nrm*{\nabla f(\vecW)}^2+\frac{f(\vecW)^2}{h(\vecW)^2}\nrm*{\nabla \tilde{\Phi}(\vecW)}^2 - \frac{f(\vecW)^2}{h(\vecW)^2}\tri*{ \nabla h(\vecW), \nabla \tilde{\Phi}(\vecW)}} \DERIV\mu_{\beta} \\
&\hspace{1in}+ \frac1{\beta}\int f(\vecW)^2 \frac{\abs*{\Delta\tilde{\Phi}(\vecW)}}{h(\vecW)} \DERIV\mu_{\beta}.\numberthis\label{eq:provesamplingsetuppipf}
\end{align*}

\subparagraph{Step a: Upper bounding relevant terms using the construction of $\tilde{\Phi}$.} 
We aim to upper bound the intermediate term in \pref{eq:provesamplingsetuppipf}. Observe as $\tilde{g}$ is non-decreasing and non-negative,
\[ \tri*{\grad h(\vecW), \grad \tilde{\Phi}(\vecW)} = \tilde{g}'\prn*{F(\vecW)}\tri*{\grad F(\vecW), \grad \tilde{\Phi}(\vecW)} \ge \tilde{g}'\prn*{F(\vecW)}\tilde{g}\prn*{F(\vecW)} \ge 0.\]
Also observe by $\tilde{L}$-smoothness of $\tilde{\Phi}$ and using \pref{lem:smoothnessgradbound}, because $\chi \in [0,1]$ and by definition of $M'$,
\[ \nrm*{\grad \tilde{\Phi}(\vecW)}^2 \le 4\tilde{L}\tilde{\Phi}(\vecW) \le 4\tilde{L} \prn*{M'+\Phi_2(\vecW)} = 4\tilde{L}\prn*{2M'+\frac1{2r_1}\nrm*{\vecW-\vecW^{\star}}^2}.\]
Therefore, as $g(x)\ge0$, using the above implies
\begin{align*}
\frac{\nrm*{\grad \tilde{\Phi}(\vecW)}^2-\tri*{\grad h(\vecW), \grad \tilde{\Phi}(\vecW)}}{h(\vecW)^2} \le \frac{\nrm*{\grad \tilde{\Phi}(\vecW)}^2}{h(\vecW)^2} \le \frac{4\tilde{L}\prn*{2M'+\frac1{2r_1}\nrm*{\vecW-\vecW^{\star}}^2}}{h(\vecW)^2}.
\end{align*}
Furthermore recall that because $\tilde{g}(x) \ge \max(0,\mprimenew x - \bprimenew)$, we have
\[ h(\vecW) \ge \max\prn*{\tilde{L},\mprimenew F(\vecW)-\bprimenew+\tilde{L}}.\]
\begin{itemize}
\item If $\vecW \in \ball(\vecW^{\star}, R)$, using $\tilde{L}/2 \ge \bprimenew$, the above is clearly at most $\frac{8\prn*{R^2+4M'r_1}}{r_1\tilde{L}}$. 
\item Otherwise, using the second part of \pref{ass:linearizableoutsideball} and $\tilde{L}/2 \ge \bprimenew$, we have 
\[ \frac{4\tilde{L}\prn*{2M'+\frac1{2r_1}\nrm*{\vecW-\vecW^{\star}}^2}}{\prn*{\mprimenew F(\vecW) - \bprimenew+\tilde{L}}^2} \le 4\tilde{L} \cdot \frac{\frac1{2r_1}\nrm*{\vecW-\vecW^{\star}}^2+2M'}{r_2^2 \mprimenew^2 \nrm*{\vecW-\vecW^{\star}}^2+\frac{\tilde{L}^2}4} \le \frac{2\tilde{L}}{r_1 r_2^2 \mprimenew^2} \lor \frac{32M'}{\tilde{L}}.\]
The last line uses the simple fact that $\frac{ta+b}{tc+d} \le \frac{a}c \lor \frac{b}d$ for all $t, a, b, c, d \ge 0$.
\end{itemize}
Define 
\[ C' := \frac{8\prn*{R^2+4M'r_1}}{r_1\tilde{L}} \lor \frac{2\tilde{L}}{r_1 r_2^2 \mprimenew^2} \lor \frac{32M'}{\tilde{L}}.\numberthis\label{eq:Cprimedefpi} \]
Here $M'$ is from \pref{eq:Mprimedef}, $\tilde{L}$ is from \pref{eq:tildeLdefpi}, and $\mprimenew$ is from \pref{eq:tailprimesdef1} or \pref{eq:tailprimesdef2} (whichever case applies here).
Consequently the above proves that for any $f$, letting $h(\vecW)=\tilde{g}\prn*{F(\vecW)}+\tilde{L}$, we have
\[ \frac{\nrm*{\grad \tilde{\Phi}(\vecW)}^2-\tri*{\grad h(\vecW), \grad \tilde{\Phi}(\vecW)}}{h(\vecW)^2} \le C'.\numberthis\label{eq:upperboundintermediateterm} \]

\subparagraph{Step b: Using the Lyapunov method.}
Applying \pref{eq:upperboundintermediateterm} in \pref{eq:provesamplingsetuppipf} and using $\tilde{L}$-smoothness of $\tilde{\Phi}$ and that $f^2 \ge 0$, we now have
\begin{align*}
&\int f(\vecW)^2 \frac{\tilde{g}\prn*{F(\vecW)}}{\tilde{g}\prn*{F(\vecW)}+\tilde{L}} \DERIV \mu_{\beta} \\
&\le \frac1{\beta} \int \nrm*{\grad f(\vecW)}^2 \DERIV \mu_{\beta}+\frac1{\beta}\int C' f(\vecW)^2 \DERIV \mu_{\beta}+\frac1{\beta}\int f(\vecW)^2 \frac{\abs*{\Delta \tilde{\Phi}(\vecW)}}{\tilde{g}(F(\vecW))+L} \DERIV \mu_{\beta} \\
&\le \frac1{\beta}\int \nrm*{\grad f(\vecW)}^2 \DERIV \mu_{\beta} + \frac1{\beta}\int C' f(\vecW)^2 \DERIV \mu_{\beta}+ \frac1{\beta} \int f(\vecW)^2 \frac{d\tilde{L}}{\tilde{g}\prn*{F(\vecW)}+\tilde{L}} \DERIV \mu_{\beta} \\
&\le \frac1{\beta}\int \nrm*{\grad f(\vecW)}^2 \DERIV \mu_{\beta} + \frac1{\beta}\int f(\vecW)^2 \prn*{d+C'} \DERIV \mu_{\beta}.
\end{align*}
Notice $\frac{\tilde{g}(t)}{\tilde{g}(t)+\tilde{L}}$ is non-decreasing as $\tilde{g}$ is non-decreasing. 
We thus obtain:
\begin{align*}
&\int f(\vecW)^2 \frac{\tilde{g}(l_b)}{\tilde{g}(l_b)+\tilde{L}} \DERIV \mu_{\beta} \\
&= \int_{\cU^c} f(\vecW)^2 \frac{\tilde{g}(l_b)}{\tilde{g}(l_b)+\tilde{L}} \DERIV \mu_{\beta}+\int_{\cU} f(\vecW)^2 \frac{\tilde{g}(l_b)}{\tilde{g}(l_b)+\tilde{L}} \DERIV \mu_{\beta} \\
&\le \int f(\vecW)^2 \frac{\tilde{g}(F(\vecW))}{\tilde{g}\prn*{F(\vecW)}+\tilde{L}} \DERIV \mu_{\beta} + \int_{\cU} f(\vecW)^2 \frac{\tilde{g}(l_b)}{\tilde{g}(l_b)+\tilde{L}} \DERIV \mu_{\beta} \\
&\le \frac1{\beta}\int \nrm*{\grad f(\vecW)}^2 \DERIV \mu_{\beta} + \frac1{\beta}\int f(\vecW)^2 \prn*{d+C'} \DERIV \mu_{\beta} + \frac{\tilde{g}(l_b)}{\tilde{g}(l_b)+\tilde{L}}\int_{\cU} f(\vecW)^2 \DERIV \mu_{\beta}. \numberthis\label{eq:provePIbeforetrick} \end{align*}
We now upper bound $\int_{\cU} f(\vecW)^2 \DERIV \mu_{\beta}$. As $\mu_{\beta, \textsc{local}} := \mu_{\beta, \textsc{local}}(l_b)$ satisfies a Poincar\'e Inequality by \pref{ass:Funimodal}, we have
\[ \mathbb{V}_{\mu_{\beta, \textsc{local}}}\prn*{f} \le \CPILOCAL \int \nrm*{\grad f(\vecW)}^2 \DERIV \mu_{\beta, \textsc{local}}.\]
Using definition of variance and $\mu_{\beta,\textsc{local}}$ in the above, we obtain that 
\[ \frac1{\mu_{\beta}(\cU)}\int_{\cU} f(\vecW)^2 \DERIV \mubeta - \frac1{\mu_{\beta}(\cU)^2} \prn*{\int_{\cU} f(\vecW) \DERIV \mubeta}^2 \le \CPILOCAL \cdot \frac1{\mu_{\beta}(\cU)} \int_{\cU} \nrm*{\grad f(\vecW)}^2 \DERIV \mubeta.\]
Recalling the definition of $f=\psi-\alpha$ for $\alpha=\frac1{\mu_{\beta}(\cU)}\int_{\cU} \psi \DERIV\mu_{\beta}$, we obtain from the above that
\begin{align*}
&\int_{\cU} f(\vecW)^2 \DERIV\mu_{\beta} \le \CPILOCAL\int_{\cU}\nrm*{\nabla f(\vecW)}^2 \DERIV\mu_{\beta}+\frac1{\mu_{\beta}\prn*{\cU}} \prn*{\int_{\cU} f(\vecW) \DERIV \mu_{\beta}}^2 \\
&\le \CPILOCAL\int\nrm*{\nabla f(\vecW)}^2 \DERIV\mu_{\beta}+\frac1{\mu_{\beta}\prn*{\cU}} \prn*{\int_{\cU}  \prn*{\psi(\vecW) -\frac1{\mu_{\beta}(\cU)}\int_{\cU} \psi(\vecW)\DERIV \mu_{\beta}}\DERIV \mu_{\beta}}^2 \\
&= \CPILOCAL\int\nrm*{\nabla f(\vecW)}^2 \DERIV\mu_{\beta}+0.
\end{align*}
Applying this in \pref{eq:provePIbeforetrick}, we obtain
\begin{align*}
\int f(\vecW)^2 \frac{\tilde{g}(l_b)}{\tilde{g}(l_b)+\tilde{L}} \DERIV \mu_{\beta}&\le \frac1{\beta}\int \nrm*{\grad f(\vecW)}^2 \DERIV \mu_{\beta} + \frac1{\beta}\int f(\vecW)^2 \prn*{d+C'} \DERIV \mu_{\beta} \\
&\hspace{1in}+ \frac{\tilde{g}(l_b)}{\tilde{g}(l_b)+\tilde{L}} \cdot \CPILOCAL \int \nrm*{\grad f(\vecW)}^2 \DERIV \mu_{\beta}.
\end{align*}
For $\beta \ge 2\prn*{1+\frac{\tilde{L}}{\tilde{g}(l_b)}} (d+C')=\Omega(d)$, this gives 
\[ \frac{\tilde{g}(l_b)}{2(\tilde{g}(l_b)+\tilde{L})}\int f(\vecW)^2 \DERIV \mu_{\beta} \le \frac1{\beta}\int \nrm*{\grad f(\vecW)}^2 \DERIV \mu_{\beta} + \frac{\tilde{g}(l_b)}{\tilde{g}(l_b)+\tilde{L}} \cdot \CPILOCAL \int \nrm*{\grad f(\vecW)}^2 \DERIV \mu_{\beta}.\]
Rearranging this inequality and converting back to $\psi$, recalling the definition of variance, and noting $\grad f = \grad \psi$ gives:
\begin{align*}
\mathbb{V}_{\mu_{\beta}}\brk*{\psi} &\le \int \prn*{\psi-\alpha}^2 \DERIV \mu_{\beta} \\
&= \int f^2 \DERIV \mu_{\beta} \\
&\le \prn*{2\CPILOCAL + \frac{2}{\beta}\prn*{1+\frac{\tilde{L}}{\tilde{g}(l_b)}}} \int \nrm*{\grad f}^2 \DERIV \mu_{\beta}\\
&=\prn*{2\CPILOCAL + \frac{2}{\beta}\prn*{1+\frac{\tilde{L}}{\tilde{g}(l_b)}}} \int \nrm*{\grad \psi}^2 \DERIV \mu_{\beta}.
\end{align*}
Recalling $\psi$ is an arbitrary test function, this shows that $\mu_{\beta}$ satisfies a Poincar\'e Inequality with a Poincar\'e constant of 
\[ 2\CPILOCAL + \frac{2}{\beta}\prn*{1+\frac{\tilde{L}}{\tilde{g}(l_b)}} \text{ for }\beta \ge 2\prn*{1+\frac{\tilde{L}}{\tilde{g}(l_b)}} (d+C'),\numberthis\label{eq:piexplicit}\]
where $\tilde{L}$ is defined in \pref{eq:tildeLdefpi} and $C'$ is defined in \pref{eq:Cprimedefpi}.
%Notice upon converting back to the settings of \pref{thm:plimpliespi}, \pref{thm:klimpliespi}, this result gives the same Poincar\'e constant as we obtained there.

\paragraph{Part 3: Proving a Log-Sobolev Inequality.}
With the above PI in hand, and under the relevant conditions given in \pref{thm:generalgflinearizabletopi}, we use the following result of \citet{cattiaux2010note}  in the form given by Proposition 15 from \citet{raginsky2017non} to prove an LSI.
\begin{theorem}[Proposition 15, \citet{raginsky2017non}]\label{thm:lsivialyapunov}
Suppose the following conditions hold:
\begin{enumerate}
\item There exists constants $\kappa,\gamma>0$ and a twice continuously differentiable function $V:\mathbb{R}^d\rightarrow[1,\infty)$ such that for all $\vecW\in\mathbb{R}^d$,  
\[ \frac{\cL V(\vecW)}{V(\vecW)} \le \kappa-\gamma \nrm*{\vecW}^2.\]
\item $\mu_{\beta}$ satisfies a Poincar\'e Inequality with constant $\CPI$.
\item There exists some constant $K \ge 0$ such that $\grad^2 F \succeq -K$.
\end{enumerate} 
Then, for any \(\delta > 0\). $\mu_{\beta}$ satisfies a Log-Sobolev Inequality with  $\CLSI=C_1+(C_2+2)\CPI$, where 
\begin{align*}
 C_1 \ldef{} \frac{2}{\gamma}\prn*{\frac1{\delta}+\frac{\beta K}2}+\delta \quad \text{and} \quad C_2 \ldef{} \frac{2}{\gamma}\prn*{\frac1{\delta}+\frac{\beta K}2}\prn*{\kappa+\gamma\int_{\mathbb{R}^d}\nrm*{\vecW}^2\DERIV \mu_{\beta}}.
\end{align*}
\end{theorem}
Use $V(\vecW) = e^{\tilde{\Phi}(\vecW)}$ in \pref{thm:lsivialyapunov}. 
Condition 2 in \pref{thm:lsivialyapunov} follows from the above part, and condition 3 in \pref{thm:lsivialyapunov} is trivially satisfied with $K=L$ by our condition on weak convexity of $F$. For condition 1, let $V(\vecW)=e^{\tilde{\Phi}(\vecW)} \ge 1$. Compute
\[ \grad V(\vecW) = e^{\tilde{\Phi}(\vecW)} \grad \tilde{\Phi}(\vecW), \Delta \tilde{\Phi}(\vecW) = e^{\tilde{\Phi}(\vecW)}\prn*{\Delta \tilde{\Phi}(\vecW)+\nrm*{\grad \tilde{\Phi}(\vecW)}^2}. \]
Therefore, 
\begin{align*}
\frac{\cL V(\vecW)}{V(\vecW)} &= \frac{V(\vecW)\prn*{\Delta \tilde{\Phi}(\vecW)+\nrm*{\grad \tilde{\Phi}(\vecW)}^2 - \tri*{\beta\grad F(\vecW), \grad \tilde{\Phi}(\vecW)}}}{V(\vecW)} \\
&= \Delta \tilde{\Phi}(\vecW)+\nrm*{\grad \tilde{\Phi}(\vecW)}^2 - \tri*{\beta\grad F(\vecW), \grad \tilde{\Phi}(\vecW)}.
\end{align*}
We now upper bound the above. Recall we showed $\tilde{\Phi}(\vecW)$ is $\tilde{L}$ smooth, hence $\Delta \tilde{\Phi}(\vecW) \le d\tilde{L}$. Now we break into cases:
\begin{itemize}
\item Consider $\vecW\in\ball(\vecW^{\star}, R+1)$. Recall for such $\vecW$, $\nrm*{\grad \tilde{\Phi}(\vecW)} \le L'$. Also recall $\tri*{\grad F(\vecW), \grad \tilde{\Phi}(\vecW)} \ge \tilde{g}(F(\vecW)) \ge 0$. Thus in this case 
\[ \frac{\cL V(\vecW)}{V(\vecW)} \le d\tilde{L}+L'.\]
\item Consider $\vecW\in\ball(\vecW^{\star}, R+1)^c$. Now, $\nrm*{\grad \tilde{\Phi}(\vecW)} = \frac1{r_1}\nrm*{\vecW-\vecW^{\star}}$. Also recall $\tri*{\grad F(\vecW), \grad \tilde{\Phi}(\vecW)} \ge \tilde{g}(F(\vecW))$. 
By construction of $\tilde{g}$, we have $\tilde{g}(x) \ge \mprimenew x - \bprimenew$ (recall \pref{eq:tildegtailgrowth}). Hence, by conditions on the growth of $F$ in this part, we obtain
\[ \tri*{\grad F(\vecW), \grad \tilde{\Phi}(\vecW)} \ge \tilde{g}(F(\vecW)) \ge \mprimenew (m\nrm*{\vecW}^2-b) - \bprimenew = m \mprimenew \nrm*{\vecW}^2 - (b \mprimenew + \bprimenew).\]
Thus in this case, we have 
\[ \frac{\mathcal{L} V(\vecW)}{V(\vecW)} \le d\tilde{L}+\frac1{r_1^2} \nrm*{\vecW-\vecW^{\star}}^2 - \beta\prn*{ m \mprimenew\nrm*{\vecW}^2-(b \mprimenew + \bprimenew)}. \]
\end{itemize}
Doing casework based on the above cases and with one application of Young's Inequality, we see that when $\beta \ge \frac{4}{r_1^2 m}$, condition 1 is of \pref{thm:lsivialyapunov} is satisfied with
\begin{align*}
\kappa &= d\tilde{L}+L'+\frac{2}{r_1^2}\nrm*{\vecW^{\star}}^2 + \beta (b \mprimenew + \bprimenew) + \frac{\beta m \mprimenew}2(R+1)^2, \\
\gamma &= \frac{\beta m \mprimenew}2.
\end{align*}
Choose $\delta=\frac1{\sqrt{\gamma}}$. As $\beta \ge 2$, we can check
\begin{align*}
C_1 &= \frac4{m \mprimenew \beta} \prn*{\sqrt{\frac{\beta m \mprimenew}2}+\frac{\beta L}2} + \sqrt{\frac{2}{\beta m \mprimenew}} \le \frac{4L+3}{2m \mprimenew} + \frac32,\\
C_2 &=2\prn*{\sqrt{\gamma}+\frac{\beta L}2}\prn*{\frac{\kappa}{\gamma}+S} \\
&\le 2\prn*{\sqrt{\frac{\beta m \mprimenew}2}+\frac{\beta L}2} \\
&\hspace{1in}\cdot \prn*{(R+1)^2+\frac{2(\bprimenew+b\mprimenew)}{m \mprimenew}+\frac{4}{\beta m \mprimenew r_1^2}\nrm*{\vecW^{\star}}^2+\frac{2\prn*{d\tilde{L}+L'}}{\beta m \mprimenew}+S}.
\end{align*}
Using $\beta \ge 2$, and our earlier upper bound on $\CPI$, this yields a Log-Sobolev constant of 
\begin{align*}
\CLSI &\le C_1+(C_2+2)\CPI \\
&\le \frac{4L+3}{2m \mprimenew} + \frac32 \\
&+4\prn*{1+\crl*{L+\sqrt{m \mprimenew}} \cdot \crl*{(R+1)^2+2\prn*{\frac{\bprimenew}{m \mprimenew} + \frac{b}m}+\frac{4}{\beta m \mprimenew r_1^2}\nrm*{\vecW^{\star}}^2+\frac{2\prn*{d\tilde{L}+L'}}{\beta m \mprimenew}+S}} \\
&\hspace{1in} \cdot \prn*{\crl*{1+\frac{\tilde{L}}{ \tilde{g}(l_b)}}+\beta\CPILOCAL},\numberthis\label{eq:lsiexplicit}
\end{align*}
for $\beta \ge 2\prn*{1+\frac{\tilde{L}}{\tilde{g}(l_b)}} (d+C') \lor \frac{4}{r_1^2 m}\ge 2$.
Again, in the above, $\tilde{L}$ comes from \pref{eq:tildeLdefpi}, $C'$ comes from \pref{eq:Cprimedefpi}, $L'$ comes from \pref{eq:Lprimedef}, and $\mprimenew, \bprimenew$ are as per \pref{eq:tailprimesdef1} or \pref{eq:tailprimesdef2}, whichever case is appropriate. This proves the desired LSI.
$\hfill\blacksquare$
}

\begin{remark}\label{rem:noFregularity}
Notice in the above proof, we did not use \pref{ass:phiselfbounding} on $F$, hence the statement of \pref{thm:generalgflinearizabletopi}. Note also that this proof establishes a PI from optimizability almost everywhere (w.r.t. Lebesgue measure $\nu$), since $\mu$ is absolutely continuous with respect to $\nu$.
\end{remark}
\begin{remark}\label{rem:linearizableboost}
We note when $\Phi$ is $L$-smooth to begin with (for example, $L=2$ when $\Phi(\vecW)=\nrm*{\vecW-\vecW^{\star}}^2$, which holds in the Linearizable example \pref{ex:formalLinearizableexample}), the construction of $\tilde{g}$ and $\tilde{\Phi}$ is unnecessary. We can just use $\Phi$ instead of $\tilde{\Phi}$, and in the above guarantees from \pref{eq:piexplicit}, \pref{eq:lsiexplicit}, using \pref{lem:smoothnessgradbound} we have 
\[ \tilde{L}=L \lor 2b', M'=0, C' = \frac{8R^2}{\min(1/2, r_1)\tilde{L}} \lor \frac{2\tilde{L}}{\min(1/2, r_1) r_2^2 \mprimenew^2}.\numberthis\label{eq:simplifiedexpressionslinearizable}\]
For example, in this case we obtain
$\mu_{\beta}$ satisfies a Poincar\'e Inequality with a Poincar\'e constant of 
\[ \CPI = 2\CPILOCAL + \frac{2}{\beta}\prn*{1+\frac{L \lor 2b'}{g(l_b)}} \text{ for }\beta \ge 2\prn*{1+\frac{L \lor 2b'}{g(l_b)}} \prn*{d+\frac{8R^2}{\min(1/2, r_1)\tilde{L}} \lor \frac{2\tilde{L}}{\min(1/2, r_1) r_2^2 \mprimenew^2}}. \]
We similarly obtain a cleaner and tighter bound for $\CLSI$ plugging the expressions from \pref{eq:simplifiedexpressionslinearizable} back into \pref{eq:lsiexplicit}. Also note the construction of $\tilde{g}$ is unnecessary if $g(x)=\lambda x$ for $\lambda \le 1$; if this is the case, we can just take $\mprimenew = \lambda$, $\bprimenew = 0$.
\end{remark}
\begin{remark}\label{rem:localconditionisall}
By tracking the proof, we see that if \pref{ass:linearizableoutsideball} holds, it suffices to have $\Phi, F, g$ satisfy \pref{eq:gfcondition} inside $\ball(\vecW^{\star}, R+1)$. This is because in our construction of $\tilde{g}$, we did not change $R$. After this, in our construction of $\tilde{\Phi}$, we only need the condition from \pref{ass:linearizableoutsideball} outside $\ball(\vecW^{\star}, R+1)$. After our construction of $\tilde{\Phi}$, the condition \pref{eq:gfcondition} is no longer used in the proof.
\end{remark}
\begin{remark}\label{rem:ourimprovementonlyapunovfull}
Consider a canonical example of non-convex, optimizable $F$: when $F$ is $\lambda$-Linearizable \citep{sekhari2021sgd, kleinberg2018alternative, priorpaper, hinder2020near}. For simplicity say $\lambda \le 1$. Thus \pref{def:optimizabilitydef} holds with $\Phi=\nrm*{\vecW-\vecW^{\star}}^2$ (which is 2-smooth) and $g(x)=\lambda x$. For 
\[ \beta \ge 2\prn*{1+\frac{2}{\lambda l_b}} \prn*{d+\frac{8R^2}{\min(r_1,1/2)} \lor \frac{4}{\lambda^2 \min(r_1,1/2) r_2^2}}, \numberthis\label{eq:betareqplus}\]
\pref{thm:generalgflinearizabletopi} gives a PI. Note as \pref{ass:phiselfbounding} is not needed for $F$, no regularity assumptions are placed on $F$. Also note the construction of $\tilde{g}$ is unnecessary here, hence we can just take $\mprimenew = \lambda$, $\bprimenew = 0$.

The concurrent work \citet{gong2024poincare, chewi2024ballistic} only consider P\LCHAR functions, which is not a natural parametrization for this example. Both approaches also do not yield a PI without further assumptions on $F$. Examining Lemma 3.3 of \citet{gong2024poincare}, they require $\beta \ge \frac{4dL}{g_0^2}$ where $g_0$ is a lower bound on the gradients outside $\cW^{\star}$ and $L$ is defined in their Assumption 4 and is analogous to the Lipschitz constant of the Hessian near $\cW^{\star}$. \citet{chewi2024ballistic} requires an upper bound on the Laplacian, which often scales with $d$, even in the standard setting when $F$ is $L$-smooth and so $\Delta F \le dL$. Following their approach to derive a PI, one needs $\beta \nrm*{\grad F}^2 \ge dL$ outside $\vecW^{\star}$ (see their page 10). 

In this Linearizable setting under \pref{ass:linearizableoutsideball}, all we can obtain for generic $F$ is $\nrm*{\grad F(\vecW)} \ge r_1 r_2 \land \frac{r_1 \lambda l_b}{R}$ outside $\cW^{\star}$. Thus the techniques of \citet{gong2024poincare, chewi2024ballistic} require 
\[ \beta \ge d(L \land L') \prn*{\frac1{r_1^2 r_2^2} \lor \frac{R^2}{\lambda^2 l_b^2 r_1^2}}. \]
Often $r_1, r_2$ could be quite small and $R$ is quite large, perhaps even dimension-dependent; these costly terms are multiplied by the dimension $d$ in the requirement for inverse temperature. This is not the case using our result \pref{thm:generalgflinearizabletopi} to obtain our inverse temperature requirement \pref{eq:betareqplus}.
\end{remark}

\subsection{Proof of Weak Poincar\'e Inequality Results \pref{thm:generalgftowpi}, \pref{corr:relaxingFunimodal}}\label{subsec:proofopttosamplingwpi}
\proof[Proof of \pref{thm:generalgftowpi}]{
For the rest of the proof, borrow the same notation as in the proof in \pref{subsec:proofopttosampling}. Consider any test function $\psi$. As in \pref{eq:fdeflyapunov}, let 
\[ f=\psi-\alpha\text{ where } \alpha=\frac1{\mu_{\beta}(\cU)}\int_{\cU} \psi \DERIV\mu_{\beta}.\] 
First, recall we can preserve \pref{def:optimizabilitydef} by replacing $\Phi$ with $\tilde{\Phi}$ and $g$ with $\tilde{g}$, as done in Part 1 of the proof in \pref{subsec:proofopttosampling}. By the work there, which was all done \textit{pointwise}, the resulting $\tilde{\Phi}$ still satisfies \pref{def:optimizabilitydef}, but \textit{now only for all $\vecW\in\cS^c$}. That is, we have 
\[\tri*{\grad \tilde{\Phi}(\vecW), \grad F(\vecW)} \ge \tilde{g}(F(\vecW))\text{ for all }\vecW \in \cS^c.\numberthis\label{eq:wpiphiprimestuff}\]
Moreover, the construction of $\tilde{\Phi}$ there using \pref{ass:linearizableoutsideball} ensures $\tilde{\Phi}$ satisfies $\tri*{\grad \tilde{\Phi}(\vecW), \grad F(\vecW)} \ge \tilde{g}(F(\vecW))$ for all $\vecW \in \ball(\vecW,R+1)^c$. Thus, $\tilde{\Phi}$ does not satisfy \pref{def:optimizabilitydef} only for $\vecW \in \cS \cap \ball(\vecW^{\star}, R+1)$, so we assume from now on that $\cS \subseteq \ball(\vecW^{\star}, R+1)$ (equivalently, take $\cS \leftarrow \cS \cap \ball(\vecW^{\star}, R+1)$). The verification of the smoothness of $\tilde{\Phi}$ did not use optimizability, and so we know that $\tilde{\Phi}$ is $\tilde{L}$-smooth over all of $\mathbb{R}^d$, where $\tilde{L}$ is again defined as in \pref{eq:tildeLdefpi}.

Next, let 
\[ B = \tilde{L} \lor G_F G_{\Phi} \ge 1, \numberthis\label{eq:wpiBdef}\]
where again $\tilde{L}$ is from \pref{eq:tildeLdefpi}, and where we define 
\[ G_F := \sup_{\vecW\in\cS} \nrm*{\grad F(\vecW)} \le L_F R, G_{\Phi} := \sup_{\vecW\in\cS}\nrm*{\grad \tilde{\Phi}(\vecW)} \le \rho_{\Phi}(M'), \numberthis\label{eq:GFPhidef}\]
Here we define $M_F = \sup_{\vecW\in\ball(\vecW^{\star}, R+1)} F(\vecW)$ and upper bound
\[ \nrm*{\grad^2 F(\vecW)}_{\OPNORM} \le \rho_{F}(M_F) := L_F,\]
where we use \pref{ass:phiselfbounding} for $F$ and that $\cS \subseteq \ball(\vecW^{\star}, R+1)$.
Thus, the following holds for all $\vecW\in \cS^c$:
\[ 0<\tilde{g}\prn*{F(\vecW)}+B \le \tri*{\grad \tilde{\Phi}(\vecW), F(\vecW)}+B = -\frac1{\beta}\cL\tilde{\Phi}(\vecW)+\frac1{\beta}\Delta\tilde{\Phi}(\vecW)+B \le -\frac1{\beta}\cL\tilde{\Phi}(\vecW)+\frac1{\beta}\abs*{\Delta\tilde{\Phi}(\vecW)}+B.\]
Define $h(\vecW)=\tilde{g}(F(\vecW))+B$. Thus for all $\vecW\in\cS^c$,
\[ 1 \le -\frac1{\beta}\cdot \frac{\cL\tilde{\Phi}(\vecW)}{h}+\frac1{\beta}\cdot \frac{\abs*{\Delta\tilde{\Phi}(\vecW)}}{h}+\frac{B}{h}.\]
Thus, as $f^2 \ge 0$, we obtain
\begin{align*}
\int f^2 \DERIV \mubeta &= \int_{\cS} f^2 \DERIV \mubeta + \int_{\cS^c} f^2 \DERIV \mubeta  \\
&\le \int_{\cS} f^2 \DERIV \mubeta +\frac1{\beta}\int_{\cS^c} f^2 \frac{-\cL\tilde{\Phi}}{h} \DERIV\mu_{\beta} + \frac1{\beta}\int_{\cS^c} f^2 \frac{\abs*{\Delta\tilde{\Phi}}}{h} \DERIV\mu_{\beta} + \int_{\cS^c} f^2 \frac{B}{h} \DERIV\mu_{\beta} \\
&\le \frac1{\beta}\int f^2 \frac{-\cL\tilde{\Phi}}{h} \DERIV\mu_{\beta} + \frac1{\beta}\int f^2 \frac{\abs*{\Delta\tilde{\Phi}}}{h} \DERIV\mu_{\beta} + \int f^2 \frac{B}{h} \DERIV\mu_{\beta}  + \prn*{\int_{\cS} f^2 \DERIV \mubeta -\frac1{\beta}\int_{\cS} f^2 \frac{-\cL\tilde{\Phi}}{h} \DERIV\mu_{\beta}}\\
&\le \frac1{\beta}\int f^2 \frac{-\cL\tilde{\Phi}}{h} \DERIV\mu_{\beta} + \frac1{\beta}\int f^2 \frac{\abs*{\Delta\tilde{\Phi}}}{h} \DERIV\mu_{\beta} + \int f^2 \frac{B}{h} \DERIV\mu_{\beta}  + \prn*{\int_{\cS} f^2 \DERIV \mubeta + \frac1{\beta}\abs*{\int_{\cS} f^2 \frac{-\cL\tilde{\Phi}}{h} \DERIV\mu_{\beta}}}.
\end{align*}

The last term in parantheses is now our error term. The first three terms will be controlled analogously to \pref{subsec:proofopttosampling}. Namely, the same application of Integration by Parts as in \pref{lem:lyapunovmethodsetup}, which never uses the optimizability condition, yields
\[ \int f^2 \frac{-\cL\tilde{\Phi}}{h} \DERIV\mu_{\beta} \le \int \prn*{ \nrm*{\nabla f}^2+\frac{f^2}{h^2}\nrm*{\nabla \tilde{\Phi}}^2 - \frac{f^2}{h^2}\tri*{ \nabla h, \nabla \tilde{\Phi}}} \DERIV\mu_{\beta}.\]
Substituting this inequality in the above, we obtain in the same way as with \pref{eq:provesamplingsetup} that
\begin{align*}
\int f(\vecW)^2 \frac{\tilde{g}\prn*{F(\vecW)}}{\tilde{g}\prn*{F(\vecW)}+B} \DERIV\mu_{\beta} &\le \frac1{\beta} \int \prn*{ \nrm*{\nabla f(\vecW)}^2+\frac{f(\vecW)^2}{h(\vecW)^2}\nrm*{\nabla \tilde{\Phi}(\vecW)}^2 - \frac{f(\vecW)^2}{h(\vecW)^2}\tri*{ \nabla h(\vecW), \nabla \tilde{\Phi}(\vecW)}} \DERIV\mu_{\beta} \\
&\hspace{1in}+ \frac1{\beta}\int f(\vecW)^2 \frac{\abs*{\Delta\tilde{\Phi}(\vecW)}}{h(\vecW)} \DERIV\mu_{\beta}\\
&\hspace{1in}+ \prn*{\int_{\cS} f(\vecW)^2 \DERIV \mubeta + \frac1{\beta}\abs*{\int_{\cS} f(\vecW)^2 \frac{-\cL\tilde{\Phi}(\vecW)}{h(\vecW)} \DERIV\mu_{\beta}}}.\numberthis\label{eq:wpiproofsetup}
\end{align*}
As discussed in \pref{sec:proofsketch}, we picked up the `error term' $\int_{\cS} f(\vecW)^2 \DERIV \mubeta + \frac1{\beta}\abs*{\int_{\cS} f(\vecW)^2 \frac{-\cL\tilde{\Phi}(\vecW)}{h(\vecW)} \DERIV\mu_{\beta}}$.

\paragraph{Step a.}
Now, we follow Part 2, Step a, \pref{subsec:proofopttosampling} to upper bound the first term in the right hand side above. Note for $\vecW \in \cS^c$, we still have \pref{eq:upperboundintermediateterm} for such $\vecW$, as the proof of \pref{eq:upperboundintermediateterm} only used optimizability pointwise. Otherwise, consider $\vecW \in \cS$. Let 
\[ G' = \sup_{t\in\mathbb{R}}\abs{g'(t)}.\numberthis\label{eq:wpigprimedef}\]
Note this is dimension free and has no $F$-dependence. Note by choice of $h(\vecW)$,
\[ -\tri*{\grad h(\vecW), \grad \tilde{\Phi}(\vecW)} \le \tilde{g}'(F(\vecW)) \nrm*{\grad F(\vecW)} \nrm*{\grad \tilde{\Phi}(\vecW)} \le G' \prn*{\nrm*{\grad F(\vecW)}^2 + \nrm*{\grad \tilde{\Phi}(\vecW)}^2}.\]
Furthermore recall that in Part 2, Step a of \pref{subsec:proofopttosampling}, without using optimizability of $F$, it was established that $\frac{\nrm*{\grad \tilde{\Phi}}^2}{h(\vecW)^2} \le C'$, where $C'$ was defined in \pref{eq:upperboundintermediateterm}. Thus, 
\[ \frac{\nrm*{\grad \tilde{\Phi}(\vecW)}^2-\tri*{\grad h(\vecW), \grad \tilde{\Phi}(\vecW)}}{h(\vecW)^2} \le  \frac{\nrm*{\grad \tilde{\Phi}(\vecW)}^2}{h(\vecW)^2} + G' \frac{\nrm*{\grad F(\vecW)}^2+\nrm*{\grad \tilde{\Phi}(\vecW)}^2}{h(\vecW)^2} \le (G'+1)C'+ G' \frac{\nrm*{\grad F(\vecW)}^2}{h(\vecW)^2}.\]
Recalling $h(\vecW) \ge B \ge 1$, an upper bound on the above is then simply 
\[ C'' := (G+1)C' + G' G_{F}^2,\numberthis\label{eq:Cprimeprimedef}\]
Here $C'$ is from \pref{eq:Cprimedefpi}, $G_F$ is as per \pref{eq:GFPhidef}, and $G'$ is as defined above. As discussed above, this bound still applies in the $\vecW \in \cS^c$ case. Thus, we have for all $\vecW$ that 
\[ \frac{\nrm*{\grad \tilde{\Phi}(\vecW)}^2-\tri*{\grad h(\vecW), \grad \tilde{\Phi}(\vecW)}}{h(\vecW)^2} \le C''.\]

\paragraph{Step b.}
From here, we can conclude a WPI analogously to Step b, \pref{subsec:proofopttosampling}, the one difference being that we need to control the `error term' $\int_{\cS} f(\vecW)^2 \DERIV \mubeta + \frac1{\beta}\abs*{\int_{\cS} f(\vecW)^2 \frac{-\cL\tilde{\Phi}(\vecW)}{h(\vecW)} \DERIV\mu_{\beta}}$. 
For convenience, let
\[ \text{err}(f) := \int_{\cS} f(\vecW)^2 \DERIV \mubeta + \frac1{\beta}\abs*{\int_{\cS} f(\vecW)^2 \frac{-\cL\tilde{\Phi}(\vecW)}{h(\vecW)} \DERIV\mu_{\beta}}.\numberthis\label{eq:wpierrdef}\]
Recalling \pref{eq:wpiproofsetup} and using that $f^2 \ge 0$ thus gives
\begin{align*}
&\int f(\vecW)^2 \frac{\tilde{g}\prn*{F(\vecW)}}{\tilde{g}\prn*{F(\vecW)}+B} \DERIV \mu_{\beta} \\
&\le \frac1{\beta} \int \nrm*{\grad f(\vecW)}^2 \DERIV \mu_{\beta}+\frac1{\beta}\int C'' f(\vecW)^2 \DERIV \mu_{\beta}+\frac1{\beta}\int f(\vecW)^2 \frac{\abs*{\Delta \tilde{\Phi}(\vecW)}}{\tilde{g}(F(\vecW))+B} \DERIV \mu_{\beta}+\text{err}(f) \\
&\le \frac1{\beta}\int \nrm*{\grad f(\vecW)}^2 \DERIV \mu_{\beta} + \frac1{\beta}\int C'' f(\vecW)^2 \DERIV \mu_{\beta}+ \frac1{\beta} \int f(\vecW)^2 \frac{d\tilde{L}}{\tilde{g}\prn*{F(\vecW)}+\tilde{L}} \DERIV \mu_{\beta}+\text{err}(f) \\
&\le \frac1{\beta}\int \nrm*{\grad f(\vecW)}^2 \DERIV \mu_{\beta} + \frac1{\beta}\int f(\vecW)^2 \prn*{d+C''} \DERIV \mu_{\beta}+\text{err}(f),
\end{align*}
where we used \pref{eq:Cprimeprimedef} and the $\tilde{L}$-smoothness of $\tilde{\Phi}$. 
Notice $\frac{\tilde{g}(t)}{\tilde{g}(t)+B}$ is non-decreasing as $\tilde{g}$ is non-decreasing. 
We thus obtain:
\begin{align*}
&\int f(\vecW)^2 \frac{\tilde{g}(l_b)}{\tilde{g}(l_b)+B} \DERIV \mu_{\beta} \\
&= \int_{\cU^c} f(\vecW)^2 \frac{\tilde{g}(l_b)}{\tilde{g}(l_b)+B} \DERIV \mu_{\beta}+\int_{\cU} f(\vecW)^2 \frac{\tilde{g}(l_b)}{\tilde{g}(l_b)+B} \DERIV \mu_{\beta} \\
&\le \int f(\vecW)^2 \frac{\tilde{g}(F(\vecW))}{\tilde{g}\prn*{F(\vecW)}+B} \DERIV \mu_{\beta} + \int_{\cU} f(\vecW)^2 \frac{\tilde{g}(l_b)}{\tilde{g}(l_b)+B} \DERIV \mu_{\beta} \\
&\le \frac1{\beta}\int \nrm*{\grad f(\vecW)}^2 \DERIV \mu_{\beta} + \frac1{\beta}\int f(\vecW)^2 \prn*{d+C''} \DERIV \mu_{\beta} + \frac{\tilde{g}(l_b)}{\tilde{g}(l_b)+B}\int_{\cU} f(\vecW)^2 \DERIV \mu_{\beta}+\text{err}(f). \numberthis\label{eq:proveWPIbeforetrick} \end{align*}
Exactly as in \pref{subsec:proofopttosampling}, using \pref{ass:Funimodal} and the definition $f=\psi-\alpha$ (the choice of $\alpha$ is crucial), we obtain
\[
\int_{\cU} f(\vecW)^2 \DERIV\mu_{\beta} \le \CPILOCAL\int\nrm*{\nabla f(\vecW)}^2 \DERIV\mu_{\beta}.
\]
Applying this in \pref{eq:proveWPIbeforetrick}, we obtain
\begin{align*}
&\int f(\vecW)^2 \frac{\tilde{g}(l_b)}{\tilde{g}(l_b)+B} \DERIV \mu_{\beta}\\
&\le \frac1{\beta}\int \nrm*{\grad f(\vecW)}^2 \DERIV \mu_{\beta} + \frac1{\beta}\int f(\vecW)^2 \prn*{d+C''} \DERIV \mu_{\beta} + \frac{\tilde{g}(l_b)}{\tilde{g}(l_b)+B} \cdot \CPILOCAL \int \nrm*{\grad f(\vecW)}^2 \DERIV \mu_{\beta} +\text{err}(f).
\end{align*}
%Again, the last step follows from an upper bound on $\int_{\cU} f(\vecW)^2 \DERIV\mu_{\beta}$ analogous to we had before, by the definition of $\alpha$.

For $\beta \ge 2\prn*{1+\frac{B}{\tilde{g}(l_b)}} (d+C'')=\Omega(d)$, this gives 
\[ \frac{\tilde{g}(l_b)}{2(\tilde{g}(l_b)+B)}\int f(\vecW)^2 \DERIV \mu_{\beta} \le \frac1{\beta}\int \nrm*{\grad f(\vecW)}^2 \DERIV \mu_{\beta} + \frac{\tilde{g}(l_b)}{\tilde{g}(l_b)+B} \cdot \CPILOCAL \int \nrm*{\grad f(\vecW)}^2 \DERIV \mu_{\beta}+\text{err}(f).\]
Rearranging this inequality and converting back to $\psi$, recalling the definition of variance, and noting $\grad f = \grad \psi$ gives:
\begin{align*}
\mathbb{V}_{\mu_{\beta}}\brk*{\psi} &\le \int \prn*{\psi-\alpha}^2 \DERIV \mu_{\beta} \\
&= \int f^2 \DERIV \mu_{\beta} \\
&\le \prn*{2\CPILOCAL + \frac{2}{\beta}\prn*{1+\frac{B}{\tilde{g}(l_b)}}} \int \nrm*{\grad f}^2 \DERIV \mu_{\beta} + 2\prn*{1+\frac{B}{\tilde{g}(l_b)}} \text{err}(f)\\
&=\prn*{2\CPILOCAL + \frac{2}{\beta}\prn*{1+\frac{B}{\tilde{g}(l_b)}}} \int \nrm*{\grad \psi}^2 \DERIV \mu_{\beta}+ 2\prn*{1+\frac{B}{\tilde{g}(l_b)}} \text{err}(f).
\end{align*}
Finally, we control the error term $\text{err}(f)$. 
First note for $\vecW\in\cS$, by definition of $B$ in \pref{eq:wpiBdef},
\[ \abs*{\frac{-\cL\tilde{\Phi}(\vecW)}{h(\vecW)}} \le \frac{\beta \nrm*{\grad F(\vecW)}\nrm*{\grad \tilde{\Phi}(\vecW)}}{g(F(\vecW))+B}+\frac{\abs*{\Delta \tilde{\Phi}}}{h(\vecW)} \le \frac{\beta G_F G_{\Phi}}{G_F G_{\Phi}}+\frac{d\tilde{L}}{\tilde{L}} \le \beta+d.\]

Next, recall $f=\psi-\alpha$ where $\alpha=\frac1{\mu_{\beta}(\cU)}\int_{\cU} \psi \DERIV\mu_{\beta}=\int_{\cU} \psi \DERIV \mubetalocal$ is defined as before. Note $\alpha \in [\inf \psi, \sup \psi]$. Note for all $\vecW$,
\begin{align*}
\psi(\vecW) - \alpha &\le \sup \psi - \inf \psi = \text{osc}(\psi),\\
\psi(\vecW) - \alpha &\ge \inf \psi - \sup \psi = -\text{osc}(\psi).
\end{align*}
Consequently, we have for all $\vecW$,
\[ f(\vecW)^2 = (\psi(\vecW) - \alpha)^2 \le \text{osc}(\psi)^2. \]
Thus, recalling $\beta \ge d$, we obtain
\begin{align*}
\text{err}(f) &= \int_{\cS} f(\vecW)^2 \DERIV \mubeta + \frac1{\beta}\abs*{\int_{\cS} f(\vecW)^2 \frac{-\cL\tilde{\Phi}(\vecW)}{h(\vecW)} \DERIV\mu_{\beta}} \\
&\le \text{osc}(\psi)^2 \mu_{\beta}(\cS) \prn*{1+\frac1{\beta}(d+\beta)} \le 3\text{osc}(\psi)^2 \mu_{\beta}(\cS).
\end{align*}
Consequently we have 
\[\mathbb{V}_{\mu_{\beta}}\brk*{\psi} \le \prn*{2\CPILOCAL + \frac{2}{\beta}\prn*{1+\frac{\tilde{L}}{\tilde{g}(l_b)}}} \int \nrm*{\grad \psi}^2 \DERIV \mu_{\beta}+ 6\prn*{1+\frac{B}{\tilde{g}(l_b)}}\mu_{\beta}(\cS) \text{osc}(\psi)^2.\]
Recalling $\psi$ is an arbitrary test function, this shows that $\mu_{\beta}$ satisfies a WPI of the form 
\[\prn*{ 2\CPILOCAL + \frac{2}{\beta}\prn*{1+\frac{B}{\tilde{g}(l_b)}}, 6\prn*{1+\frac{B}{\tilde{g}(l_b)}}\mu_{\beta}(\cS) } \text{ for }\beta \ge 2\prn*{1+\frac{B}{\tilde{g}(l_b)}} (d+C''),\numberthis\label{eq:wpiexplicit}\]
where $B$ is defined in \pref{eq:wpiBdef} and $C''$ is defined in \pref{eq:Cprimeprimedef}.
$\hfill\blacksquare$
}
\begin{remark}
Notice that in the region $\cS$ where GF/GD do not work, one would generally expect $\nrm*{\grad F(\vecW)}$ and thus $G_F$ to be very small. Moreover, the dependence on $F$-dependent constants above can be optimized in the above analysis; we made little effort to do so. 
\end{remark}
\begin{remark}
Note that the construction of $\tilde{\Phi}$ is unnecessary if $\Phi$ is smooth, and in this case the expressions simplify analogously to \pref{rem:linearizableboost}. However, in this setting, we cannot assume $\cS \subseteq \ball(\vecW^{\star},R+1)$.
\end{remark}

\proof[Proof of \pref{corr:relaxingFunimodal}]{
If we only have a $(\CWPILOCAL, \deltalocal)$-WPI for $\mubetalocal$ rather than \pref{ass:Funimodal}, we can proceed as follows to prove a WPI for $\mu_{\beta}$.
Perform the exact same moves as in \pref{subsec:proofopttosampling} up until establishing \pref{eq:provePIbeforetrick}, including our choice of arbitrary test function $\psi$ and $f$ defined in terms of $\psi$, none of which utilize \pref{ass:Funimodal}. Follow the exact same notation as in that proof. These same exact steps again give \pref{eq:provePIbeforetrick}:
\[ \int f^2 \frac{\tilde{g}(l_b)}{\tilde{g}(l_b)+\tilde{L}} \DERIV \mu_{\beta} \le \frac1{\beta}\int \nrm*{\grad f}^2 \DERIV \mu_{\beta} + \frac1{\beta}\int f^2 \prn*{d+C'} \DERIV \mu_{\beta} + \frac{\tilde{g}(l_b)}{\tilde{g}(l_b)+\tilde{L}}\int_{\cU} f^2 \DERIV \mu_{\beta}.\numberthis\label{eq:provePIbeforetricknew}\]
Now rather than utilizing a PI for $\mubetalocal$ which we do not have, use the $(\CWPILOCAL, \deltalocal)$-WPI for $\mubetalocal$ on the test function $f$ to obtain
\[ \mathbb{V}_{\mubetalocal}(f) \le \CWPILOCAL \int \nrm*{\grad f}^2 \DERIV \mubetalocal + \deltalocal \osc(f)^2. \]
The left hand side above also equals 
\[ \int f^2 \DERIV \mubetalocal - \prn*{\int f \DERIV \mubetalocal}^2 = \frac1{\mu_{\beta}(\cU)} \int_{\cU} f^2 \DERIV \mubeta - \frac1{\mu_{\beta}(\cU)^2} \prn*{\int_{\cU} f\DERIV\mubeta}^2.\]
That is, we have 
\[\frac1{\mu_{\beta}(\cU)} \int_{\cU} f^2 \DERIV \mubeta - \frac1{\mu_{\beta}(\cU)^2} \prn*{\int_{\cU} f\DERIV\mubeta}^2 \le \frac{\CWPILOCAL}{\mu_{\beta}(\cU)} \int_{\cU} \nrm*{\grad f}^2 \DERIV \mubeta + \deltalocal \osc(f)^2. \]
Recalling the definition of $f$ in terms of $\psi$, the above rearranges to 
\begin{align*}
\int_{\cU} f^2 \DERIV \mubeta &\le\CWPILOCAL \int_{\cU} \nrm*{\grad f}^2 \DERIV \mubeta +\mu_{\beta}(\cU) \cdot \deltalocal \osc(f)^2 \\
&\hspace{1in}+ \frac1{\mu_{\beta}\prn*{\cU}} \prn*{\int_{\cU}  \prn*{\psi(\vecW) -\frac1{\mu_{\beta}(\cU)}\int_{\cU} \psi(\vecW)\DERIV \mu_{\beta}}\DERIV \mu_{\beta}}^2 \\
&\le \CWPILOCAL \int \nrm*{\grad f}^2 \DERIV \mubeta +\deltalocal \osc(f)^2.
\end{align*}
Applying this in \pref{eq:provePIbeforetricknew}, we obtain
\begin{align*}
\int f^2 \frac{\tilde{g}(l_b)}{\tilde{g}(l_b)+\tilde{L}} \DERIV \mu_{\beta} &\le \frac1{\beta}\int \nrm*{\grad f}^2 \DERIV \mu_{\beta} + \frac1{\beta}\int f^2 \prn*{d+C'} \DERIV \mu_{\beta} \\
&\hspace{1in}+ \frac{\tilde{g}(l_b)}{\tilde{g}(l_b)+\tilde{L}} \prn*{ \CWPILOCAL \int \nrm*{\grad f}^2 \DERIV \mu_{\beta} + \deltalocal \osc(f)^2}.
\end{align*}
If $\beta \ge 2\prn*{1+\frac{\tilde{L}}{\tilde{g}(l_b)}} (d+C')=\Omega(d)$, this gives 
\[ \frac{\tilde{g}(l_b)}{2(\tilde{g}(l_b)+\tilde{L})}\int f^2 \DERIV \mu_{\beta} \le \frac1{\beta}\int \nrm*{\grad f}^2 \DERIV \mu_{\beta} + \frac{\tilde{g}(l_b)}{\tilde{g}(l_b)+\tilde{L}} \prn*{ \CWPILOCAL \int \nrm*{\grad f}^2 \DERIV \mu_{\beta} + \deltalocal \osc(f)^2}.\]
Now, we rearrange above the inequality and convert back to $\psi$. Recalling the definition of variance and noting $\grad f = \grad \psi$, and noting $\psi$ is just a constant shift of $f$ and hence $\text{osc}(\psi)=\text{osc}(f)$, we obtain:
\begin{align*}
\mathbb{V}_{\mu_{\beta}}\brk*{\psi} &\le \int \prn*{\psi-\alpha}^2 \DERIV \mu_{\beta} \\
&= \int f^2 \DERIV \mu_{\beta} \\
&\le \prn*{2\CWPILOCAL + \frac{2}{\beta}\prn*{1+\frac{\tilde{L}}{\tilde{g}(l_b)}}} \int \nrm*{\grad f}^2 \DERIV \mu_{\beta} + 2\deltalocal \osc(f)^2\\
&=\prn*{2\CWPILOCAL + \frac{2}{\beta}\prn*{1+\frac{\tilde{L}}{\tilde{g}(l_b)}}} \int \nrm*{\grad \psi}^2 \DERIV \mu_{\beta} + 2\deltalocal\osc(\psi)^2.
\end{align*}
Recalling $\psi$ is an arbitrary test function, this shows that $\mu_{\beta}$ satisfies a WPI with constants 
\[ \prn*{2\CWPILOCAL + \frac{2}{\beta}\prn*{1+\frac{\tilde{L}}{\tilde{g}(l_b)}}, 2\deltalocal}\text{ for }\beta \ge 2\prn*{1+\frac{\tilde{L}}{\tilde{g}(l_b)}} (d+C').\numberthis\label{eq:wpiexplicitcorr}\]
Again, $\tilde{L}$ comes from \pref{eq:tildeLdefpi}, $C'$ comes from \pref{eq:Cprimedefpi}.

The extension to the setting of \pref{thm:generalgftowpi} follows the exact same steps, and shows that 
$\mu_{\beta}$ satisfies a Weak Poincar\'e Inequality of the form 
\[\prn*{ 2\CWPILOCAL + \frac{2}{\beta}\prn*{1+\frac{B}{\tilde{g}(l_b)}}, 6\prn*{1+\frac{B}{\tilde{g}(l_b)}}\mu_{\beta}(\cS) + 2\deltalocal} \text{ for }\beta \ge 2\prn*{1+\frac{B}{\tilde{g}(l_b)}} (d+C''),\]
where again $B$ is defined in \pref{eq:wpiBdef} and $C''$ is defined in \pref{eq:Cprimeprimedef}.
$\hfill\blacksquare$
}

\subsection{Proofs of \pref{corr:standardregularitysampling}, \pref{corr:newregularitysampling}}\label{subsec:proofofcorollaries}
\proof[Proof of \pref{corr:standardregularitysampling}]{
First, apply \pref{thm:generalgflinearizabletopi} to obtain
\[ \CPI = O(\CPILOCAL+1/\beta).\]
\begin{itemize}
    \item Now, the first part on sampling via LMC under \pref{ass:holdercontinuous} follows directly as a corollary of Theorem 7 of \citet{chewi21analysis} on sampling from targets satisfying a PI, which we apply with $\beta L$ in place of $L$ there as our potential in question is $\beta F$, and with R\'enyi divergence of order $q=1$ (hence we obtain a result in $\KL$) and LOI inequality of order $\alpha=1$. The implementation for the step size is \textit{exactly} the same as in these theorems and the corresponding implementation in \citet{chewi21analysis}. In particular the step size $h$ is given by 6.10 of \citet{chewi21analysis}; the only change is changing $L$ to $\beta L$ exactly as mentioned above, and applying the new bounds for initialization in this setting now from \pref{lem:initdivergencecontrolled}. We appeal to \pref{lem:initdivergencecontrolled} to control the initialization, $\KL(\pi_0||\mu_{\beta})$, and the R\'enyi Divergence of order 2, which is $\ln(\chi^2(\pi_0||\mu_{\beta})+1)$. This justifies that the explicit $\beta, d$ dependence of the initialization is $\tilde{O}(\beta)$ for $\beta=\Omega(d)$ up to log factors (see more discussion in \pref{rem:initdivergencenormfactor}). Thus, as a direct corollary of Theorem 7 of \citet{chewi21analysis}, we see that LMC satisfies the following guarantee:
\[ \KL\prn*{\pi_T||\mu_{\beta}} \le \epsilon \text{ after }T=\tilde{O}\prn*{d \prn*{\CPILOCAL+\frac1{\beta}}^{1+\frac1{s}} \beta^{1+\frac3{s}}\epsilon^{-\frac1{s}} \cdot \max\crl*{1,\frac{\beta^{s/2}}{d}}}\text{ iterations}.\]
Applying Pinkser's Inequality yields the desired.

The term $\max\crl*{1,\frac{\beta^{s/2}}{d}}$ warrants some discussion. It arises here in the maximum of Theorem 7, \citet{chewi21analysis}. The second term there does not dominate, and it seems reasonable that the third term there does not dominate, as we justify in \pref{rem:initdivergencenormfactor}. However, now the fourth term in the maximum could dominate, and we argue in \pref{lem:initdivergencecontrolled} that we can take it to be $\tilde{O}(\beta)$. This gives the factor $\max\crl*{1,\frac{\beta^{s/2}}{d}}$.

For more details on the implementation of $\gamma$ here, here $\gamma \le \frac1{768 Th} \le 1$ as per Proposition 29, \citet{chewi21analysis}. Since $\gamma \le 1$, applying \pref{lem:initdivergencecontrolled} gives the claimed bounds on the initialization. $T$ is the iteration count reported above, and the step size $h$ is given by 6.10 of \citet{chewi21analysis}, with the only explicit change of changing $L$ to $\beta L$ and using the new bounds on initialization.

\item The second part on sampling under the Proximal Sampler follows directly from Theorem 5.4, \citet{altschuler2024faster}, on sampling from targets satisfying a PI. The implementation for the step size is \textit{exactly} the same as in these theorems and the corresponding implementation in \citet{altschuler2024faster}, where we take the smoothness constant in their result equal to $\beta L$, the smoothness constant of our potential $\beta F$. Here we can initialize $\pi_0$ as in \pref{corr:standardregularitysampling} for \textit{any} $\gamma \le 1$, and simply use the first part of \pref{lem:initdivergencecontrolled} to argue the initial divergence $\ln(\chi^2(\pi_0||\mu_{\beta}))$ is controlled by $\tilde{O}(\beta+d)$ (again see more discussion in \pref{rem:initdivergencenormfactor}).
\end{itemize}
This completes the proof. $\hfill\blacksquare$
}

Note that the above is simply a corollary of our main results, and is \textit{not} the focus of our work.

\begin{remark}\label{rem:whynolsi}
Notice we only used the PI from \pref{thm:generalgflinearizabletopi} above. Indeed, there is little gain in using the LSI vs PI from \pref{thm:generalgflinearizabletopi} in the proof above. This is certainly not because is no gain in an LSI; rather, it is because our LSI bound loses about a factor of $\beta S$ for $\beta = \Omega(d)$, and so combining \pref{thm:generalgflinearizabletopi} with pre-existing results on sampling under LSI does not give better results.
\end{remark}
\proof[Proof of \pref{corr:newregularitysampling}]{
We first show that \pref{ass:tamedassumptionsimpler} implies the following assumption from \citet{lytras2024tamed}, allowing us to use their results:
\begin{assumption}[Assumption 1 from \citet{lytras2024tamed}]\label{ass:tamedassumption}
Suppose $F$ satisfies the following properties, from Assumption 1, \citet{lytras2024tamed}:
\begin{itemize}
\item Polynomial Lipschitz Continuity: for some $s_1, L'_1>0$, we have for all $\vecW_1,\vecW_2\in\mathbb{R}^d$, 
\[ \nrm*{\grad F(\vecW_1)-\grad F(\vecW_2)} \le L'_1\prn*{1+\nrm*{\vecW_1}+\nrm*{\vecW_2}}^{s_1} \nrm*{\vecW_1-\vecW_2}.\]
\item Weak Dissipativity: for some $s_2 \ge 1$, $A_2, b_2>0$, we have for all $\vecW\in\mathbb{R}^d$, 
\[ \tri*{\grad F(\vecW), \vecW} \ge A_2 \nrm*{\vecW}^{s_2}-b_2.\]
\item Polynomial Jacobian Growth: for some $L_3, s_3>0$ and all $k \ge 2$ for which the following is well-defined, we have for all $\vecW\in\mathbb{R}^d$,
\[ \max\prn*{\nrm*{\grad F(\vecW)}, \nrm*{\grad^k F(\vecW)}_{\OPNORM}} \le L_3(1+\nrm*{\vecW})^{2s_3}.\]
\end{itemize}
\end{assumption}
To verify this, take $k=2$ in \pref{ass:tamedassumptionsimpler}, and note for any $\vecW =t\vecW_1+(1-t)\vecW_2$ for $0\le t \le 1$ that
\[ \nrm*{\grad^2 F(\vecW)} \le L_3(1+\nrm*{t\vecW_1+(1-t)\vecW_2})^{2s_3} \le L_3(1+\nrm*{\vecW_1}+\nrm*{\vecW_2})^{2s_3}.\]
Consequently as this holds for all $\vecW$ in the line segment $\overline{\vecW_1 \vecW_2}$, we obtain
\[ \nrm*{\grad F(\vecW_1)-\grad F(\vecW_2)} \le L_3\prn*{1+\nrm*{\vecW_1}+\nrm*{\vecW_2}}^{2s_3} \nrm*{\vecW_1-\vecW_2},\]
and so from \pref{ass:tamedassumptionsimpler}, we have \pref{ass:tamedassumption} with $L'_1=L_3$, $s_1=2s_3$.

Now to establish \pref{corr:newregularitysampling}, we directly apply Theorems 2 and 3 of \citet{lytras2024tamed}. These results show that their relevant algorithm can yield a distribution $\pi_T$ with $\KL\prn*{\pi_T || \mu_{\beta}} \le \epsilon$ for large enough $T$. In particular:
\begin{itemize}
    \item Theorem 2 of \citet{lytras2024tamed} shows that under \pref{ass:tamedassumption}, if $\mu_{\beta}$ satisfies a Log-Sobolev Inequality with constant $\CLSI$, then via their algorithm wd-TULA we have
    \[ \KL\prn*{\pi_T || \mu_{\beta}} \le \epsilon\text{ within }T = \tilde{O}\prn*{\frac{\textup{poly}(d, \beta) \CLSI}{\epsilon} \log\prn*{\frac{\KL\prn*{\pi_0||\mu_{\beta}}}{\epsilon}}}\text{ iterations}.\]
    \item Theorem 3 of \citet{lytras2024tamed} shows that under \pref{ass:tamedassumption}, if $\mu_{\beta}$ satisfies a Poincar\'e Inequality with constant $\CPI$, then via their algorithm reg-TULA we can take
    \[ \KL\prn*{\pi_T || \mu_{\beta}} \le \epsilon\text{ within }T = \tilde{O}\prn*{\textup{poly}\prn*{d, \beta, \CPI, \frac1{\epsilon}}\log\prn*{\frac{\KL\prn*{\pi_0||\hat{\mu}_{\beta}}}{\epsilon}}}\text{ iterations}.\]
    Here, $\hat{\mu}_{\beta}$ corresponds to $e^{-\prn*{\beta F(\vecW)+\eta \nrm*{\vecW}^{2r+2}}}/ Z$, where $r$ is taken large enough in terms of the exponents $s_1, s_2, s_3$ from \pref{ass:tamedassumption}. The degree of these polynomials also depends on $s_1, s_2, s_3$.
\end{itemize}
Note Assumption 1 of \citet{lytras2024tamed} is phrased in terms of the true potential $\beta F$ rather than $F$. Their results have polynomial $d$ dependence, but to convert these results to our setting where $\beta=\Omega(d)$, we need to track their proofs and find the explicit dependency on their parameters $A, L, L', b$, which are scaled up by $\beta$ for us. 

We explicitly make this conversion here for the reader's convenience: converting to their notation we have 
\[ L' = \beta L'_1 = \beta L_3, A=\beta A_2, b = \beta b_2, L = \beta L_3.\]
The powers do not change: converting to their notation we still have $l'=s_1$, $a=s_2$, $l=s_3$.
For the rest of this discussion, we follow the notation of \citet{lytras2024tamed} so the reader can easily reference their work.

We find that this dependency is polynomial in their guarantees from Theorems 2 and 3. In particular, we carefully track this for $\hat{C}$ from their Theorem 2 and their $\hat{C}, \dot c$ from their Theorem 3, and see the dependencies on these is polynomial with respect to $d, A', K, L, L', b$ from their Assumption 1. By consequence the dependence on $\beta$ is also polynomial.\footnote{This is to be expected; in many results on discrete-time LMC, e.g. \citet{chewi21analysis}, dependence on smoothness constants (which are also scaled up by $\beta$ here) are polynomial.} However such dependence on problem-dependent $A', K, L, L', b$ is not made as explicit in \citet{lytras2024tamed}, and so we explicitly track this here. For more details:
\begin{itemize}
    \item Consider their Theorem 2. The convergence rate there is given in terms of $\CLSI, \KL(\pi_0||\mu_{\beta})$, and $\hat{C}$. $\hat{C}$ bounds the discretization error, and through the proof of Lemma A.5, $\hat{C}$ is in turn given by a polynomial function of $C_{1,p}, C_p$ for integers $p \ge 0$ from their Lemmas A.3 and A.4. These quantities control various moment bounds. In turn, these are all given in terms of the $C_p$ from their Lemma A.3 and polynomial factors in $A', L, L', d$ (recall $A', L, L'$ are $\beta$ times our smoothness constants). $C_p$ here is at most $(\ln C_{\mu})^{2p}$ where $C_{\mu}$ is defined in their Lemma A.2 and controls the growth of particular exponential moments. Tracking the proof of Lemma A.2, we can see that $C_{\mu} \le \exp\{ \textup{poly}\prn*{A, L, L', b, d}\}$. Thus $C_p \le \textup{poly}\prn*{A, L, L', b, d}$, and so $\hat{C}\le \textup{poly}\prn*{A, L, L', b, d}$. 
    \item Consider their Theorem 3. This is derived from their Theorem 7, where the convergence rate there is given in terms of $\hat{C}$, which again controls discretization error, and $\dot c$, which governs the Log-Sobolev constant of a particular regularized version of the potential $\beta F$. The regularization is in particular given by $\beta F(\vecW)+\lambda \nrm*{\vecW}^{2r+2}$. Here $\lambda$ denotes the step size and we can without loss of generality take $\lambda \le 1$. 

    First we consider $\hat{C}$. Analyzing the proof of Theorem 7, we see that it is given by the sum of $C^{\text{reg}}_{\text{tam}}$ and $C^{\text{reg}}_{\text{onestep}}$ from Lemmas C.2, C.3. In turn, these quantities are controlled exactly the same way by the moment bounds as in Lemmas A.5, and in turn Lemmas A.3 and A.4, except now we are dealing with the regularized potential $\beta F + \lambda \nrm*{\vecW}^{2r+2}$ rather than the original potential $\beta F$ (this is shown for example in their Lemma C.6). As noted in the article, we can prove analogous moment bounds the same way, with dependence that is still $\textup{poly}\prn*{A, L, L', b, d}$. This is because the proof of their Lemma A.6 shows the regularized potential still satisfies their Assumption 1 parts A1 and A2 and a result analogous to Lemma A.1, with smoothness parameters only a universal constant shift from $A, L, L', b$ for regularization $\lambda \le 1$. These are all the conditions needed to prove Lemma A.2, which in turn give the desired bounds Lemma A.3 and A.4, for the regularized potential.

    Next we consider $\dot c$. The dependence of $\dot c$ on $\lambda$ is given in Proposition 3.8, \citet{lytras2024tamed}, which upon converting to our notation, is $\prn*{\frac1{\lambda}}^{\frac1{r+1}+\frac{s_1}{2r-s_1}}$. We need $\frac1{r+1}+\frac{s_1}{2r-s_1}\le1$ to obtain a meaningful convergence rate, and indeed we can make $\frac1{r+1}+\frac{s_1}{2r-s_1}\le \frac12$ by taking $r$ large enough in terms of $s_1$. The dependence of $\dot c$ on all other parameters is given from their equation C.8 in the proof of their Proposition A.4 (we note that the third term in that equation is a typo and should read, following their notation, $\frac{K_{\lambda}}{A_{\text{reg}}}$ from using Theorem 3.15 of \citet{menz2014poincare}). We can check that, by what we have argued on moment control in the above paragraph, all the other parameters $A_{\text{reg}}, K_{\lambda}, \pi_{\text{reg}}(\nrm*{x}^2)$ and Poincar\'e constant of the Gibbs measure of the regularized potential all depend polynomially on $A', L, L', b, d$. Hence $\dot c$ depends polynomially on $A', L, L', b, d$.

    We conclude upon applying the same rationale as Theorem 7 and Corollary 4 of \citet{lytras2024tamed}.
\end{itemize}
One additional point of consideration is these results contain dependence on initial divergences 
\[ \KL\prn*{\pi_0||\mu_{\beta}}, \KL\prn*{\pi_0||\hat{\mu}_{\beta}}.\]
We argue that these both can be controlled by $\tilde{O}(d\beta)$ in \pref{lem:controlregularizedklinit}, given appropriate initialization. As noted on footnote 1 of page 7 of \citet{lytras2024tamed}, or just by tracking their proof, we note that their result holds for any initialization (at the expense of a different price for initialization $\KL\prn*{\pi_0||\mu_{\beta}}, \KL\prn*{\pi_0||\hat{\mu}_{\beta}}$). Note since these initializations are polynomial in $d,\beta$, they do not affect the claimed rate or \pref{corr:newregularitysampling} (as they appear in the logarithm, as per \citet{lytras2024tamed}). Putting all this together, combining with Points 1 and 2, and using Pinkser's Inequality gives \pref{corr:newregularitysampling}.
$\hfill\blacksquare$
}

We emphasize that we just cite the result of \citet{lytras2024tamed} and made no attempt to optimize the polynomial dependency. The focus on our work is on proving isoperimetric inequalities. Moreover, while the dependence indicated above is polynomial, again note the degree of the polynomials in question depends on the exponents $s_1, s_2, s_3$ from \pref{ass:tamedassumption}.

\subsection{Proofs of \pref{subsec:localoptimizability}}\label{subsec:localconditionproofs}
We first verify that $\hat{\mu}_{\beta}$, $\mu_{\beta}$ are indeed close in $\TV$ distance:
\begin{lemma}\label{lem:closetvlocalcondition}
Defining $\delta$ as in \pref{corr:localconditionsampling}, we have $\TV(\hat{\mu}_{\beta}, \mu_{\beta}) \le 3\delta$.
\end{lemma}
\proof{
Let $I = \int_{\ball(\vecW, R-1)} e^{-\beta F(\vecW)}\DERIV \vecW$. By construction of $\hat{F}$, we also have $I = \int_{\ball(\vecW, R-1)} e^{-\beta \hat{F}(\vecW)}\DERIV \vecW$. Let $I_1 = \int_{\ball(\vecW, R-1)^c} e^{-\beta F(\vecW)}\DERIV \vecW$, $I_2 = \int_{\ball(\vecW, R-1)^c} e^{-\beta \hat{F}(\vecW)}\DERIV \vecW$. Note $I_2 \le I_1$ as $\hat{F} \ge F$ on $\ball(\vecW, R-1)^c$. Consequently, recalling the definition of $\delta$, we have 
\[ 1 \ge \frac{I}{I+I_2} \ge \frac{I}{I+I_1} \ge 1-\delta, \text{ thus } 0 \le \frac{I_1}{I+I_1}, \frac{I_2}{I+I_2} \le \delta.\]
Now consider any subset $\cA \subseteq \mathbb{R}^d$, and let $\cA_1=\cA \cap \ball(\vecW, R-1)$, $\cA_2=\cA \cap \ball(\vecW, R-1)^c$. Note $F, \hat{F}$ agree on $\cA_1$ and so $\int_{\cA_1} e^{-\beta F(\vecW)}\DERIV \vecW = \int_{\cA_1} e^{-\beta \hat{F}(\vecW)}\DERIV \vecW = xI$ for some $x \in [0,1]$. Let $Y_1 = \int_{\cA_2} e^{-\beta F(\vecW)}\DERIV \vecW$, let $Y_2=\int_{\cA_2^c} e^{-\beta \hat{F}(\vecW)}\DERIV \vecW$, and note $Y_1 \le I_1$, $Y_2 \le I_2$. Thus we obtain
\begin{align*}
\abs*{\hat{\mu}_{\beta}(\cA) - \mu_{\beta}(\cA)} &= \abs*{\frac{xI}{I+I_1} - \frac{xI}{I+I_2} + \frac{Y_1}{I+I_1} - \frac{Y_2}{I+I_2}}\\
&\le \abs*{\frac{xI}{I+I_1} - \frac{xI}{I+I_2}} + \abs*{\frac{Y_1}{I+I_1} - \frac{Y_2}{I+I_2}} \\
&\le x \abs*{\frac{I}{I+I_1} - \frac{I}{I+I_2}} + \frac{Y_1}{I+I_1}+\frac{Y_2}{I+I_2} \\
&\le \delta+\delta+\delta=3\delta.
\end{align*}
This applies for all $\cA \subset \mathbb{R}^d$, and we conclude. $\hfill\blacksquare$
}

\proof[Proof of \pref{prop:obtainlocalcondition}]{
Let $\cU = \ball(\cW^{\star}, r(l_b))$ for any $l_b$ satisfying \pref{ass:Funimodal}.
\paragraph{Part 1: Modifying the Interpolation Argument}
Recall for a suitable bump function $\chi_F \in [0,1]$ which we will define later, we defined
\[ \tilde{F}(\vecW) := \begin{cases} F(\vecW)&:\nrm*{\vecW-\vecW^{\star}} \le R-1 \\ F(\vecW)+ \chi_F(\vecW) \cdot \lambdareg(\nrm*{\vecW-\vecW^{\star}}^{2}+1)&:R-1<\nrm*{\vecW-\vecW^{\star}} < R \\ F(\vecW)+ \lambdareg(\nrm*{\vecW-\vecW^{\star}}^{2}+1) &: R\le\nrm*{\vecW-\vecW^{\star}}, \end{cases}\]
where $\lambdareg = L$.\footnote{In fact, $\lambdareg$ can be any upper bound on $L$, which can be seen by tracking the following proof.}
Also let 
\[ L_{b,1} = \inf_{R-1 \le \nrm*{\vecW-\vecW^{\star}} \le R} F(\vecW).\]
By assumption that $\ball(\cW^{\star}, r(l_b)) \subseteq \ball(\vecW^{\star}, R-1)$, we have $L_{b,1} \ge l_b$.

We now show that we can perform the same interpolation steps as in the proof of \pref{thm:generalgflinearizabletopi} in \pref{subsec:proofopttosampling}, Step 1, to create $\tilde{\Phi}$, except using $\tilde{F}$ in place of $F$. From here, very similar steps as the proof of \pref{thm:generalgflinearizabletopi} in \pref{subsec:proofopttosampling} prove that $\hat{\mu}_{\beta} \propto \exp(-\beta \hat{F})$ satisfies a PI. To this end, define the interpolators as follows. First define
\[ M=\crl*{\sup_{\vecW\in\ball(\vecW^{\star}, R)} \Phi(\vecW)+\sup_{\vecW\in\ball(\vecW^{\star}, R)}  F(\vecW)} \lor \frac1{\lambda} \prn*{\frac14 g(L_{b,1})+1}.\numberthis\label{eq:Mdeflocal}\]
Now let $\chi(\vecW) = p(\nrm*{\vecW-\vecW^{\star}}-(R-1))$ be the interpolator from the proof in \pref{subsec:proofopttosampling}, where $p(x)=\frac{e^{-1/x^2}}{e^{-1/x^2}+e^{-1/(1-x^2)}}$. Recall the derivatives of $p$, and hence $\nrm*{\grad \chi(\vecW)}, \nrm*{\grad^2 \chi(\vecW)}_{\OPNORM}$, are upper bounded by $B$ for a universal ($F$-independent) constant $B$, and that $p$ is differentiable to all orders. As per \pref{lem:scalenonnegative}, we know $p$ is increasing on $[0,1]$ as well. (We extend $p$ to $[0,1]$ by $p(0)=0, p(1)=1$, which clearly preserves all these properties.)

Let $\sigma_{\Phi}$ be a bijection from $[0,1]$ to itself such that $p(\sigma_{\Phi}(1/2))=1/2$. Clearly we can choose $\sigma_{\Phi}$ to be infinitely differentiable, increasing, and with first and second derivatives bounded by a universal, $F$-independent constant. Now define define the interpolator $\chi_{\Phi}$ for $\Phi$ by
\[ p_{\Phi} = p\circ \sigma_{\Phi}, \chi_{\Phi}(\vecW) =  p_{\Phi}(\nrm*{\vecW-\vecW^{\star}}-(R-1)).\]
Consequently, $\chi_{\Phi}(1/2)=1/2$, $\chi_{\Phi}$ is increasing, and $\chi_{\Phi}$ has gradient norm and Hessian operator norm bounded by a universal constant $B_{\Phi}$.

Next let
\[ c_F := \frac{g(L_{b,1})}{8\lambdareg (R^{2}+1) \rho_{\Phi}(M)}, \tthresF = 1/2.\]
Let $\sigma_F$ be a bijection from $[0,1]$ to itself such that $p(\sigma_{\Phi}(1/2))=c_F$. Clearly we can choose $\sigma_{F}$ to be infinitely differentiable, increasing, and with first and second derivatives bounded by a \textit{$c_F$-dependent} constant, which in turn depends on $F, \Phi$ in turn. Let $\tilde{\chi}_F$ be defined by
\[q_F = p \circ \sigma_F, \tilde{\chi}_F(\vecW) = q_F(\nrm*{\vecW-\vecW^{\star}}-(R-1)).\]
Hence $q_F$ is increasing and $q_F(1/2)=c_F$. Now define the interpolator $\chi_F$ for $F$ by 
\[\chi_F(\vecW) = \int_0^{\nrm*{\vecW-\vecW^{\star}}-(R-1)} q_F(t) \DERIV t. \numberthis\label{eq:interpolateFlocalconditiondef}\]
It follows that $\chi_F$ is increasing. Also define $p_F(x) = \int_0^x q_F(t) \DERIV t$. Thus $p'_F = q_F$ and $p'_F$ is increasing, $p'_F(1/2)=c_F$, and that
\[ \chi_F(\vecW) = p_F(\nrm*{\vecW-\vecW^{\star}}-(R-1)).\]
Also, notice for $\nrm*{\vecW-\vecW^{\star}}-(R-1) \le \tthresF$,
\[ \chi_F(\vecW) = \int_0^{\nrm*{\vecW-\vecW^{\star}}-(R-1)} p'_F(t) \DERIV t \le \sup_{0 \le t \le \nrm*{\vecW-\vecW^{\star}}-(R-1)} p'_F(t), \text{ thus } p_F(t) \le c_F\text{ for }t \le 1/2.\numberthis\label{eq:boundintegralinterpolateF}\]
It also follows by the above discussion that $\chi_F$ has gradient norm and Hessian operator norm bounded by an $F$-dependent parameter $B_F$. 

Finally, let 
\[ \Phi_2 = \cwgt \nrm*{\vecW-\vecW^{\star}}^2+2M \]
where $\cwgt$ is defined by
\[\cwgt = \frac{g(L_{b,1})}{\lambdareg (R-1) ((R-1)^2+1)c_F} \lor \frac{2\rho_{\Phi}(M) R}{(R-1)^2}.\]
This defines how much we regularize by $\nrm*{\vecW-\vecW^{\star}}^2$ to ensure this construction is successful. Notice $\Phi_2 \ge \Phi$ on $\ball(\vecW^{\star}, R)$. In terms of $\Phi_2$, define 
\[ \tilde{\Phi}(\vecW) := \chi_{\Phi}(\vecW) \Phi_2(\vecW) +(1-\chi_{\Phi}(\vecW))\Phi(\vecW).\numberthis\label{eq:localtildephidef}\]

We first show:
\begin{lemma}\label{lem:regularizeFsmooth}
$\hat{F}$ is smooth with smoothness constant $O(1)$ (here $O(\cdot)$ hides problem-dependent parameters).
\end{lemma}
\proof{
This is evident for $\nrm*{\vecW-\vecW^{\star}} \le R-1$, $\nrm*{\vecW-\vecW^{\star}} \ge R$, where it is straightforward to verify that $\nrm*{\grad^2 \tilde{F}} \le 3L$. Otherwise, we have 
\begin{align*}
\grad \tilde{F} &= \grad F + \grad \chi_F \cdot \lambdareg (\nrm*{\vecW-\vecW^{\star}}^2+1) + \chi_F \cdot 2\lambdareg (\vecW-\vecW^{\star}),
\end{align*}
and
\begin{align*}
\grad^2 \tilde{F} &= \grad^2 F + \grad^2 \chi_F \cdot \lambdareg (\nrm*{\vecW-\vecW^{\star}}^2+1) + \grad \chi_F \cdot 2\lambdareg (\vecW-\vecW^{\star})^T \\
&\hspace{1in}+ \grad \chi_F\cdot 2\lambdareg (\vecW-\vecW^{\star}) + 2\lambdareg \chi_F. 
\end{align*}
Recalling $\lambdareg=L$, Triangle Inequality thus gives
\[ \nrm*{\grad^2 \tilde{F}} \le L+LB_F (R^2+1)+4LB_F R + 2L B_F. \]
This proves this Lemma. $\hfill\blacksquare$
}

The benefit of regularizing is shown via the following Lemma.
\begin{lemma}\label{lem:regularizeroutsideniceproperties}
For $\vecW \in \ball(\vecW^{\star}, R)^c$, we have 
\[ \tri*{\grad \Phi_2(\vecW), \grad \tilde{F}(\vecW)} \ge g(F(\vecW)).\]
\end{lemma}
\proof[Proof of \pref{lem:regularizeroutsideniceproperties}]{
For such $\vecW$, 
\begin{align*}
\tri*{\grad \Phi_2(\vecW), \grad F(\vecW)} &= \tri*{\grad \Phi_2(\vecW), \grad F(\vecW) + 2\lambdareg (\vecW-\vecW^{\star})} \\
&= 2\cwgt\prn*{\tri*{\vecW-\vecW^{\star}, \grad F(\vecW)} + 2 \lambdareg\nrm*{\vecW-\vecW^{\star}}^2}.
\end{align*}
Thus we have
\begin{align*}
\tri*{\grad \Phi_2(\vecW), \grad F(\vecW)} &\ge 2\cwgt \prn*{ 2\lambdareg\nrm*{\vecW-\vecW^{\star}}^{2} - L \nrm*{\vecW-\vecW^{\star}}} \ge L\nrm*{\vecW-\vecW^{\star}}^{2} \ge g(F(\vecW)),
\end{align*}
where the last inequality follows from $L$-smoothness of $F$ and that $g(x)=\lambda x$ for $\lambda \le 1$. $\hfill\blacksquare$
}

From \pref{lem:regularizeroutsideniceproperties} and the definition of $\Phi_2(\cdot)$, we directly obtain the following Corollary.
\begin{corollary}\label{corr:localinnerproductnonnegative}
For $\vecW$ with $\nrm*{\vecW-\vecW^{\star}} \in [R-1,R]$, we have 
\[ \tri*{\vecW-\vecW^{\star}, \grad \tilde{F}(\vecW)} \ge 0.\]
\end{corollary}

Now we break into cases and show that $\tilde{\Phi}$ is still a valid Lyapunov function, in an appropriate sense:
\begin{itemize}
    \item For $\vecW \in \ball(\vecW^{\star}, R-1)$, as $\chi_{\Phi}, \chi_F \equiv 0$ holds identically in this set, we have 
\[ \tri*{\grad \tilde{\Phi}(\vecW), \grad \tilde{F}(\vecW)} \equiv \tri*{\grad \Phi(\vecW), \grad F(\vecW)} \ge g(F(\vecW)).\]
\item For $\vecW \in \ball(\vecW^{\star}, R)^c$, as $\chi_F, \chi_{\Phi} \equiv 1$ identically in this set, we have by \pref{lem:regularizeroutsideniceproperties}
\[ \tri*{\grad \tilde{\Phi}(\vecW), \grad \tilde{F}(\vecW)} = \tri*{\grad \Phi_2(\vecW), \grad \tilde{F}(\vecW)} \ge g(F(\vecW)). \]

\item For $\vecW \in \ball(\vecW^{\star}, R-1)^c \cap \ball(\vecW^{\star}, R)$, we have 
\[ \grad \tilde{\Phi}(\vecW)=\chi_{\Phi}(\vecW) \grad \Phi_2(\vecW)+(1-\chi_{\Phi}(\vecW))\grad\Phi(\vecW)+\grad \chi_{\Phi}(\vecW) \Phi_2(\vecW) - \grad \chi_{\Phi}(\vecW) \Phi(\vecW).\]
First let $L_{b,1}$ denote the minimum value of $F$ in $\ball(\vecW^{\star}, R-1)^c \cap \ball(\vecW^{\star}, R)$. Note $L_{b,1} \ge l_b$ by assumption that $\ball(\cW^{\star}, r(l_b)) \subseteq \ball(\vecW^{\star}, R-1)$.
This means
\begin{align*}
\tri*{\grad \tilde{\Phi}(\vecW), \grad \tilde{F}(\vecW)} &= (1-\chi_{\Phi}(\vecW))\tri*{\grad \Phi(\vecW), \grad \tilde{F}(\vecW)} +\chi_{\Phi}(\vecW)\tri*{\grad \Phi_2(\vecW), \grad \tilde{F}(\vecW)}\\
&\hspace{1in}+(\Phi_2(\vecW)-\Phi(\vecW)) \tri*{\grad \chi_{\Phi}(\vecW), \grad \tilde{F}(\vecW)}.\numberthis\label{eq:regFintermediatelowerbound}
\end{align*}
Note in this region,
\[ \grad\tilde{F}(\vecW) = \grad F(\vecW)+ \grad \chi_F(\vecW) \cdot \lambdareg(\nrm*{\vecW-\vecW^{\star}}^{2}+1) + \chi_F(\vecW) \cdot 2\lambdareg \nrm*{\vecW-\vecW^{\star}}(\vecW-\vecW^{\star}).\]
Also recall that 
\[ \grad \chi_{\Phi}(\vecW) = p'_{\Phi}(\nrm*{\vecW-\vecW^{\star}}-(R-1)) \frac{\vecW-\vecW^{\star}}{\nrm*{\vecW-\vecW^{\star}}}, \grad \chi_F(\vecW) = p'_F(\nrm*{\vecW-\vecW^{\star}}-(R-1)) \frac{\vecW-\vecW^{\star}}{\nrm*{\vecW-\vecW^{\star}}}.\]
Define
\begin{align*}
A &= \tri*{\grad \Phi(\vecW), \grad F(\vecW)} \ge g(F(\vecW)) \ge 0,\\
B_1 &= \frac{p'_F(\nrm*{\vecW-\vecW^{\star}}-(R-1))}{\nrm*{\vecW-\vecW^{\star}}}\lambdareg(\nrm*{\vecW-\vecW^{\star}}^{2}+1)\tri*{\grad \Phi(\vecW), \vecW-\vecW^{\star}},\\
B_2 &= \chi_F(\vecW) \cdot 2\lambdareg \nrm*{\vecW-\vecW^{\star}}\tri*{\grad \Phi(\vecW), \vecW-\vecW^{\star}}, \\
C_1 &= \cwgt \lambdareg (\nrm*{\vecW-\vecW^{\star}}^{2}+1) \tri*{\grad \chi_F(\vecW), \vecW-\vecW^{\star}}\\
&=\cwgt \lambdareg (\nrm*{\vecW-\vecW^{\star}}^{2}+1) \nrm*{\vecW-\vecW^{\star}} p'_F(\nrm*{\vecW-\vecW^{\star}}-(R-1)) \ge 0, \\
&=B_1 \cwgt \frac{\nrm*{\vecW-\vecW^{\star}}^2}{\tri*{\grad \Phi(\vecW), \vecW-\vecW^{\star}}},\numberthis\label{eq:C1relation}\\
C_2 &= 2 \cwgt \lambdareg  \chi_F(\vecW) \nrm*{\vecW-\vecW^{\star}}^{3} \ge 0\\
&= B_2 \cwgt \frac{\nrm*{\vecW-\vecW^{\star}}^2}{\tri*{\grad \Phi(\vecW), \vecW-\vecW^{\star}}}\numberthis\label{eq:C2relation},\\
C_3 &= \cwgt \tri*{\grad F(\vecW), \vecW-\vecW^{\star}}\ge 0.
\end{align*}
It is clear that $C_1, C_2 \ge 0$, and $C_3 \ge 0$ follows by \pref{ass:localconditionassumption}. In the above, $A, C_1, C_2$ are favorable terms, and $B_1, B_2$ are terms that could be negative that we must control.

Recalling the definition of $\Phi_2$, that for $\vecW \in \ball(\vecW^{\star}, R)$ we have $\Phi(\vecW) \le M$, and furthermore using \pref{corr:localinnerproductnonnegative}, we obtain from \pref{eq:regFintermediatelowerbound} that
\begin{align*}
\tri*{\grad \tilde{\Phi}(\vecW), \grad \tilde{F}(\vecW)} &\ge (1-\chi_{\Phi}(\vecW))(A+B_1+B_2)\\
&+\prn*{\cwgt \chi_{\Phi}(\vecW)+ \frac{M p'_{\Phi}(\nrm*{\vecW-\vecW^{\star}}-(R-1))}{R}} \tri*{\grad \tilde{F}(\vecW), \vecW-\vecW^{\star}} \\
&\ge (1-\chi_{\Phi}(\vecW))(A+B_1+B_2) + \cwgt \chi_{\Phi}(\vecW) \tri*{\grad \tilde{F}(\vecW), \vecW-\vecW^{\star}} \\
&= (1-\chi_{\Phi}(\vecW))(A+B_1+B_2)+\chi_{\Phi}(\vecW) \prn*{C_1+ C_2 + C_3}.
\end{align*}
We aim to find a lower bound on the above. 
We break into cases:
\begin{enumerate}
    \item Suppose $\nrm*{\vecW-\vecW^{\star}} < R-1+\tthresF$. In this case by \pref{corr:localinnerproductnonnegative}, it remains to lower bound $(1-\chi_{\Phi}(\vecW))(A+B_1+B_2)$ by a positive constant. 
    (This is where it becomes very useful to have independent interpolators $\chi_F, \chi_{\Phi}$.)
    
    By construction of $\chi_{\Phi}$, for $\nrm*{\vecW-\vecW^{\star}} < R-1+\tthresF$, recall we have 
    \[ \chi_{\Phi}(\vecW) = p_{\Phi}(\vecW-\vecW^{\star}-(R-1)) \le \frac12.\]
    That is, we still `weight' $\Phi$ substantially in the construction of $\tilde{\Phi}$.
    Furthermore for such $\vecW$, recall by \pref{eq:boundintegralinterpolateF} that
    \begin{align*}
    F(\vecW) &\ge g(L_{b,1}).\\
    p'_F(\nrm*{\vecW-\vecW^{\star}}-(R-1)) &\le c_F.\\
    \chi_F(\nrm*{\vecW-\vecW^{\star}}-(R-1)) &= p_F(\nrm*{\vecW-\vecW^{\star}}-(R-1)) \le c_F.
    \end{align*}
    That is, we do not weight the regularizer much yet.
    
    Thus we obtain
    \begin{align*}
    \abs*{B_1} &\le p'_F(\nrm*{\vecW-\vecW^{\star}}-(R-1)) \cdot \lambdareg (R^{2}+1) \rho_{\Phi}(M) \le \frac14 g(L_{b,1}).\\
    \abs*{B_2} &\le p_F(\nrm*{\vecW-\vecW^{\star}}-(R-1)) \cdot 2\lambdareg R^{2} \rho_{\Phi}(M) \le \frac14 g(L_{b,1}).
    \end{align*}
    Consequently we have 
    \[ A+B_1+B_2 \ge g(F(\vecW))- \frac12 g(L_{b,1}) \ge \frac12 g(F(\vecW))  + \frac12 g(L_{b,1}) - \frac12 g(L_{b,1})=\frac12 g(F(\vecW)). \]
    Recalling $C_1,C_2,C_3 \ge 0$, we obtain
    \[ \tri*{\grad \tilde{\Phi}(\vecW), \grad \tilde{F}(\vecW)} \ge (1-\chi_{\Phi}(\vecW))(A+B_1+B_2) \ge \frac14 g(F(\vecW)) \ge \frac14 g(L_{b,1}).\]
    
    \item Suppose $\nrm*{\vecW-\vecW^{\star}} \ge R-1+\tthresF$. In this case $A+B_1+B_2<0$ is possible. The benefit however is that $\cwgt$ comes into play, and allows for $C_1, C_2$ to dominate. 
    The relations \pref{eq:C1relation}, \pref{eq:C2relation} between $B_1, C_1$ and $B_2, C_2$ earlier, that $\vecW\in\ball(\vecW^{\star}, R)$ in this case, and the choice of $\cwgt$ together imply that
    \[ B_2+C_2 \ge 0, B_1+\frac{C_1}2 \ge 0.\]
    Notice here by construction of $\chi_{\Phi}$ that in this case, we have $\chi_{\Phi}(\vecW) \ge \frac12$.
    Consequently, we have 
    \begin{align*}
      \tri*{\grad \tilde{\Phi}(\vecW), \grad \tilde{F}(\vecW)}  &\ge (1-\chi_{\Phi}(\vecW))(A+B_1+B_2)+\chi_{\Phi}(\vecW) \prn*{C_1+ C_2 + C_3} \\
      &\ge \frac12 (B_1+B_2)+\frac12 (C_1+C_2) \\
      &\ge \frac14 C_1 + \frac14 C_2 + \frac12 B_1 \ge \frac14 C_1.
    \end{align*}
    By choice of $\cwgt$, and since
     \[ p'_F(\nrm*{\vecW-\vecW^{\star}}-(R-1)) \ge c_F\text{ for }\nrm*{\vecW-\vecW^{\star}} \ge R-1+\tthresF,\]
     we have for such $\vecW$,
    \[ \frac14 C_1 \ge \frac14\cwgt \lambdareg ((R-1)^2+1)(R-1) c_F \ge \frac14 g(L_{b,1}).\]
    This last step follows by definition of $\cwgt$.
\end{enumerate}
Thus in either case, we have
\[\tri*{\grad \tilde{\Phi}(\vecW), \grad \tilde{F}(\vecW)} \ge \frac14 g(L_{b,1}).\]
\end{itemize}
Putting these cases together, we obtain:
\begin{enumerate}
    \item For $F(\vecW) \ge M$, then we must have $\vecW\in\ball(\vecW^{\star}, R)^c$ and so
    \[ \tri*{\grad \tilde{\Phi}(\vecW), \grad \tilde{F}(\vecW)} \ge g(F(\vecW)).\]
    \item For $F(\vecW) \in [l_b, M)$, then as $F$ is non-decreasing and as $L_{b,1} \ge l_b$,
    \[ \tri*{\grad \tilde{\Phi}(\vecW), \grad \tilde{F}(\vecW)} \ge \frac14 g(l_b).\]
    \item For $F(\vecW) \le l_b$, we must have $\vecW\in\ball(\vecW^{\star}, R-1)$ and so 
    \[ \tri*{\grad \tilde{\Phi}(\vecW), \grad \tilde{F}(\vecW)} \ge g(F(\vecW)).\]
\end{enumerate}
We now construct a non-decreasing, infinitely differentiable function $\tilde{h}$ analogously to the definition of $\tilde{g}$ from \pref{subsec:proofopttosampling}. Notice $\frac14 g(L_{b,1})\le g(M)$ as $L_{b,1} \le M$ and as $g$ is non-decreasing. 
Now for some small constant $1 > \delta>0$, we can interpolate to create $\tilde{h}$ as follows:
\[ \tilde{h}(x) = \begin{cases}\frac18 g(x)=\frac18 \lambda x &: x \le l_b  \\ \text{smooth interpolation to }\frac14 g(l_b) &: l_b   < x < l_b+\delta \\ \frac14 g(l_b) &: l_b+\delta  \le x \le M \\ \text{ smooth interpolation to }\lambda x &: M< x < M+\delta  \\ \lambda x &: M+\delta\le x\end{cases}.\numberthis\label{eq:tildehdeflocal}\]
These interpolators can be defined analogously as in the definition of $\tilde{g}$, from \pref{subsec:proofopttosampling}, so that $\tilde{h}$ is non-decreasing and differentiable, and so that $\tilde{h}(x) \le \lambda x= g(x)$ for $x\in [M, M+\delta]$ (because we took $M$ so that we have $\lambda x \ge \frac14 g(L_{b,1})+1 \ge \frac14 g(l_b)+1$ for $x \ge M$), and $\tilde{h}(x) \le \frac18 g(l_b) \le \frac14 g(l_b)$ for $x \in [\tilde{l}, \tilde{l}+\delta]$. Moreover, note $\tilde{h}(x) = \lambda x$ for $x \ge \tilde{M} + 1$. 

Noting $\tilde{h}(x) \ge 0$, define
\[ \mprimenew = \lambda, \bprimenew = \lambda(M+1)',\numberthis\label{eq:tailprimesdeflocal}\]
where $M$ is defined as per \pref{eq:Mdeflocal}. Consequently we always have $\tilde{h}(x) \ge \mprimenew x - \bprimenew$.

Therefore, for all $\vecW\in\mathbb{R}^d$ we have
\[ \tri*{\grad \tilde{\Phi}(\vecW), \grad \tilde{F}(\vecW)} \ge \tilde{h}\prn*{F(\vecW)}.\numberthis\label{eq:localconditiontildePhi}\]
We can also check now similarly to Part 1 of \pref{subsec:proofopttosampling} that
\[ \nrm*{\grad^2 \tilde{\Phi}}_{\OPNORM} \le L' + B_{\Phi}\prn*{R^2 \cwgt + 4M}+\prn*{\cwgt+L'}\cdot 1+2B_{\Phi}\prn*{L'+R\cwgt},\]
where
\[ L' = \sup_{\vecW\in \ball(\vecW^{\star}, R)} \rho_{\Phi}\prn*{\Phi(\vecW)}.\numberthis\label{eq:localMLprimedef}\]
Consequently, $\tilde{\Phi}$ is again $\tilde{L}$-smooth, where we now define
\[ \tilde{L} :=\prn*{L' + B_{\Phi}\prn*{R^2 \cwgt + 4M}+\prn*{\cwgt+L'}\cdot 1+2B_{\Phi}\prn*{L'+R\cwgt}} \lor 2\bprimenew \lor 1.\numberthis\label{eq:localLtildedef}\]

\paragraph{Part 2: Proving a PI with the same idea as before.}
From here, the finish is analogous to the proof of \pref{thm:generalgflinearizabletopi}. We omit straightforward details that are checked verbatim as there. Because of \pref{eq:localconditiontildePhi}, letting $h(x)=\tilde{h}(x)+B$ where $B=\tilde{L}$, \pref{lem:lyapunovmethodsetup} gives for any test function $f$:
\begin{align*}
\int f(\vecW)^2 \frac{\tilde{h}\prn*{F(\vecW)}}{\tilde{h}\prn*{F(\vecW)}+\tilde{L}} \DERIV\mu_{\beta} &\le \frac1{\beta} \int \prn*{ \nrm*{\nabla f(\vecW)}^2+\frac{f(\vecW)^2}{h(\vecW)^2}\nrm*{\nabla \tilde{\Phi}(\vecW)}^2 - \frac{f(\vecW)^2}{h(\vecW)^2}\tri*{ \nabla h(\vecW), \nabla \tilde{\Phi}(\vecW)}} \DERIV\mu_{\beta} \\
&\hspace{1in}+ \frac1{\beta}\int f(\vecW)^2 \frac{\abs*{\Delta\tilde{\Phi}(\vecW)}}{h(\vecW)} \DERIV\mu_{\beta}.\numberthis\label{eq:pisetuplocal}
\end{align*}

\subparagraph{Step a: Upper bounding intermediate terms.}
Using $\tilde{L}$-smoothness of $\tilde{\Phi}$, and that $\tilde{h}(x) \ge \mprimenew x - \bprimenew$, $\tilde{L}/2 \ge \bprimenew$, $F(\vecW) \ge r_2\nrm*{\vecW-\vecW^{\star}}$, we obtain analogously to Step a in \pref{subsec:proofopttosampling} that
\[ \frac{\nrm*{\grad \tilde{\Phi}(\vecW)}^2-\tri*{\grad h(\vecW), \grad \tilde{\Phi}(\vecW)}}{h(\vecW)^2} \le C',\]
where now
\[ C' := \frac{8\prn*{R^2+8Mr_1}}{r_1\tilde{L}} \lor \frac{2\tilde{L}}{r_1 r_2^2 \mprimenew^2} \lor \frac{64M}{\tilde{L}}.\numberthis\label{eq:Cprimedefpilocal},\]
where $\tilde{L}$ is defined in \pref{eq:localLtildedef}, $\mprimenew=\lambda$, and $M$ is defined in \pref{eq:Mdeflocal}.

\subparagraph{Step b: Finishing the proof of PI identically to before.}
Consider an arbitrary test function $\psi$ and define $f$ in terms of $\psi$ identically as in \pref{subsec:proofopttosampling}, \pref{eq:fdeflyapunov}.

Now using $C'$ to upper bound the right hand side of \pref{eq:pisetuplocal}, we obtain
\[\int f(\vecW)^2 \frac{\tilde{h}\prn*{F(\vecW)}}{\tilde{h}\prn*{F(\vecW)}+\tilde{L}} \DERIV \mu_{\beta} \le \frac1{\beta}\int \nrm*{\grad f(\vecW)}^2 \DERIV \mu_{\beta} + \frac1{\beta}\int f(\vecW)^2 \prn*{d+C'} \DERIV \mu_{\beta}.\]
The only difference is the $\tilde{h}$ rather than $\tilde{g}$ in the left hand side, and that now $C'$ is defined in \pref{eq:Cprimedefpilocal}, rather than \pref{eq:Cprimedefpi}.

Now recalling that $\tilde{h}$ is non-decreasing, we obtain from the above that
\[ \int f(\vecW)^2 \frac{\tilde{h}(l_b)}{\tilde{h}(l_b)+\tilde{L}} \DERIV \mu_{\beta} \le \frac1{\beta}\int \nrm*{\grad f(\vecW)}^2 \DERIV \mu_{\beta} + \frac1{\beta}\int f(\vecW)^2 \prn*{d+C'} \DERIV \mu_{\beta} + \frac{\tilde{h}(l_b)}{\tilde{h}(l_b)+\tilde{L}}\int_{\cU} f(\vecW)^2 \DERIV \mu_{\beta}.\]
An analogous manipulation using \pref{ass:Funimodal} to upper bound $\int_{\cU} f(\vecW)^2 \DERIV \mu_{\beta}$, using the choice of $\alpha$, now proves 
\begin{align*}
\int f(\vecW)^2 \frac{\tilde{h}(l_b)}{\tilde{h}(l_b)+\tilde{L}} \DERIV \mu_{\beta}&\le \frac1{\beta}\int \nrm*{\grad f(\vecW)}^2 \DERIV \mu_{\beta} + \frac1{\beta}\int f(\vecW)^2 \prn*{d+C'} \DERIV \mu_{\beta} \\
&+ \frac{\tilde{h}(l_b)}{\tilde{h}(l_b)+\tilde{L}} \cdot \CPILOCAL \int \nrm*{\grad f(\vecW)}^2 \DERIV \mu_{\beta}.
\end{align*}
If $\beta \ge 2\prn*{1+\frac{\tilde{L}}{\tilde{h}(l_b)}} (d+C')=\Omega(d)$, where $\tilde{h}$ is as per \pref{eq:tildehdeflocal}, $C'$ is as per \pref{eq:Cprimedefpilocal}, $\tilde{L}$ is as per \pref{eq:localLtildedef}, we obtain
\[ \mathbb{V}_{\mu_{\beta}}\brk*{\psi}  \le \prn*{2\CPILOCAL + \frac{2}{\beta}\prn*{1+\frac{\tilde{L}}{\tilde{h}(l_b)}}} \int \nrm*{\grad \psi}^2 \DERIV \mu_{\beta}.\]
Recalling $\psi$ is an arbitrary test function, this gives the desired Poincar\'e Inequality. We furthermore verified that $\hat{F}$ is $O(1)$-smooth in \pref{lem:regularizeFsmooth}, so this finishes the proof. $\hfill\blacksquare$
}

\begin{remark}\label{rem:localconditionsamplingextend}
Note if we instead have an upper bound of the form $\nrm*{\grad F(\vecW)} \le L(\nrm*{\vecW-\vecW^{\star}}^s+1)$ rather than smoothness, one can instead add regularization in the form $\lambdareg (\nrm*{\vecW-\vecW^{\star}}^{s+1}+1)$. To capture other forms of $g(\cdot)$, one can perform similar ideas of lower bounding $g(x)$ by a $\tilde{g}(x)$ that grows linearly for large enough $x$, as done in \pref{subsec:proofopttosampling}. One can also tighten the PI to an LSI as in \pref{subsec:proofopttosampling}. These details follow the exact same argument as in \pref{subsec:proofopttosampling} and are straightforward to verify. 
\end{remark}
\begin{remark}\label{rem:computeprobdependentparam}
Notice to construct $\grad F$, all the problem-dependent parameters used in the construction can be computed with oracle access to $F$, knowledge of $\vecW^{\star}, R$, except for $\rho_{\Phi}(M)$ (to define $c_F$) and $L$ (to define $\lambdareg$). However, for $\rho_{\Phi}(M)$, $L$, it suffices to use a \textit{upper bound} on them, as can be seen through the above proof. Consequently we can construct a suitable $\hat{F}$ via appropriate cross-validation on these parameters.
\end{remark}

\section{Technical Helper Results}
\begin{lemma}[Lemma 2.1, \citet{srebro2010smoothness}]\label{lem:smoothnessgradbound}
If some $G$ is non-negative and $L$-smooth, then 
\[ \nrm*{\grad G(\vecW)} \le \sqrt{4L G(\vecW)}.\]
\end{lemma}
% \begin{lemma}[Theorem 2 of \cite{karimi2016linear}]\label{lem:FPLquadgrowth}
% If $F$ is P\LCHAR with parameter $\lambda$, then
% \[ F(\vecW) \ge \frac{\lambda}4 \nrm*{\vecW-\vecW_p}^2,\]
% where $\vecW_p$ denotes the projection of $\vecW$ onto $\cW^{\star}$.
% %If $F$ is P\LCHAR with parameter $\lambda$, then $F(\vecW) \ge \frac{\lambda}8 \nrm*{\vecW}^2 - \frac{\lambda}2 \prn*{\sup_{\vecW' \in \cW^{\star}}\nrm*{\vecW'}}^2$. Moreover 
% \end{lemma}
% \proof{
% By Theorem 2 of \cite{karimi2016linear}, we know that if $F$ is P\LCHAR with parameter $\lambda$, then
% \[ F(\vecW) \ge \frac{\lambda}4 \nrm*{\vecW-\vecW_p}^2,\]
% where $\vecW_p$ denotes the projection of $\vecW$ onto $\cW^{\star}$. Rewrite
% \begin{align*}
% \nrm*{\vecW-\vecW_p}^2&\ge\nrm*{\vecW}^2 -2\nrm*{\vecW}\nrm*{\vecW_p} \ge \nrm*{\vecW}^2 -2\nrm*{\vecW}\prn*{\sup_{\vecW' \in \cW^{\star}}\nrm*{\vecW'}}\ge \nrm*{\vecW}^2 - \prn*{\frac{\nrm*{\vecW}^2 }2+2\crl*{\sup_{\vecW' \in \cW^{\star}}\nrm*{\vecW'}}^2}.
% \end{align*}
% The last step uses AM-GM. Rearranging gives the Lemma.
% }
\begin{lemma}\label{lem:initdivergencecontrolled}
Suppose $F$ is $L$-H\"{o}lder continuous with parameter $s \in (0,1]$. Let $M = \int \nrm*{\cdot} \DERIV \mu_{\beta}$. Additionally define $\hat{F}(\vecW) = F(\vecW)+\frac{\gamma}{2\beta}\max\prn*{0, \nrm*{\vecW}-R}^2$ for $\gamma>0$, $\hat{\mu}_{\beta}=e^{-\beta \hat{F}}/Z$. With initialization $\pi_0 \sim \cN\prn*{\vecOrigin, \frac{1}{2\beta L+\gamma} I_d}$, we have the following:
\[ \ln\prn*{\chi^2(\pi_0||\mu_{\beta})+1}, \KL(\pi_0||\mu_{\beta}) \le \beta L + \beta F(\vecOrigin) + 2 + \frac{d}{2}\ln\prn*{4M^2\prn*{\beta L+\gamma/2}}, \]
\[ \ln\prn*{\chi^2(\pi_0||\hat{\mu}_{\beta})+1}, \KL(\pi_0||\hat{\mu}_{\beta}) \le \beta L + \beta F(\vecOrigin) + 2 + \frac{d}{2}\ln\prn*{4\hat{M}^2\prn*{\beta L+\gamma/2}}.\]
\end{lemma}
\proof{
Since R\'enyi divergence (for the definition, see e.g. \citet{chewi2024log}) is increasing in its order, and as $\KL$ divergence is R\'enyi divergence of order 1 and $\ln(\chi^2+1)$ is R\'enyi divergence of order 2, it suffices to show these upper bounds for the R\'enyi divergence of order $\infty$, $\cR_{\infty}(\cdot||\cdot)$. This is the supremum of the log ratio of the probability density functions. Now the proof follows by exactly the same argument as the proof of Lemmas 31 and 32 from \citet{chewi21analysis}. We highlight it here by proving the second upper bound. Let $V=\beta F$, $\hat{V}=\beta\hat{F}$. Then we can compare the ratio of their unnormalized densities:
\begin{align*}
\exp\prn*{\hat{V}(\vecW)-\prn*{L\beta+\frac{\gamma}2} \nrm*{\vecW}^2} &\le \exp\prn*{\hat{V}(\vecW)-\hat{V}(\vecOrigin)+\hat{V}(\vecOrigin)-\prn*{\beta L+\frac{\gamma}2} \nrm*{\vecW}^2} \\
&\le \exp\prn*{\beta L \nrm*{\vecW}^{s+1} + \frac{\gamma}2\max\{0, \nrm*{\vecW}-R\}^2+\beta F(\vecOrigin) -\prn*{L\beta+\frac{\gamma}2} \nrm*{\vecW}^2} \\
&\le \exp\prn*{\beta L + \beta F(\vecOrigin)}.
\end{align*}
Here we used the inequality $x^{s+1}\le x^2+1$ for all $x \ge 0$ (as $s \le 1$) and $\hat{V}(\vecW)-\hat{V}(\vecOrigin) = \beta\prn*{F(\vecW)-F(\vecOrigin)}+\frac{\gamma}2\max\{0, \nrm*{\vecW}-R\}^2 \le \beta L\nrm*{\vecW}^{s+1}+\frac{\gamma}2\max\{0, \nrm*{\vecW}-R\}^2$.

Now analogously to the proof of Lemma 31 of \citet{chewi21analysis}, we compare the partition functions, arguing through the intermediate quantity $\int \exp\prn*{-\hat{V}(\vecW) - \delta \nrm*{\vecW}^2} \DERIV \vecW$:
\[ \frac{\int \exp\prn*{-\hat{V}(\vecW) - \delta \nrm*{\vecW}^2} \DERIV \vecW}{\int \exp\prn*{-\hat{V}(\vecW)} \DERIV \vecW} \ge \frac12 \exp\prn*{-4\delta \hat{M}^2}, \frac{\int \exp\prn*{-\hat{V}(\vecW) - \delta \nrm*{\vecW}^2} \DERIV \vecW}{\prn*{\frac{\pi}{\beta L+\gamma/2}}^{d/2}} \le \prn*{\frac{\beta L+\gamma/2}{\delta}}^{d/2}.\]
Taking $\delta=\frac1{4\hat{M}^2}$ and rearranging the above gives
\[ \cR_{\infty}\prn*{\pi_0||\hat{\mu}_{\beta}} \le \beta L + \beta F(\vecOrigin) + 2 + \frac{d}{2}\ln\prn*{4\hat{M}^2\prn*{\beta L+\gamma/2}}.\]
For the first upper bound, we do the same steps with $V$ in place of $\hat{V}$. The first upper bound still holds, and the second two inequalities comparing the partition functions still hold, except $\hat{M}$ is replaced by $M$ instead. Taking $\delta=\frac1{4M^2}$, we obtain the first inequality. $\hfill\blacksquare$
}
\begin{remark}\label{rem:initdivergencenormfactor}
For an upper bound on $M$ and $\hat{M}$, note if $F$ is $L$-smooth and dissipative, that is $\tri*{\vecW, \grad F(\vecW)} \ge m\nrm*{\vecW}^2-b$ for $m,b>0$, then following the notation from \pref{thm:generalgflinearizabletopi}, Cauchy-Schwartz gives that 
\[ M^2 \le S \le \frac{b+d/\beta}m = O(1).\]
The bound on $S$ follows from \citet{raginsky2017non}. If $F$ is dissipative with parameters $m,b$ it is easy to check $\hat{F}$ is also dissipative with the same parameters, so we also have the same upper bound on $\hat{M}$. Notice also for $F = \nrm*{\vecW}^{\alpha}$ and $\beta=\Omega(d)$ that $M=O(1)$. Therefore, we believe it is reasonable to suppose the right hand side of the above two lines is $\tilde{O}(\beta)$ for $\beta=\Omega(d)$.\footnote{Since we are in the low temperature setting corresponding to optimization, the norm is a $\beta$ factor \textit{smaller} than in the standard sampling setting. }
\end{remark}
\begin{remark}\label{rem:wstarmoment}
As will be clear in the following proof, it is also possible to replace each instance of $\vecW$ with $\vecW-\vecW^{\star}$ for a fixed $\vecW^{\star}\in\cW^{\star}$, if we know such a $\vecW^{\star}$. Our initialization then changes to Gaussian initialization centered at $\vecW^{\star}$. This can be done to give somewhat better bounds, but we do not pursue it for simplicity.
\end{remark}

\begin{lemma}\label{lem:controlregularizedklinit}
Suppose $F$ satisfies \pref{ass:tamedassumption}. Taking $\pi_0(\vecW) \propto \exp\prn*{-2\nrm*{\vecW}^{2s_3'}}$ where $s_3' = \max(s_3+\frac12, r+1)$, we have 
\[ \KL(\pi_0||\mu_{\beta}), \KL(\pi_0||\hat{\mu}_{\beta}) \le \tilde{O}(d\beta).\]
Here $\hat{\mu}_{\beta}$ comes from Theorem 3, \citet{lytras2024tamed}; it is defined explicitly in our proof of \pref{thm:generalgflinearizabletopi}.
\end{lemma}
\proof{
First notice by \pref{ass:tamedassumption}, we can check that for some $L_1, L_2>0$, we have $F(\vecW) \le L_1 \nrm*{\vecW}^{2s_3+1}+L_2$. Thus $F(\vecW), F(\vecW)+ \frac{\eta}{\beta}\nrm*{\vecW}^{2r+2} \le L_1 \nrm*{\vecW}^{2s_3'}+L_2$ where $s_3' = \max(s_3+\frac12, r+1)$. Now we adopt the proof of Lemma 5, \citet{raginsky2017non}. Analogously to how C.11 was derived there, we have 
\[ \KL(\pi_0||\mu_{\beta}) \le \log \nrm*{\pi_0}_{\infty}+\log \Lambda + \beta \int_{\mathbb{R}^d} \pi_0(\vecW) F(\vecW)\DERIV \vecW,\numberthis\label{eq:klinitcontrol}\]
where $\Lambda$ denotes the partition function of $\mu_{\beta}$. We upper bound each part of the above sum:
\begin{itemize}
    \item The partition function: By the second part of \pref{ass:linearizableoutsideball}, we have 
    \begin{align*}
    \Lambda &= \int_{\mathbb{R}^d} e^{-\beta F(\vecW)}\DERIV \vecW \\
    &\le e^{\beta \sup_{\vecW \in \ball(\vecW^{\star}, R)} F(\vecW)} \int_{\mathbb{R}^d} e^{-\beta r_2 \nrm*{\vecW-\vecW^{\star}}} \DERIV \vecW \\
    &= e^{\beta \sup_{\vecW \in \ball(\vecW^{\star}, R)} F(\vecW)} \frac{2\pi^{d/2}}{\Gamma(d/2)} (\beta r_2)^{-d} \Gamma(d) \\
    &\le e^{\beta \sup_{\vecW \in \ball(\vecW^{\star}, R)} F(\vecW)} \cdot \frac{4\pi^{d/2} \cdot d^d \sqrt{\pi}}{(\beta r_2)^d}. 
    \end{align*}
    Here $\Gamma(\cdot)$ denotes the Gamma function. We evaluated the integral by Lemma 8.5 of \citet{chen2024langevin}, and used straightforward properties of $\Gamma(\cdot)$ in the above.
    \item The $\infty$ norm: Since $\pi_0(\vecW) \propto \exp\prn*{-2\nrm*{\vecW}^{2s_3'}}$, it follows that its normalizing constant is
    \[ Z = \int_{\mathbb{R}^d} \exp\prn*{-2\nrm*{\vecW}^{2s_3'}} \DERIV \vecW = \frac{2\pi^{d/2}}{\Gamma(d/2)} \cdot \frac1{2s_3'} 2^{-\frac{d-2}{2s_3'}} \Gamma\prn*{\frac{d}{2s_3'}} \ge \frac{\pi^{d/2}}{2 \pi^{1/2} s_3' d^{d/2}2^{\frac{d-2}{2s_3'}}}.\]
    The computation of this integral follows from analogous steps as in Lemmas 5.1 and 8.5, \citet{chen2024langevin} (there the result is stated for a particular range on $s_3'$, but this is not needed). It follows that for all $\vecW\in\mathbb{R}^d$,
    \[ \log \pi_0 = -2\nrm*{\vecW}^{2s} - \log Z \le -\log Z \le \log (2s_3' \pi^{1/2}) + \frac{d}{2} \log\prn*{\frac{d2^{2s'_3}}{\pi}}. \]
    \item The last term: Since $F(\vecW) \le L_1 \nrm*{\vecW}^{2s_3'}+L_2$, 
    \[ \int_{\mathbb{R}^d} \pi_0(\vecW) F(\vecW) \DERIV \vecW \le \int_{\mathbb{R}^d} \pi_0(\vecW) F(\vecW) \DERIV \vecW \le L_1 \int_{\mathbb{R}^d} \pi_0(\vecW) \nrm*{\vecW}^{2s_3'} \DERIV \vecW + L_2.\]
    By Jensen's Inequality, we have 
    \[ \int_{\mathbb{R}^d} \pi_0(\vecW) \nrm*{\vecW}^{2s_3'} = \mathbb{E}_{\pi_0}\brk*{\log \exp \crl*{\nrm*{\vecW}^{2s_3'}}} \le \log \mathbb{E}_{\pi_0}\brk*{\exp \crl*{\nrm*{\vecW}^{2s_3'}}}.\]
    Let $Z$ denote the normalizing constant of $\pi_0$, as in the above. Note by choice of $\pi_0$, 
    \begin{align*}
    \mathbb{E}_{\pi_0}\brk*{\exp \prn*{\nrm*{\vecW}^{2s_3'}}} &= \frac1{Z}\int \exp\prn*{\nrm*{\vecW}^{2s_3'}-2\nrm{\vecW}^{2s_3'}}\DERIV \vecW \\
    &= \frac1{Z} \int \exp\prn*{-\nrm{\vecW}^{2s_3'}}\DERIV \vecW\\
    &= \frac{\frac{2\pi^{d/2}}{\Gamma(d/2)} \cdot \frac1{2s_3'}  \Gamma\prn*{\frac{d}{2s_3'}} }{\frac{2\pi^{d/2}}{\Gamma(d/2)} \cdot \frac1{2s_3'} 2^{-\frac{d-2}{2s_3'}} \Gamma\prn*{\frac{d}{2s_3'}}} = e^{\ln 2 \cdot \frac{d-2}{2s_3'}}.
    \end{align*}
    Here, we evaluated the above integral analogously to how we computed $Z$. Putting all this together yields 
    \[ \int_{\mathbb{R}^d} \pi_0(\vecW) F(\vecW) \DERIV \vecW \le L_1 \cdot \frac{d-2}{s_3'} + L_2.\]
\end{itemize}
Putting all these steps together yields 
\begin{align*}
&\KL(\pi_0||\mu_{\beta}) \\
&\le \log \nrm*{\pi_0}_{\infty}+\log \Lambda + \beta \int_{\mathbb{R}^d} \pi_0(\vecW) F(\vecW)\DERIV \vecW \\
&\le \log (2s_3' \pi^{1/2}) + \frac{d}{2} \log\prn*{\frac{d2^{2s_3'}}{\pi}} + \beta \sup_{\vecW \in \ball(\vecW^{\star}, R)} F(\vecW) +d \log\prn*{\frac{4\pi^{\frac12 + \frac1{2d}} d}{\beta r_2}}+ \beta \prn*{L_1 \cdot \frac{d-2}{s_3'} + L_2}\\
&= \tilde{O}\prn*{d\beta}.
\end{align*}
The calculation for $\KL(\pi_0||\hat{\mu}_{\beta})$ follows from an analogous argument, using \pref{eq:klinitcontrol}. We just replace $F(\vecW)$ by $F(\vecW)+ \frac{\eta}{\beta}\nrm*{\vecW}^{2r+2}$, and thanks to the definition of $s'_3$, all the bounds above go through. $\hfill\blacksquare$
}

\begin{lemma}\label{lem:constructsuchachi}
We can construct a $\chi(\vecW) \in [0,1]$ such that:
\begin{itemize}
    \item $\chi \equiv 0$ on $B(\vecW^{\star}, R)$ and $\chi \equiv 1$ on $B(\vecW^{\star}, R+1)^c$.
    \item $\chi(\vecW)$ is differentiable to all orders.
    \item $\nrm*{\grad \chi(\vecW)}, \nrm*{\grad^2 \chi(\vecW)}_{\OPNORM} \le B$ for some universal constant $B>0$.
    \item $\tri*{\grad \chi(\vecW), \grad F(\vecW)} \ge 0$.
\end{itemize}
\end{lemma}
\proof{
The construction is to let 
\[ \chi(\vecW) = \begin{cases} 0:\nrm*{\vecW-\vecW^{\star}} \le R \\ 1:\nrm*{\vecW-\vecW^{\star}} \ge R+1 \\ \frac{e^{-\frac{1}{(\nrm*{\vecW-\vecW^{\star}}-R)^{2}}}}{e^{-\frac{1}{(\nrm*{\vecW-\vecW^{\star}}-R)^{2}}}+e^{-\frac{1}{1-(\nrm*{\vecW-\vecW^{\star}}-R)^{2}}}}:R<\nrm*{\vecW-\vecW^{\star}}<R+1.\end{cases}\]
Clearly $\chi \in [0,1]$ and also the first property is satisfied. The second property is satisfied because $\tilde{\chi}(x)=\frac{e^{-\frac{1}{x^{2}}}}{e^{-\frac{1}{x^{2}}}+e^{-\frac{1}{1-x^{2}}}}$ is infinitely differentiable on $(0,1)$, and $\tilde{\chi}(0)=0$, $\tilde{\chi}(1)=1$. In particular, on $(0,1)$, $e^{-\frac{1}{x^{2}}}$ and $e^{-\frac{1}{1-x^{2}}}$ are both infinitely differentiable, which can be verified by a straightforward induction argument, and their sum is lower bounded by a constant $[0,1]$. Therefore, the quotient $\tilde{\chi}(x)$ is infinitely differentiable. Therefore, $\tilde{\chi}$ interpolates between 0 and 1 on $(0,1)$ in an infinitely differentiable way. Because $R>0$, the composition of $\tilde{\chi}$ and $\nrm*{\vecW-\vecW^{\star}}-R$ is infinitely differentiable, as both these maps are. 

For the next two properties, we directly do the calculation. They are both obvious when $\nrm*{\vecW-\vecW^{\star}} \le R$ or $\nrm*{\vecW-\vecW^{\star}} \ge R+1$, so we check these two properties when $R<\nrm*{\vecW-\vecW^{\star}}<R+1$. We first prove the last property. We do so using the intuitive geometric approach of comparing the angle that $\grad \chi(\vecW)$ and $\grad F(\vecW)$ make with $\vecW-\vecW^{\star}$, and showing the sum of their angles is at most $\frac{\pi}2$.

First, by \pref{ass:linearizableoutsideball}, we have for $R+1 > \nrm*{\vecW-\vecW^{\star}} > R$ that
\[ \frac{\tri*{\vecW-\vecW^{\star}, \grad F(\vecW)}}{\nrm*{\vecW-\vecW^{\star}} \nrm*{\grad F(\vecW)}} \ge \frac{r_1 F(\vecW)}{\nrm*{\vecW-\vecW^{\star}} \nrm*{\grad F(\vecW)}} \ge 0. \]
This means 
\[ \theta\tri*{\grad F(\vecW), \vecW-\vecW^{\star}} \le \cos^{-1}(0)= \frac{\pi}2.\numberthis\label{eq:gradFangle}\]
Notice $\grad(\nrm*{\vecW-\vecW^{\star}})=\frac{\vecW-\vecW^{\star}}{\nrm*{\vecW-\vecW^{\star}}}$. Thus, by Chain Rule, we have
\begin{align*}
&\grad \chi(\vecW) \\
&= \frac{e^{-\frac{1}{(\nrm*{\vecW-\vecW^{\star}}-R)^{2}}} \cdot \frac{2}{(\nrm*{\vecW-\vecW^{\star}}-R)^3} \cdot \frac{\vecW-\vecW^{\star}}{\nrm*{\vecW-\vecW^{\star}}}}{e^{-\frac{1}{(\nrm*{\vecW-\vecW^{\star}}-R)^{2}}}+e^{-\frac{1}{1-(\nrm*{\vecW-\vecW^{\star}}-R)^{2}}}} \\
&+ \frac{e^{-\frac1{(\nrm*{\vecW-\vecW^{\star}}-R)^2}}\prn*{e^{-\frac1{(\nrm*{\vecW-\vecW^{\star}}-R)^2}}\cdot\frac{2}{(\nrm*{\vecW-\vecW^{\star}}-R)^3} \cdot \frac{\vecW-\vecW^{\star}}{\nrm*{\vecW-\vecW^{\star}}}+e^{-\frac{1}{1-(\nrm*{\vecW-\vecW^{\star}}-R)^{2}}}\cdot \frac{-2(\nrm*{\vecW-\vecW^{\star}}-R)}{(1-(\nrm*{\vecW-\vecW^{\star}}-R)^2)^2}\cdot \frac{\vecW-\vecW^{\star}}{\nrm*{\vecW-\vecW^{\star}}}}}{\prn*{e^{-\frac{1}{(\nrm*{\vecW-\vecW^{\star}}-R)^{2}}}+e^{-\frac{1}{1-(\nrm*{\vecW-\vecW^{\star}}-R)^{2}}}}^2}.
\end{align*}
Thus,
\[ \grad \chi(\vecW)=\tilde{p}(\nrm*{\vecW-\vecW^{\star}}-R) \cdot \frac{\vecW-\vecW^{\star}}{\nrm*{\vecW-\vecW^{\star}}},\]
where 
\[ \tilde{p}(x) = \frac{e^{-\frac{1}{x^2}} \cdot \frac{2}{x^3}}{e^{-\frac{1}{x^2}}+e^{-\frac{1}{1-x^2}}}+\frac{e^{-\frac{1}{x^2}}\prn*{e^{-\frac1{x^2}} \cdot \frac{2}{x^3} + e^{-\frac1{1-x^2}} \cdot \frac{-2x}{(1-x^2)^2}}}{(e^{-\frac{1}{x^2}}+e^{-\frac{1}{1-x^2}})^2}\]
is just a scalar. In \pref{lem:scalenonnegative}, we prove $\tilde{p}(x) \ge 0$ for all $x \in [0,1]$, therefore 
\[ \tri*{\grad \chi(\vecW), \vecW-\vecW^{\star}} = \frac{\tilde{p}(\nrm*{\vecW-\vecW^{\star}}-R)}{\nrm*{\vecW-\vecW^{\star}}} \nrm*{\vecW-\vecW^{\star}}^2 = \nrm*{\grad \chi(\vecW)} \nrm*{\vecW-\vecW^{\star}}.\]
Thus, the vectors $\grad \chi(\vecW), \vecW-\vecW^{\star}$ are collinear and point in the same direction: 
\[ \theta\tri*{\grad \chi(\vecW), \vecW-\vecW^{\star}}=0.\numberthis\label{eq:gradchiangle}\]
Combining \pref{eq:gradchiangle} and \pref{eq:gradFangle}, it is clear that $\theta\tri*{\grad \chi(\vecW), \grad F(\vecW)} \le \frac{\pi}2$, hence $\tri*{\grad \chi(\vecW), \grad F(\vecW)} \ge 0$.

For the third property, we clearly only need to check it when $\nrm*{\vecW-\vecW^{\star}} \in [R,R+1]$. The above calculation verifies it directly for the gradient Euclidean norm, as it shows that
\[ \nrm*{\grad \chi(\vecW)} = \tilde{p}(\nrm*{\vecW-\vecW^{\star}}-R) \le \sup_{t \in (0,1)}\tilde{p}(t).\]
We conclude this part for the gradient, noting $\tilde{p}$ is a univariate function with no explicit $d$ dependence, which can be extended to be bounded and differentiable to all orders on $[0,1]$ (because $\lim_{t\rightarrow 0}e^{-1/t} \frac1{t^p}=0$ for all $p<\infty$, and similarly for the limits to 1). For the Hessian operator norm, applying Chain Rule to the above shows
\begin{align*}
\grad^2 \chi(\vecW) &= \tilde{p}'(\nrm*{\vecW-\vecW^{\star}}-R) \cdot\frac{1}{\nrm*{\vecW-\vecW^{\star}}^2} (\vecW-\vecW^{\star})(\vecW-\vecW^{\star})^T \\
&\hspace{1in}+ \tilde{p}(\nrm*{\vecW-\vecW^{\star}}-R) \cdot \frac{1}{\nrm*{\vecW-\vecW^{\star}}} I_d\\
&\hspace{1in}- \tilde{p}(\nrm*{\vecW-\vecW^{\star}}-R) \cdot \frac1{\nrm*{\vecW-\vecW^{\star}}^2} \cdot \frac1{\nrm*{\vecW-\vecW^{\star}}}(\vecW-\vecW^{\star})(\vecW-\vecW^{\star})^T.
\end{align*}
The same rationale as before justifies that $\tilde{p}'(\cdot)$ is a univariate function with no explicit $d$ dependence, which can be extended to be bounded and differentiable to all orders on $[0,1]$. Recalling $\nrm*{\vecW-\vecW^{\star}} \in (R,R+1)$, it follows that $\tilde{p}'(\nrm*{\vecW-\vecW^{\star}}-R)$ is upper bounded by universal constant $\sup_{t \in (0,1)}\tilde{p}'(t)<\infty$. Using the fact that 
\[ \nrm*{(\vecW-\vecW^{\star})(\vecW-\vecW^{\star})^T}_{\OPNORM} \le \nrm*{\vecW-\vecW^{\star}}^2\]
when $R+1 > \nrm*{\vecW-\vecW^{\star}} > R$, we obtain
\begin{align*}
\nrm*{\grad^2 \chi(\vecW)}_{\OPNORM} &\le \sup_{t \in (0,1)}\tilde{p}'(t) \cdot \frac{\nrm*{\vecW-\vecW^{\star}}^2}{\nrm*{\vecW-\vecW^{\star}}^2}+\sup_{t \in (0,1)}\tilde{p}(t) \cdot \frac1{R}+\sup_{t \in (0,1)}\tilde{p}(t) \cdot \frac{\nrm*{\vecW-\vecW^{\star}}^2}{\nrm*{\vecW-\vecW^{\star}}^3} \\
&\le \sup_{t \in (0,1)}\tilde{p}'(t)+2\sup_{t \in (0,1)}\tilde{p}(t).
\end{align*}
The last step follows as we have $R \ge 1$ without loss of generality. The proof is complete. $\hfill\blacksquare$
}
\begin{lemma}\label{lem:scalenonnegative}
For $x \in [0,1]$, we have 
\[\tilde{p}(x) = \frac{e^{-\frac{1}{x^2}} \cdot \frac{2}{x^3}}{e^{-\frac{1}{x^2}}+e^{-\frac{1}{1-x^2}}}+\frac{e^{-\frac{1}{x^2}}\prn*{e^{-\frac1{x^2}} \cdot \frac{2}{x^3} + e^{-\frac1{1-x^2}} \cdot \frac{-2x}{(1-x^2)^2}}}{(e^{-\frac{1}{x^2}}+e^{-\frac{1}{1-x^2}})^2} \ge 0.\]
\end{lemma}
\proof{
Simplifying, it is enough to show that 
\[ \frac2{x^3}\prn*{e^{-\frac{1}{x^2}}+e^{-\frac{1}{1-x^2}}} + e^{-\frac1{x^2}} \cdot \frac{2}{x^3} + e^{-\frac1{1-x^2}} \cdot \frac{-2x}{(1-x^2)^2} \ge 0.\]
If $x \le \frac{\sqrt{2}}2$, that is $x^2 \le \frac12$, then notice $\frac{1}{x^3} \ge \frac{x}{(1-x^2)^2}$, which proves the above. Thus from now on suppose $x \ge \frac{\sqrt{2}}2$. Rewrite the above desired inequality as 
\begin{align*}
&\frac{1}{x^3} \prn*{2e^{-\frac{1}{x^2}} + e^{-\frac{1}{1-x^2}}} - \frac{x}{(1-x^2)^2} e^{-\frac1{1-x^2}} \ge 0 \\
\iff &(1-x^2)^2 (2e^{-\frac{1}{x^2}} + e^{-\frac{1}{1-x^2}}) \ge x^4 e^{-\frac1{1-x^2}} \\
\iff &2(1-x^2)^2 e^{-\frac1{x^2}} \ge (2x^2-1)  e^{-\frac1{1-x^2}} \\
\iff &e^{\frac1{1-x^2}-\frac1{x^2}} \ge \frac{2x^2-1}{2(1-x^2)^2}.
\end{align*}
Notice $\frac1{1-x^2}-\frac1{x^2} \ge 0$ since $2x^2 \ge 1$, thus by series expansion, it suffices to show 
\[ 1+\frac1{1-x^2}-\frac1{x^2}+\frac12\prn*{\frac1{1-x^2}-\frac1{x^2}}^2+ \frac16\prn*{\frac1{1-x^2}-\frac1{x^2}}^3 \ge\frac{2x^2-1}{2(1-x^2)^2}. \]
Explicitly expanding the above, because $0 \le x \le 1$, this is equivalent to the inequality
\[ 6x^{6}\left(1-x^{2}\right)^{3}+6x^{4}\left(1-x^{2}\right)^{2}\left(2x^{2}-1\right)+\left(2x^{2}-1\right)^{3}+3\left(2x^{2}-1\right)^{2}x^{2}\left(1-x^{2}\right)-3x^{6}\left(1-x^{2}\right)\left(2x^{2}-1\right)\ge 0\]
for $x \in [\frac{\sqrt{2}}2, 1]$. Replacing $x^2$ by $x$, the left hand side of the above expands to
\[ h(x) = -6x^{6}+36x^{5}-69x^4+65x^3-33x^2+9x-1.\]
We want to show $h(x) \ge 0$ for $x \in \brk*{\frac12, 1}$. This can be directly checked by computer, but we also give a proof by hand. Noting $h(\frac12), h'(\frac12), h''(\frac12) \ge 0$, it is enough to show $h'''(x) \ge 0$ on $[\frac12, 1]$, or equivalently 
\[h_3(x):=-120x^{3}+360x^{2}-276x+65 \ge 0 \forall x \in [\frac12, 1].\]
However differentiating and applying the quadratic formula, we can check $h_3(x)$ attains a minimum value on $[\frac12, 1]$ at $x=1-\sqrt{\frac7{30}} \approx 0.517$, and that this minimum value is strictly positive. This completes the proof. $\hfill\blacksquare$
}

\end{document}